\documentclass[lettersize,journal]{IEEEtran}

\usepackage{amsmath,amsfonts}
\usepackage{algorithmic}
\usepackage{array}
\usepackage[caption=false,font=normalsize,labelfont=sf,textfont=sf]{subfig}
\usepackage{textcomp}
\usepackage{stfloats}
\usepackage{url}
\usepackage{verbatim}
\usepackage{graphicx}
\usepackage{multirow}
\usepackage{xcolor}
\usepackage{subfloat}
\usepackage{hhline}
\usepackage{booktabs}
\usepackage[caption=false]{subfig}
\def\BibTeX{{\rm B\kern-.05em{\sc i\kern-.025em b}\kern-.08em
    T\kern-.1667em\lower.7ex\hbox{E}\kern-.125emX}}
\usepackage{balance}

\usepackage{hyperref}
\hypersetup{
    colorlinks=true,
    linkcolor=blue,
    filecolor=magenta,      
    urlcolor=cyan,
    pdftitle={Overleaf Example},
    pdfpagemode=FullScreen,
    }

\def\thesubsubsectiondis{\unskip\arabic{subsubsection})}

\begin{document}
\title{Low-Light Image and Video Enhancement: A Comprehensive Survey and Beyond}
\author{Shen Zheng, Yiling Ma*, Jinqian Pan*, Changjie Lu*, Gaurav Gupta
\thanks{Shen Zheng is with the School of Computer Science, Carnegie Mellon University, Pittsburgh, PA, USA (email: shenzhen@andrew.cmu.edu). Jinqian Pan is with the Center for Data Science, New York University, New York, NY, USA (email: jp6218@nyu.edu) . Yiling Ma, Changjie Lu, and Gaurav Gupta is with the School of Science and Technology, Wenzhou-Kean University, Wenzhou, Zhejiang, China (email: mayili@kean.edu, lucha@kean.edu, ggupta@kean.edu).}
\thanks{* Indicates equal contribution. }
}


\markboth{Journal of \LaTeX\ Class Files,~Vol.~18, No.~9, September~2020}%
{How to Use the IEEEtran \LaTeX \ Templates}

\maketitle

\begin{abstract}

This paper presents a comprehensive survey of low-light image and video enhancement, addressing two primary challenges in the field. The first challenge is the prevalence of mixed over-/under-exposed images, which are not adequately addressed by existing methods. In response, this work introduces two enhanced variants of the SICE dataset: SICE\_Grad and SICE\_Mix, designed to better represent these complexities. The second challenge is the scarcity of suitable low-light video datasets for training and testing. To address this, the paper introduces the Night Wenzhou dataset, a large-scale, high-resolution video collection that features challenging fast-moving aerial scenes and streetscapes with varied illuminations and degradation. This study also conducts an extensive analysis of key techniques and performs comparative experiments using the proposed and current benchmark datasets. The survey concludes by highlighting emerging applications, discussing unresolved challenges, and suggesting future research directions within the LLIE community. The datasets are available at \href{https://github.com/ShenZheng2000/LLIE_Survey}{https://github.com/ShenZheng2000/LLIE\_Survey}.

   
\end{abstract}

\begin{IEEEkeywords}
Low-Light Image and Video Enhancement, Low-Level Vision, Deep Learning, Computational Photography.
\end{IEEEkeywords}


\section{Introduction}

\begin{figure}[t]
    \centering
    
    \subfloat[Under Exposure]{\includegraphics[width=2.8cm]{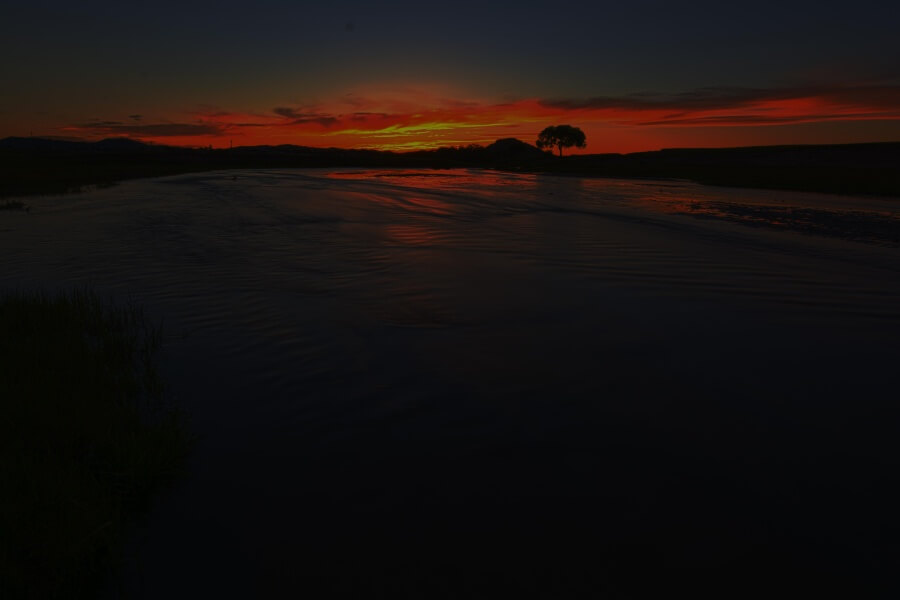}}
    \subfloat[Over Exposure]{\includegraphics[width=2.8cm]{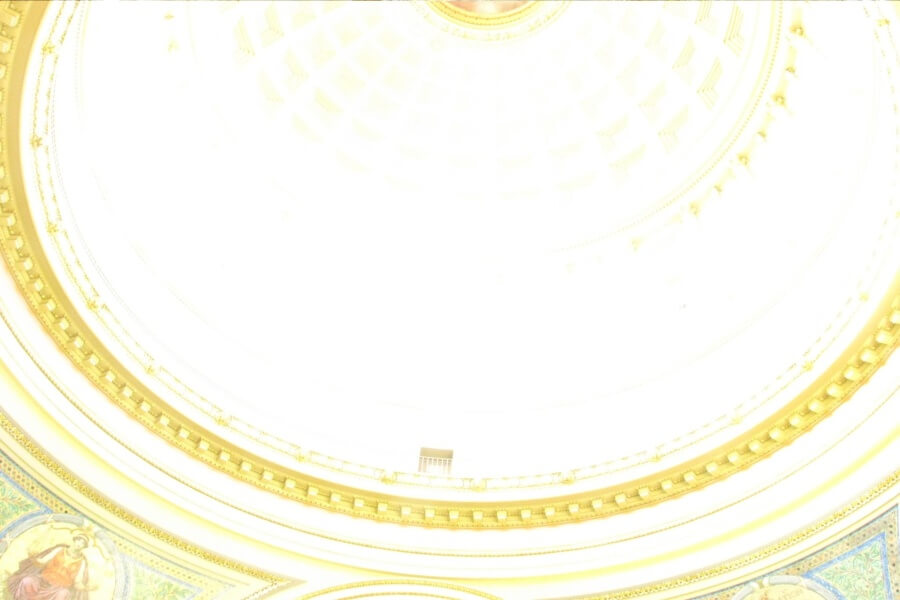}}
     \subfloat[Uneven Exposure]{\includegraphics[width=2.8cm]{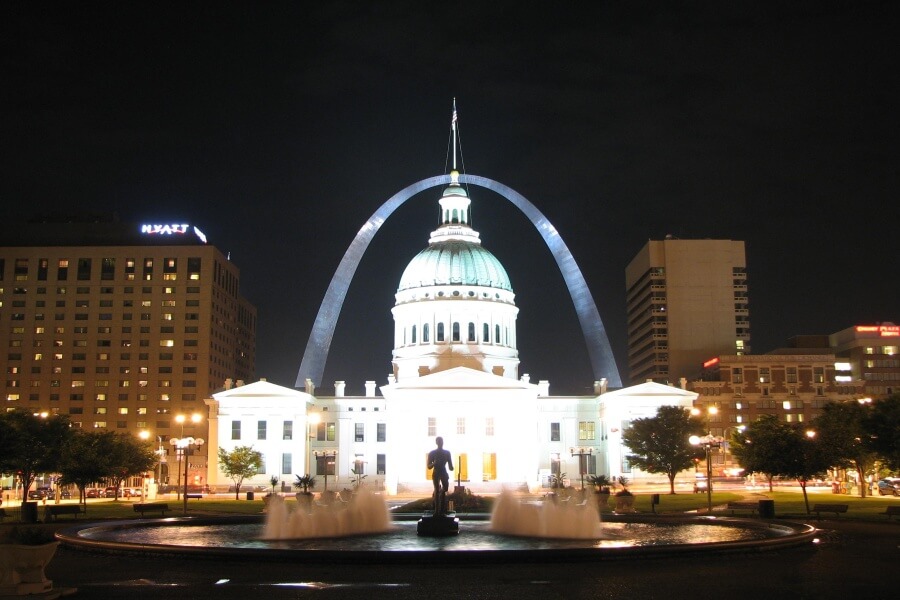} }
      \\
      \subfloat[Front Light]{\includegraphics[width=2.8cm]{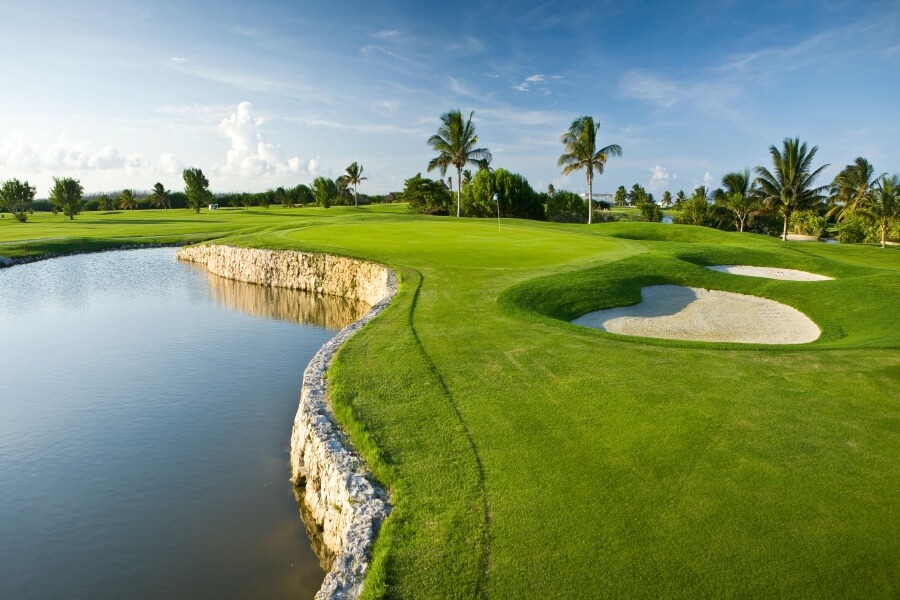}} 
      \subfloat[Back Light]{\includegraphics[width=2.8cm]{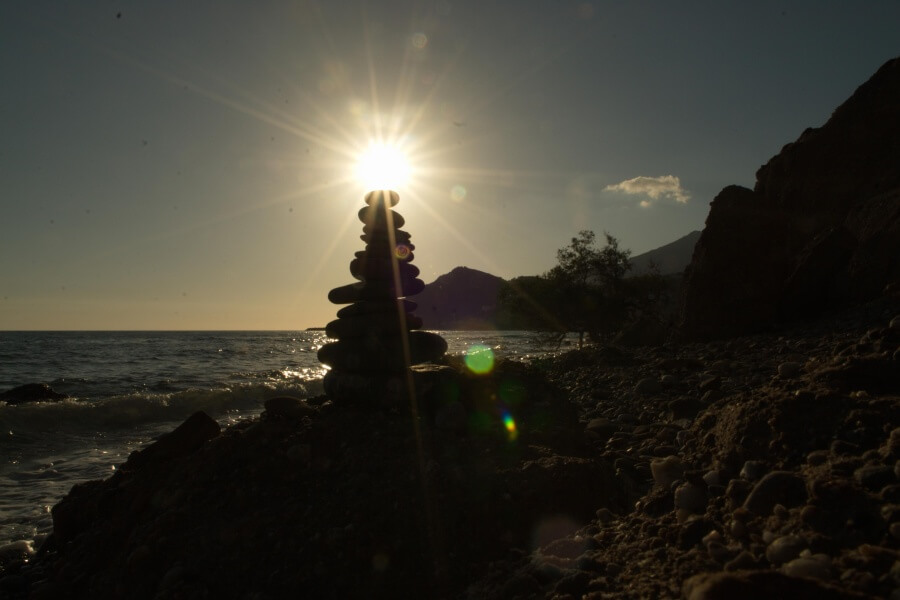}}
      \subfloat[Side Light]{ \includegraphics[width=2.8cm]{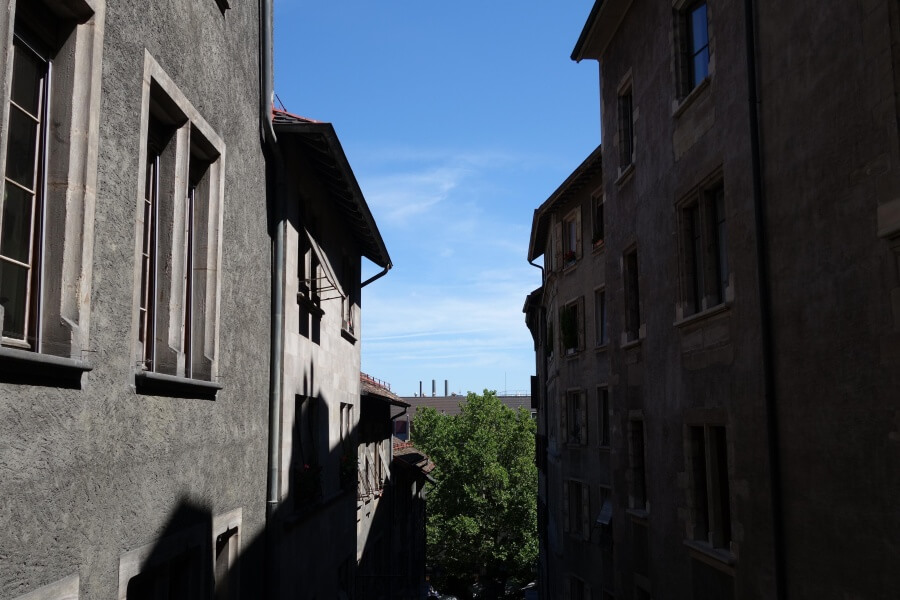}}
      \\
    \caption{\textbf{Real-world Images from the SICE~\cite{cai2018learning} Dataset.} These images exhibit diverse exposure and lighting, making visual aesthetics and scene understanding challenging. }
    \label{fig:header}
\end{figure}

Images are often captured under sub-optimal illumination conditions. Due to environmental factors (e.g., poor lightening, incorrect beam angle) or technical constraints (e.g., small ISO, short exposure)~\cite{li2021low}, these images could have deteriorated features, and low contrast (Shown in Fig. \ref{fig:header}), which not only deteriorate the low-level perceptual quality but also degrade the high-level vision tasks such as object detection~\cite{loh2019getting}, semantic segmentation~\cite{wang2022sfnet}, and depth estimation~\cite{lamba2020harnessing}.


\begin{figure*}
    \centering
    \includegraphics[width=18cm]{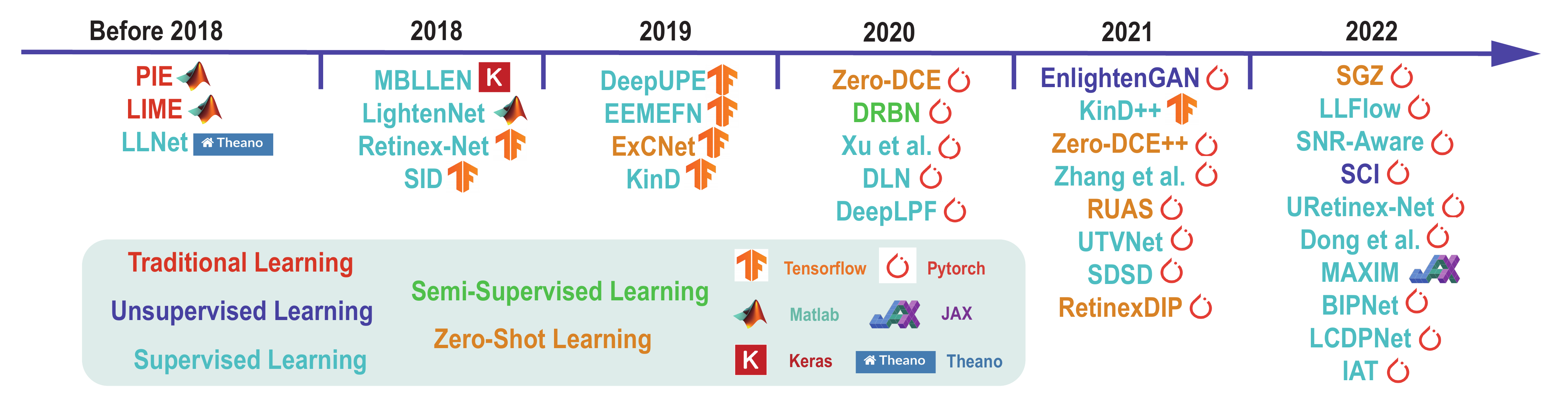}
    \caption{\textbf{A Milestone for Recent Representative Low-Light Image and Video Enhancement Methods.} (1) Traditional Learning methods: PIE~\cite{fu2015probabilistic} and LIME~\cite{guo2016lime}. (2) Unsupervised Learning methods: EnlightenGAN~\cite{jiang2021enlightengan} and SCI~\cite{ma2022toward}. (3) Semi-Supervised Learning method: DRBN~\cite{yang2020fidelity}. (4): Zero-Shot Learning methods: ExCNet~\cite{zhang2019zero}, Zero-DCE~\cite{guo2020zero}, Zero-DCE++~\cite{li2021learning}, RUAS~\cite{liu2021retinex}, RetinexDIP~\cite{zhao2021retinexdip}, and SGZ~\cite{zheng2022semantic}. (5) Supervised Learning methods: LLNet~\cite{lore2017llnet}, MBLLEN~\cite{lv2018mbllen}, LightenNet~\cite{li2018lightennet}, Retinex-Net~\cite{pham2020low}, SID~\cite{cheng2016learning}, DeepUPE~\cite{wang2019underexposed}, EEMEFN~\cite{zhu2020eemefn}, KinD~\cite{zhang2019kindling}, Xu et al.~\cite{xu2020learning}, DLN~\cite{wang2020lightening}, DeepLPF~\cite{moran2020deeplpf}, KinD++~\cite{zhang2021beyond}, Zhang et al.~\cite{zhang2021learning}, UTVNett~\cite{zheng2021adaptive}, SDSD~\cite{wang2021seeing}, LLFlow~\cite{wang2022low}, SNR-Aware~\cite{xu2022snr}, URetinex-Net~\cite{wu2022uretinex}, Dong et al.~\cite{dong2022abandoning}, MAXIM~\cite{tu2022maxim}, BIPNet~\cite{dudhane2022burst}, LCDPNet~\cite{wang2022local}, and IAT~\cite{cui2022illumination}.}
    \label{fig:Timeline}
\end{figure*}



The aforesaid problem can be addressed in a logical manner from the camera side. The brightness of the images will undoubtedly improve with higher ISO and exposure. However, boosting ISO causes noise, whereas prolonged exposure produces motion blur~\cite{cheng2016learning}, which makes the images look even worse. The other viable option is to enhance the visual appeal of low-light images using image editing tools like Photoshop or Lightroom. But both tools demand artistic taste and take a long time on large datasets.

Contrasting with camera and software approaches that require manual efforts, Low-Light Image Enhancement (LLIE) is designed to autonomously enhance the visibility of images captured in low-light conditions. It is an active research field that is related to various system-level applications, such as visual surveillance~\cite{yang2019coarse}, autonomous driving~\cite{li2021deep}, unmanned aerial vehicle~\cite{samanta2018log}, photography~\cite{yuan2007image}, remote sensing~\cite{campbell2011introduction}, microscopic imaging~\cite{pepperkok2006high}, and underwater imaging~\cite{li2019underwater}. 

In pre-deep learning eras, the only option for LLIE is the traditional approaches. Most traditional LLIE methods utilize Histogram Equalization~\cite{pizer1987adaptive,pizer1990contrast,arici2009histogram,lee2013contrast,nakai2013color,ying2017new,wu2017contrast}, Retinex theory~\cite{land1971lightness,land1977retinex,lee2013adaptive,wang2013naturalness,wang2014variational,fu2016weighted,cai2017joint}, Dehazing 
\cite{dong2010fast,li2015low,chiang2011underwater,li2016underwater}, or Statistical methods ~\cite{celik2011contextual,liang2015contrast,yu2017low,ying2017new,su2017low}. While these traditional approaches have solid theoretical foundations, in practice they deliver unsatisfactory results.



    
The popularity of deep learning LLIE approaches can be attributed to their superior effectiveness, efficiency, and generalizability. Deep learning-based LLIE methods can be divided into the following categories: supervised learning~\cite{lore2017llnet,lv2018mbllen,li2018lightennet,wei2018deep,cheng2016learning,wang2019underexposed,zhu2020eemefn,zhang2019kindling,xu2020learning,wang2020lightening,moran2020deeplpf,zhang2021beyond,zhang2021learning,zheng2021adaptive,wang2021seeing,wang2022low,xu2022snr,wu2022uretinex,dong2022abandoning,tu2022maxim,dudhane2022burst,wang2022local,cui2022illumination}, unsupervised learning~\cite{jiang2021enlightengan,ma2022toward}, semi-supervised learning~\cite{yang2020fidelity} and zero-shot learning~\cite{zhang2019zero,guo2020zero,li2021learning,liu2021retinex,zhao2021retinexdip,zheng2022semantic} methods. In the past five years, there have been a handful of publications on deep learning-based LLIE (See Fig. \ref{fig:Timeline} and Tab. \ref{tab:summary}). Each of these learning algorithms exhibits its own set of strengths and limitations. For instance, unsupervised and zero-shot learning perform on unknown datasets, whereas supervised learning achieves state-of-the-art performance on benchmark datasets. It is crucial to carefully examine previous developments since they can offer a detailed knowledge, highlight present challenges, and suggest potential research directions for the LLIE community.

\begin{figure*}
    \centering
    \includegraphics[width=\textwidth]{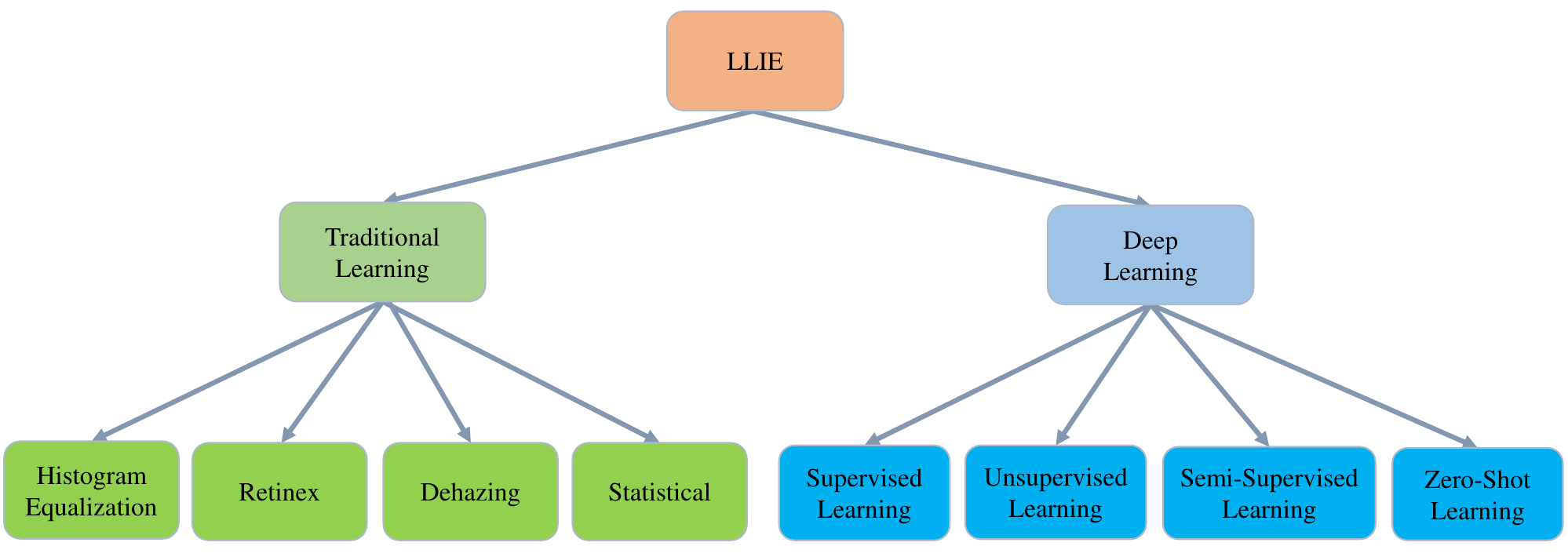}
    \caption{\textbf{A Hierarchical Taxonomy of Learning Strategies for Low-Light Image and Video Enhancement Methods.}}
    \label{fig:taxonomy}
\end{figure*}

\begin{table*}
\setlength\tabcolsep{0.1pt}
\tiny
\centering
\caption{Table Summary for recent representative low-light image and video enhancement methods. }
\begin{tabular}{l|c|c|c|c|c|c|l}
\toprule
\textbf{Name}                                               & \textbf{Publications} & \textbf{Network Structure} & \textbf{Loss Functions}                                                                                                                                                               & \textbf{Evaluation Metrics}                                                                                                         & \textbf{Training Dataset}                                                  & \textbf{Testing Dataset}                                                                                 & \textbf{One-line Summary}                                                                                      \\ \hline
PIE~\cite{fu2015probabilistic}            & TIP                   & -                        & -                                                                                                                                                                                   & \begin{tabular}[c]{@{}c@{}}Contrast Gain, LOE, \\ NIQE\end{tabular}                                                                 & -                                                                        & Custom Dataset                                                                                           & Probabilistic method for image enhancement via illumination \& reflectance estimation.                         \\ \hline
LIME~\cite{guo2016lime}                   & TIP                   & -                        & -                                                                                                                                                                                   & LOE                                                                                                                                 & -                                                                        &  LIME    &   LLIE using illumination map.                                                                        \\ \hline
LLNet~\cite{lore2017llnet}                 & PR                    & U-Net                      & L2, KL                                                                                                                                                                                & PSNR, SSIM                                                                                                                          & NORB                                                                       & CVG-UGR                                                                                                  & Deep autoencoder for adaptive high dynamic range image brightening.                                           \\ \hline
MBLLEN~\cite{lv2018mbllen}                & BMVC                  & Multi-scale                & SSIM, Region, Perceptual                                                                                                                                                              & \begin{tabular}[c]{@{}c@{}}PSNR, SSIM, VIF, \\ LOE, TMQI, AB\end{tabular}                                                           & Custom Dataset                                                             & Custom Dataset                                                                                           & Multi-branch fusion network for LLIE/LLVE.                                                                   \\ \hline
LightenNet~\cite{li2018lightennet}        & PRL                   & Others                     & L2                                                                                                                                                                                    & \begin{tabular}[c]{@{}c@{}}PSNR, SSIM, MSE, \\ US\end{tabular}                                                                      & Custom Dataset                                                             & Custom Dataset                                                                                           & Retinex-based CNN for enhancing weakly illuminated images.                                                \\ \hline
Retinex-Net~\cite{wei2018deep}            & BMVC                  & Multi-scale                & \begin{tabular}[c]{@{}c@{}}Reconstruction, L1, \\ Illumination Smoothness\end{tabular}                                                                                                & NIQE                                                                                                                                & LOL                                                                        & \begin{tabular}[c]{@{}c@{}}MEF, NPE, LIME, \\ DICM, VV, Fusion, \\ LOL, RAISE\end{tabular}               & Retinex network for LLIE via decomposition \& illumination adjustment.                       \\ \hline
SID~\cite{cheng2016learning}              & CVPR                  & U-Net, Multi-scale         & L1                                                                                                                                                                                    & PSNR, SSIM                                                                                                                          & SID                                                                        & SID                                                                                                      & End-to-end trained FCN for low-light image processing.                                 \\ \hline
DeepUPE~\cite{wang2019underexposed}       & CVPR                  & Others                     & \begin{tabular}[c]{@{}c@{}}Reconstruction, Color, \\ Smoothness\end{tabular}                                                                                                          & PSNR, SSIM, US                                                                                                                      & Custom Dataset                                                             & MIT, Custom Dataset                                                                                      & Enhancement of underexposed images using constrained intermediate illumination.                 \\ \hline
EEMEFN~\cite{zhu2020eemefn}               & AAAI                  & U-Net                      & L1, Edge                                                                                                                                                                              & PSNR, SSIM                                                                                                                          & SID                                                                        & SID                                                                                                      & Edge-Enhanced Multi-Exposure Fusion Network for enhancing extreme low-light images.                                      \\ \hline
ExCNet~\cite{zhang2019zero}               & ACMMM                 & Others                     & Energy Minimization                                                                                                                                                                   & CDIQA, LOD, US                                                                                                                      & Custom Dataset                                                             & IEpsD                                                                                                    & Zero-shot CNN for back-lit image restoration via `S-curve' estimation.                                         \\ \hline
KinD~\cite{zhang2019kindling}             & ACMMM                 & U-Net                      & \begin{tabular}[c]{@{}c@{}}Reflectance Similarity, \\ Illumination Smoothness, \\ Mutual Consistency, \\ L1, L2, SSIM, \\ Texture Similarity, \\ Illumination Adjustment\end{tabular} & \begin{tabular}[c]{@{}c@{}}PSNR, SSIM, NIQE, \\ LOE\end{tabular}                                                                    & LOL                                                                        & \begin{tabular}[c]{@{}c@{}}LOL, LIME, NPE, \\ MEF\end{tabular}                                           & Retinex network with light adjustment \& degradation removal for LLIE.                        \\ \hline
Zero-DCE~\cite{guo2020zero}               & CVPR                  & Others                     & \begin{tabular}[c]{@{}c@{}}Spatial Consistency, \\ Exposure Control, \\ Color Constancy, \\ Illumination Smoothness\end{tabular}                                                      & \begin{tabular}[c]{@{}c@{}}PSNR, SSIM, MAE, \\ PI, US, Precision, \\ Recall, Time\end{tabular}                                      & SICE                                                                       & \begin{tabular}[c]{@{}c@{}}NPE, LIME, MEF, \\ DICM, VV, SICE, \\ Dark Face\end{tabular}                  & Zero-shot network for LLIE using high-order curves \& dynamic range adjustment.               \\ \hline
DRBN~\cite{yang2020fidelity}              & CVPR                  & U-Net, Recursive           & Perceptual, Detail, Quality                                                                                                                                                           & \begin{tabular}[c]{@{}c@{}}PSNR, SSIM, \\ SSIM-GC\end{tabular}                                                                      & LOL                                                                        & LOL                                                                                                      & Semi-supervised deep recursive band network using band decomposition for LLIE.                \\ \hline
Xu et al.~\cite{xu2020learning}           & CVPR                  & U-Net, Multi-scale         & L2, Perceptual                                                                                                                                                                        & PSNR, SSIM                                                                                                                          & Custom Dataset                                                             & SID, Custom Dataset                                                                                      & Frequency-based decomposition model for LLIE.                                           \\ \hline
DLN~\cite{wang2020lightening}             & TIP                   & Others                     & SSIM, TV                                                                                                                                                                              & \begin{tabular}[c]{@{}c@{}}PSNR, SSIM, NIQE, \\ US, \#Params, Time, \\ Size\end{tabular}                                            & Custom Dataset                                                             & Custom Dataset                                                                                           & Deep Lightening Network with Lightening Back-Projection blocks for LLIE.                                                \\ \hline
DeepLPF~\cite{moran2020deeplpf}            & CVPR                  & U-Net                      & L1, SSIM                                                                                                                                                                              & \begin{tabular}[c]{@{}c@{}}PSNR, SSIM, LPIPS, \\ \#Params\end{tabular}                                                              & MIT, SID                                                                   & MIT, SID                                                                                                 & Deep Local Parametric Filters model with spatially localized filters for LLIE.                    \\ \hline
EnlightenGAN~\cite{jiang2021enlightengan} & TIP                   & U-Net, Multi-scale         & \begin{tabular}[c]{@{}c@{}}Adversarial, \\ Self-Feature Preserving\end{tabular}                                                                                                       & NIQE, US, Accuracy                                                                                                                  & Custom Dataset                                                             & \begin{tabular}[c]{@{}c@{}}MEF, LIME, NPE, \\ VV, DICM, ExDARK, \\ BDD100k\end{tabular}                  & Unsupervised GAN with global-local discriminator \& attention for LLIE.                       \\ \hline
KinD++~\cite{zhang2021beyond}            & IJCV                  & U-Net                      & \begin{tabular}[c]{@{}c@{}}Reflectance Similarity, \\ Illumination Smoothness, \\ Mutual Consistency, \\ L1, L2, SSIM, \\ Texture Similarity, \\ Illumination Adjustment\end{tabular} & \begin{tabular}[c]{@{}c@{}}PSNR, SSIM, NIQE, \\ LOE, DeltaE, US,\\ Bradley-Terry\end{tabular}                                       & LOL                                                                        & \begin{tabular}[c]{@{}c@{}}LOL, DICM, LIME, \\ NPE, MEF, SICE\end{tabular}                               & KinD extension.                                                                                                \\ \hline
Zero-DCE++~\cite{li2021learning}           & TPAMI                 & Others                     & \begin{tabular}[c]{@{}c@{}}Spatial Consistency, \\ Exposure Control, \\ Color Constancy, \\ Illumination Smoothness\end{tabular}                                                      & \begin{tabular}[c]{@{}c@{}}PSNR, SSIM, MAE, \\ PI, US, Precision, \\ Recall, Time, \#Params, \\ FLOPs\end{tabular}                  & SICE                                                                       & SICE                                                                                                     & Zero-DCE extension.                                                                                            \\ \hline
Zhang et al.~\cite{zhang2021learning}     & CVPR                  & U-Net, Optical Flow        & L1, Consistency                                                                                                                                                                       & \begin{tabular}[c]{@{}c@{}}PSNR, SSIM, AB, \\ MABD, WE, US\end{tabular}                                      & Custom Dataset                                                             & Custom Dataset                                                                                           & Optical flow modelfor temporal stability in LLVE.                                     \\ \hline
RUAS~\cite{liu2021retinex}                & CVPR                  & NAS, Unfolding             & \begin{tabular}[c]{@{}c@{}}Cooperative, \\ Similarity, TV\end{tabular}                                                                                                                & \begin{tabular}[c]{@{}c@{}}PSNR, SSIM, Time, \\ \#Params, FLOPs\end{tabular}                                                        & MIT, LOL                                                                   & \begin{tabular}[c]{@{}c@{}}MIT, LOL, DarkFace, \\ ExtremelyDarkFace\end{tabular}                         & Retinex-inspired model with Neural Architecture Search for LLIE.                                   \\ \hline
UTVNet~\cite{zheng2021adaptive}           & ICCV                  & U-Net, Unfolding           & TV                                                                                                                                                                                    & \begin{tabular}[c]{@{}c@{}}PSNR, SSIM, L2-Lab, \\ LPIPS\end{tabular}                                                                & sRGB-SID                                                                   & sRGB-SID                                                                                                 & Adaptive unfolding TV network with noise level approximation for LLIE.                             \\ \hline
SDSD~\cite{wang2021seeing}                & ICCV                  & Others                     & \begin{tabular}[c]{@{}c@{}}Progressive Alignment, \\ Self-Supervised Noise Estimation, \\ Illumination Map Prediction\end{tabular}                                                    & PSNR, SSIM, US                                                                                                                      & SDSD, SMID                                                                 & SDSD, SMID                                                                                               & Retinex method with self-supervised noise reduction for LLVE.                            \\ \hline
RetinexDIP~\cite{zhao2021retinexdip}       & TCSVT                 & U-Net                      & \begin{tabular}[c]{@{}c@{}}Illumination Smoothness, \\ Illumination Consistency, \\ Reflectance, \\ Reconstruction, \\ SSIM, TV\end{tabular}                                          & NIQE, NIQMC, CPCQI                                                                                                                  & -                                                                        & \begin{tabular}[c]{@{}c@{}}DICM, ExDark, Fusion, \\ LIME, NASA, NPE, \\ VV\end{tabular}                  & Retinex zero-shot method using 'generative' decomposition for LLIE.                              \\ \hline
SGZ~\cite{zheng2022semantic}              & WACV                  & U-Net, Recurrent           & \begin{tabular}[c]{@{}c@{}}Spatial Consistency, \\ RGB, Brightness, \\ TV, Semantic\end{tabular}                                                                                      & \begin{tabular}[c]{@{}c@{}}PSNR, SSIM, MSE, \\ UNIQUE, BRISQUE, Time, \\ \#Params, FLOPs, US, \\ mIoU, mPA\end{tabular}             & SICE                                                                       & \begin{tabular}[c]{@{}c@{}}NPE, LIME, MEF, \\ DICM, VV, LOL, \\ DarkBDD, DCS\end{tabular}                & Zero-shot LLIE/LLVE network via light deficiency estimation \& semantic segmentation. \\ \hline
LLFlow~\cite{wang2022low}                & AAAI                  & Normalizing Flow           & NLL                                                                                                                                                                                   & PSNR, SSIM, LPIPS                                                                                                                   & LOL, VE-LOL                                                                & LOL, VE-LOL                                                                                              & Normalizing flow-based model conditioned on low-light images/features.                                    \\ \hline
SNR-Aware~\cite{xu2022snr}                & CVPR                  & Transformer                & \begin{tabular}[c]{@{}c@{}}Charbonnier,\\ Perceptual\end{tabular}                                                                                                                     & PSNR, SSIM, US                                                                                                                      & \begin{tabular}[c]{@{}c@{}}LOL, SID, \\ SMID, SDSD\end{tabular}            & \begin{tabular}[c]{@{}c@{}}LOL, SID, SMID, \\ SDSD\end{tabular}                                          & Signal-to-Noise-Ratio aware transformer for LLIE.                                                                      \\ \hline
SCI~\cite{ma2022toward}                   & CVPR                  & Others                     & Fidelity, Smoothness                                                                                                                                                                  & \begin{tabular}[c]{@{}c@{}}PSNR, SSIM, EME, \\ DE, LOE, NIQE, \\ Size, FLOPs, Time, \\ Precision, Recall, mAP, \\ mIoU\end{tabular} & \begin{tabular}[c]{@{}c@{}}MIT, LSRW, \\ ACDC, DarkFace\end{tabular}       & \begin{tabular}[c]{@{}c@{}}MIT, LSRW, \\ Dark Face, \\ UG2+Prize Challenge, \\ ACDC, ExDark\end{tabular} & Self-Calibrated Illumination learning for LLIE.                                              \\ \hline
URetinex-Net~\cite{wu2022uretinex}         & CVPR                  & Unfolding                  & \begin{tabular}[c]{@{}c@{}}Initialization, \\ Unfolding Optimization, \\ Illumination Adjustment\end{tabular}                                                                         & \begin{tabular}[c]{@{}c@{}}MAE, PSNR, SSIM, \\ LPIPS, Time\end{tabular}                                                             & LOL                                                                        & LOL, SICE, MEF                                                                                           & Retinex-based Deep Unfolding Network with unfolding optimization for LLIE.                         \\ \hline
Dong et al.~\cite{dong2022abandoning}    & CVPR                  & U-Net                      & L1                                                                                                                                                                                    & PSNR, SSIM                                                                                                                          & MCR, SID                                                                   & MCR, SID                                                                                                 & Fusion of colored \& synthesized monochrome raw image for LLIE.                                  \\ \hline
MAXIM~\cite{tu2022maxim}                  & CVPR                  & Transformer, Multi-scale   & \begin{tabular}[c]{@{}c@{}}Charbonnier, \\ Frequency Reconstruction\end{tabular}                                                                                                      & PSNR, \#Params, FLOPs                                                                                                               & MIT, LOL                                                                   & MIT, LOL                                                                                                 & Multi-axis MLP transformer for low-level image processing  tasks.                                             \\ \hline
BIPNet~\cite{dudhane2022burst}            & CVPR                  & U-Net, Multi-scale         & L1                                                                                                                                                                                    & PSNR, SSIM, LPIPS                                                                                                                   & SID                                                                        & SID                                                                                                      & Burst image enhancement using pseudo-burst features.                                               \\ \hline
LCDPNet~\cite{wang2022local}              & ECCV                  & U-Net, Multi-scale         & \begin{tabular}[c]{@{}c@{}}L2, TV,\\ Cosine Similarity\end{tabular}                                                                                                                   & PSNR, SSIM                                                                                                                          & Custom Dataset                                                             & MSEC, Custom Dataset                                                                                     & Over-/under-exposed region localization and enhancement via local color distribution.               \\ \hline
IAT~\cite{cui2022illumination}             & BMVC                  & Transformer                & L1                                                                                                                                                                                    & \begin{tabular}[c]{@{}c@{}}PSNR, SSIM, PI, \\ mAP, Time, mIoU, \\ FLOPs, \#Params\end{tabular}                                      & \begin{tabular}[c]{@{}c@{}}LOL, MIT, \\ ExDark, ACDC, \\ TYOL\end{tabular} & \begin{tabular}[c]{@{}c@{}}LOL, MIT, ExDark, \\ ACDC, TYOL\end{tabular}                                  & Illumination Adaptive Transformer with ISP parameter adjustment.                               \\ \bottomrule
\end{tabular}
\label{tab:summary}
\end{table*}

Three recent surveys have been conducted on LLIE. 
Wang et al.~\cite{wang2020experiment} provide an overview of traditional learning-based LLIE techniques. 
Liu et al.~\cite{liu2021benchmarking} propose a new LLIE dataset named VE-LOL, review LLIE methods, and introduce a joint image enhancement and face detection network named EDTNet. 
Li et al.~\cite{li2021low} unveil a new LLIE dataset named LLIV-Phone, reviews deep learning-based LLIE methods, and design an online demo platform for LLIE methods. 


The existing surveys have the following limitations. 

\begin{itemize}
    \item Their proposed dataset has either overexposure \textit{or} underexposure for single images. This assumption contradicts images from the real world, which frequently contain both overexposure \textit{and} underexposure.
    \item Their proposed dataset contains few videos, and even these videos are filmed in fixed shooting positions. This oversimplification is also inconsistent with real-world videos that are often captured in motion.
    \item These studies emphasize low-level perceptual quality and high-level vision tasks while neglecting system-level application, which is essential when LLIE approaches are implemented in real-world products. 
\end{itemize}

This paper makes the following contributions to existing LLIE surveys:

\begin{itemize}
    \item We present the most recent comprehensive survey on low-light image and video enhancement. In particular, we conducted an extensive qualitative and quantitative comparison with various full-reference and non-reference evaluation metrics and made a modularized discussion focusing on structures and strategies.  Based on these analysis, we identity the emerging system-level applications, point out the open challenges and suggest directions for future works.
    \item We introduce two image datasets named SICE\_Grad and SICE\_Mix. They are the first datasets that include both overexposure and underexposure in single images. This preliminary effort highlights the LLIE community's unresolved mixed over- and underexposure challenge.
    \item We propose Night Wenzhou, a large-scale high-resolution video dataset. Night Wenzhou is captured during fast motions and contains diverse illuminations, various landscapes, and miscellaneous degradation. It will facilitate the application of LLIE methods to real-world challenges like autonomous driving and UAV. 
\end{itemize}


\begin{figure}[t]
    \centering
    \includegraphics[width=9cm]{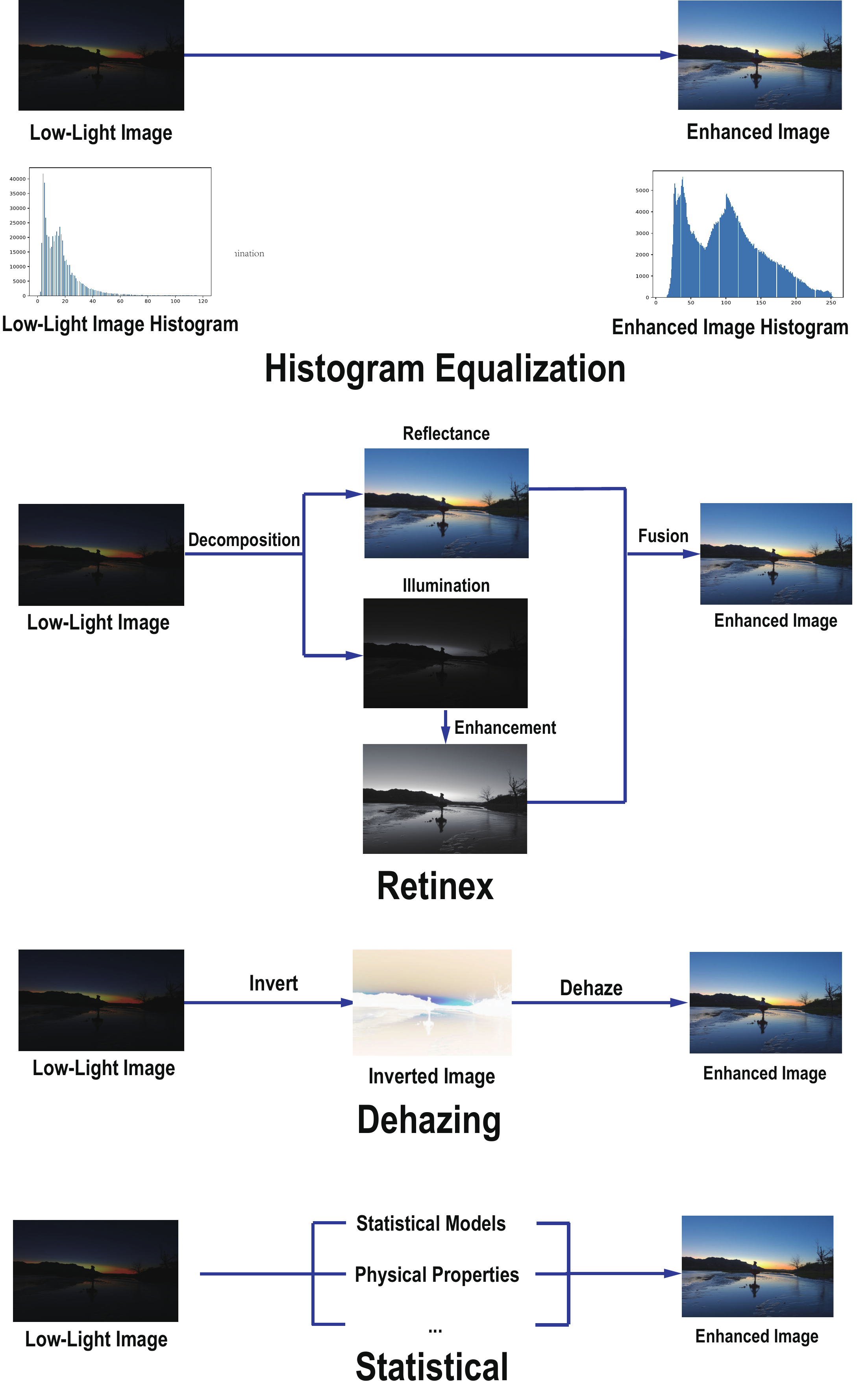}
    \caption{\textbf{Learning Strategies for Traditional Learning-based Low-Light Image and Video Enhancement.} See Section \ref{learning_strategies} for details.  }
    \label{fig:ls}
\end{figure}

\begin{figure}[t]
    \centering
    \includegraphics[width=9cm]{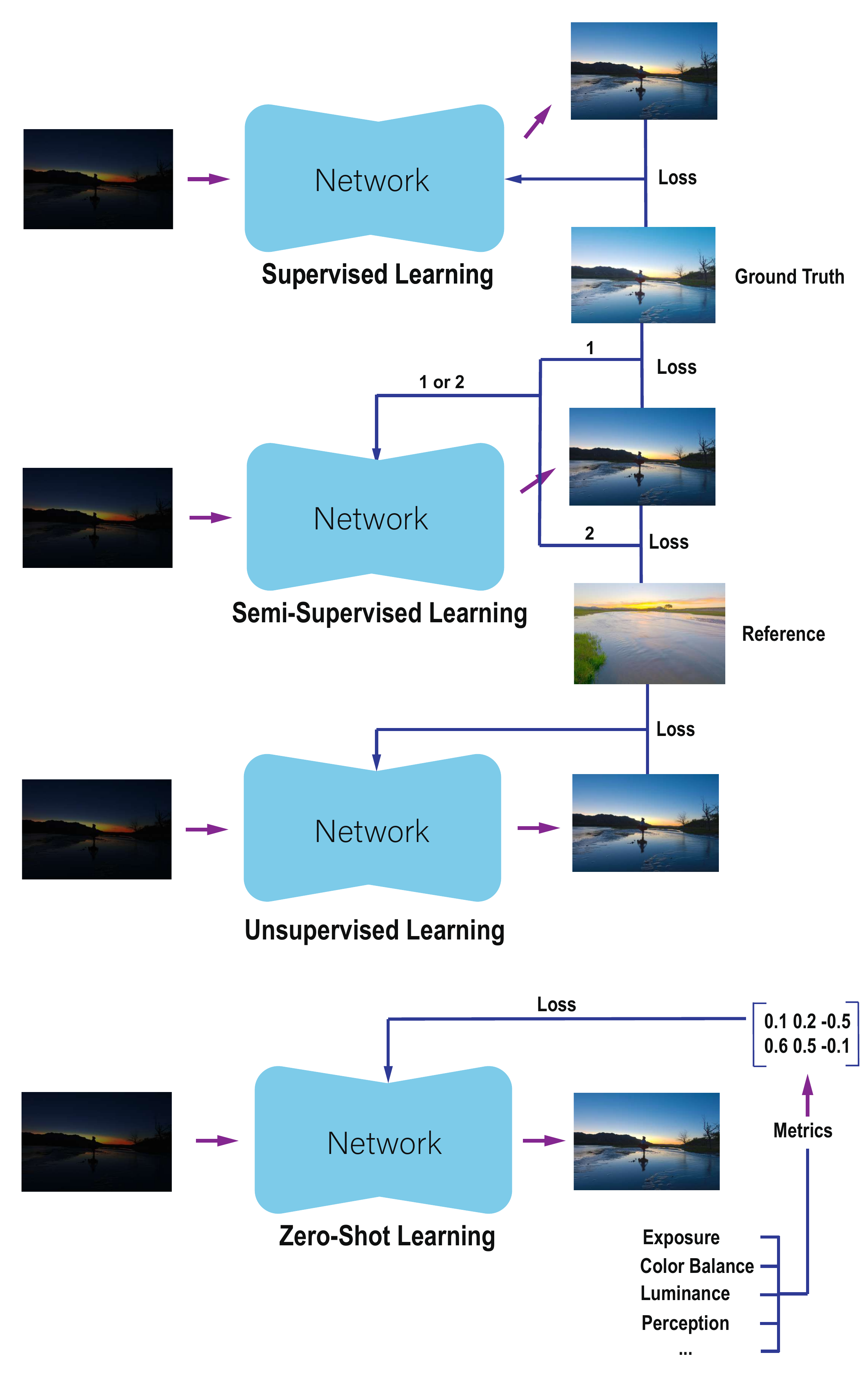}
    \caption{\textbf{Learning Strategies for Deep Learning-based Low-Light Image and Video Enhancement.}  See Section \ref{learning_strategies} for details. }
    \label{fig:learning_strategy}
\end{figure}

\begin{figure*}[t]
    \centering
    \subfloat[Learning Strategy]{\includegraphics[width=0.33\linewidth]{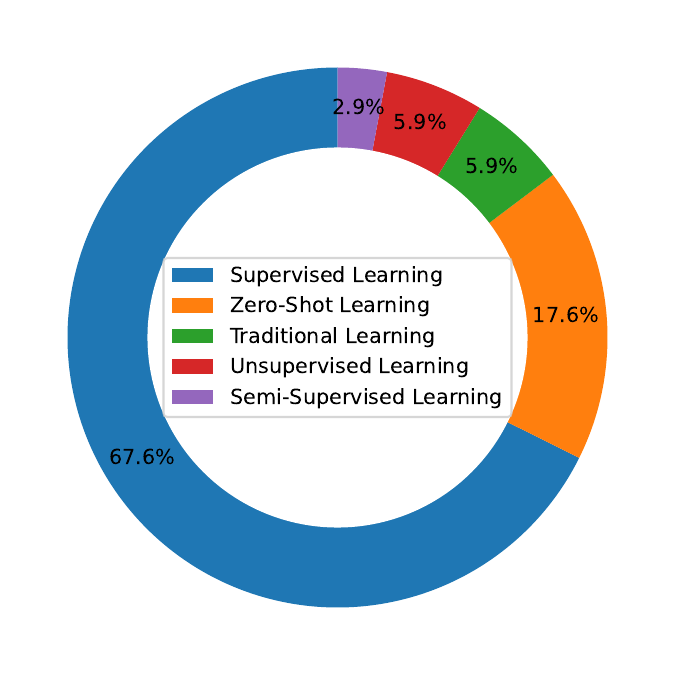}}
    \subfloat[Network Structure]{\includegraphics[width=0.33\linewidth]{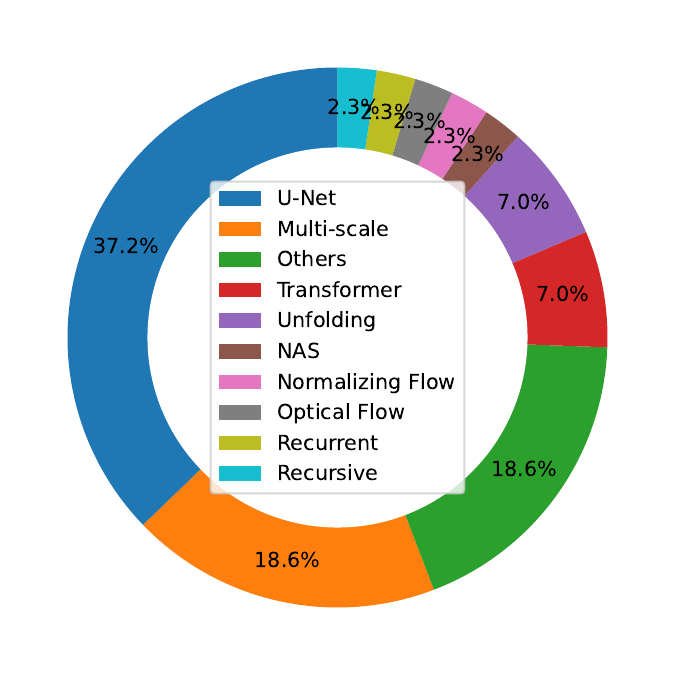}}
    \subfloat[Loss Functions]{\includegraphics[width=0.33\linewidth]{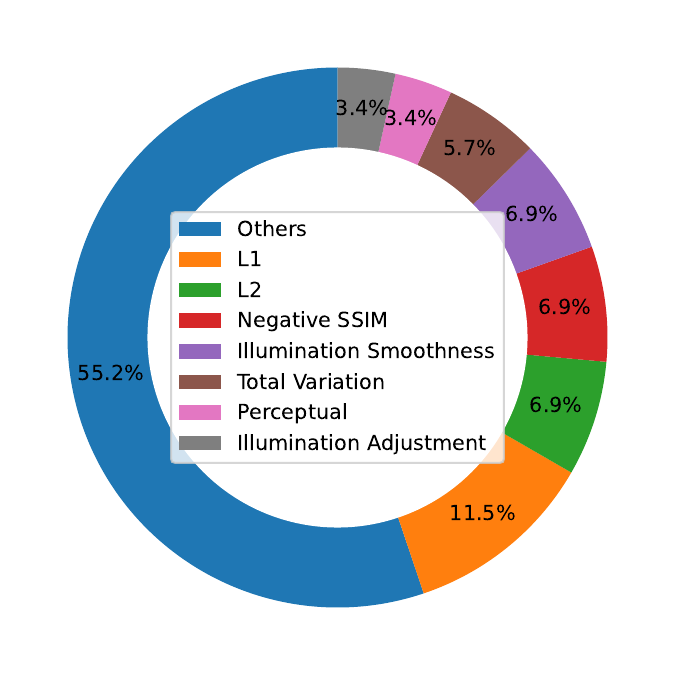}}  \\
    \subfloat[Evaluation Metrics]{\includegraphics[width=0.33\linewidth]{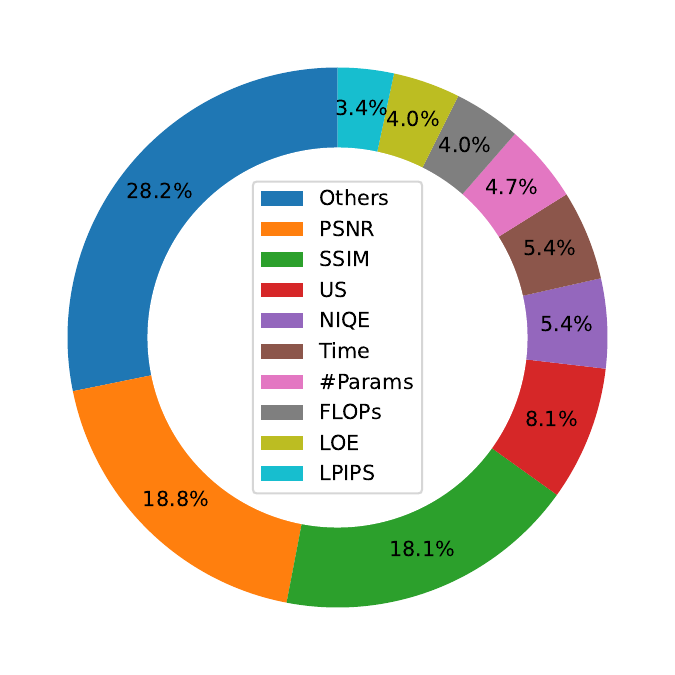} }
    \subfloat[Training Datasets]{\includegraphics[width=0.33\linewidth]{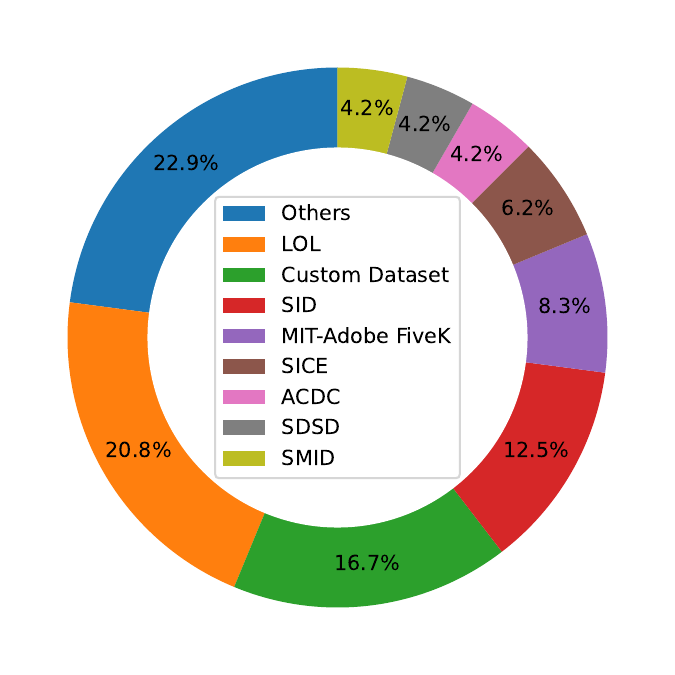}}
    \subfloat[Testing Datasets]{\includegraphics[width=0.33\linewidth]{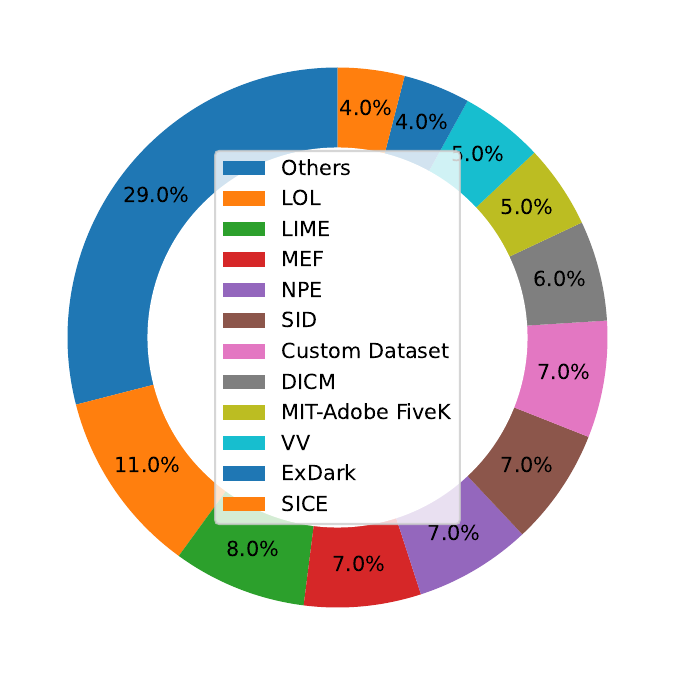}}
     \\
     
    
    \caption{\textbf{Donut Charts Summarizing Characteristics for Low-Light Image and Video Enhancement Methods.} See Section \ref{sec:methods_review} for details. }
    \label{fig:methods_donut}
\end{figure*}

The rest of the paper is organized as follows. Section \ref{sec:methods_review} provides a systematic overview of existing LLIE methods (Fig. \ref{fig:taxonomy} and \ref{fig:methods_donut}). Section \ref{sec:datasets} introduces the benchmark datasets and the proposed datasets. Section \ref{sec:evaluations} makes empirical analysis and comparisons for representative LLIE methods. Section \ref{sec:applications} identifies the system-level applications. Section \ref{sec:future_prospects} discusses the open challenges and the corresponding future works. Section \ref{sec:conclusion} provides the concluding remarks.


    \section{Methods Review} \label{sec:methods_review}

\subsection{Selection Criteria}


We select the LLIE methods according to the following rubrics. 

\begin{itemize}
    \item We focus on LLIE methods in the recent 5 years (2018-2022) and lay emphasis on deep learning-based LLIE methods in the recent 2 years (2021-2022) because of their rapid development.
    \item We pick LLIE methods published in prestigious conferences (e.g., CVPR) and journals (e.g., TIP) with official codes to ensure credibility and authenticity.
    \item For the paper published in the same year, we prefer works with more citations and Github stars.
    \item We include LLIE methods that significantly surpass previous state-of-the-art benchmark LLIE datasets. 
\end{itemize}

\subsection{Learning Strategies} \label{learning_strategies}

Before 2017, traditional learning-based (\textbf{TL}) methods are the ad-hoc solution for LLIE. As shown in Fig. \ref{fig:ls}, mainstream traditional learning methods in LLIE utilize Histogram Equalization, Retinex, Dehazing or Statistical techniques. Since 2017, deep learning-based (\textbf{DL}) methods have started to dominate this field. Supervised learning (67.6 \%) is so far the most popular strategy. From 2019, there are several methods using zero-shot learning (17.6 \%), unsupervised learning (5.9 \%), and semi-supervised (2.9 \%), as shown in Fig. \ref{fig:learning_strategy}.

    

     \noindent \textbf{(TL) Histogram Equalization:}
    Histogram Equalization (HE)-based methods spread out the frequent intensity values of an image to improve its global contrast. In this way, the low-contrast region of an image gains higher contrast, and the visibility improves. The original HE-based method~\cite{pizer1987adaptive} consider global adjustment only, leading to poor local illumination and amplified degradation (e.g., noise, blur, and artifacts). The follow-up works attempt to address these issues using different priors and constraints. For example, Pizer et al.~\cite{pizer1990contrast} perform HE on partitioned regions with local contrast constraints to suppress noise. Ibrahim et al.~\cite{ibrahim2007brightness} utilize mean brightness preservation to prevent visual deterioration, whereas Arici et al.~\cite{arici2009histogram} additionally integrate contrast adjustment, noise robustness, and white/black stretching. The later works integrate gray-level differences~\cite{lee2013contrast}, depth information~\cite{nakai2013color}, weighting matrix~\cite{ying2017new}, and visual importance~\cite{wu2017contrast} to guide low-light image enhancement towards better fine-grained details. On the flip side, the HE-based methods' performance gains are predominantly attributed to manually-designed constraints upon the vanilla histogram equalization.

    \noindent \textbf{(TL) Retinex:}
    Retinex-based methods are based on the Retinex theory of color vision~\cite{land1971lightness,land1977retinex}, which assumes that an image can be decomposed into a reflectance map and an illumination map. The enhanced image can be obtained by fusing the enhanced illumination map and the reflectance map. Lee et al.~\cite{lee2013adaptive} is the pioneering work incorporating Retinex theory into image enhancement. After that, Wang et al.~\cite{wang2013naturalness} utilizes lightness-order-error and bi-log transform to improve the naturalness and details during enhancement. Another work by Wang et al.~\cite{wang2014variational} leverages Gibbs distributions as priors for the reflectance and illumination and gamma distributions as priors for the network parameters. Fu et al.~\cite{fu2016weighted} design a weighted variational model (instead of logarithmic transform) for better prior modeling and edge preservation. Guo et al.~\cite{guo2016lime} introduce a structure prior to refining the initial illumination map. Cai et al.~\cite{cai2017joint} propose a shape prior for structural information preservation, a texture prior for reflection estimation, and an illumination prior for luminous modeling, respectively. Still, the Retinex-based models extensively rely on hand-crafted priors to achieve satisfactory image enhancement. 

    \noindent \textbf{(TL) Dehazing:}
    Dehazing-based methods treat the inverted low-light images as haze images and then apply dehazing algorithms to enhance the image~\cite{dong2010fast}. Instead of attending the whole image, Li et al.~\cite{li2015low} decompose the images into base layers and details layers and use a dark channel prior~\cite{he2010single}-guided dehazing process for image enhancement. Although less prevalent in LLIE, the dehazing-based methods have been widely used in underwater image enhancement. For example, Chiang et al.~\cite{chiang2011underwater} leverage wavelength compensation, depth map estimation, and dehazing. The other work by Li et al.~\cite{li2016underwater} builds an underwater image enhancement pipeline using the minimum information loss principle, histogram prior, and dehazing. Nonetheless, the dehazing largely depends on the insubstantial dehazing assumptions for performing enhancement.


    \noindent \textbf{(TL) Non-HE Statistical:} 
    Statistical methods involve statistical models and physical properties for image processing and enhancement. Compared with other traditional learning-based methods, statistical methods require a solid mathematical foundation and expert domain knowledge. The pioneering statistical method by Celik et al.~\cite{celik2011contextual} performs contrast adjustment using 2-D interpixel contextual information. Similar to HE-based approaches, the follow-up works to improve the previous research using additional constraints and assumptions. For example, Liang et al.~\cite{liang2015contrast} propose a variational model based on discrete total variation for local contrast adjustment. Yu et al.~\cite{yu2017low} utilize Gaussian surrounding function for light estimation, followed by light-scattering attenuation with information loss constraint for light refinement. Ying et al.~\cite{ying2017new} utilizes a weighting matrix for illumination estimation and a camera response model for best exposure ratio finding. Su et al.~\cite{su2017low} exploit noise level function and a just noticeable difference model for noise suppression during image enhancement. Nevertheless, the Statistical methods are associated with computationally expensive optimization processes for performance improvements.

    \noindent \textbf{(TL) Hybrid Methods:} 
    Hybrid traditional learning methods aims to synergize the strengths of techniques like HE, Retinex, Dehazing and Statistical (Non-HE) for enhanced performances. HE-MSR-COM~\cite{han2023low} combines histogram equalization and multiscale Retinex to extract illumination from low-frequency HE-enhanced images and edge details from high-frequency Retinex-enhanced images for image enhancement. Li et al.~\cite{li2015underwater} presents an underwater image enhancement method that integrates dehazing, color correction, histogram equalization, saturation, intensity stretching, and bilateral filtering. Galdran et al.~\cite{galdran2018duality} explores the connection between Retinex and dehazing by applying Retinex to hazy images with inverted intensities. However, hybrid methods, in attempting to merge multiple techniques, can sometimes accumulate the weaknesses of each approach and may fail to deliver the expected synergistic benefits.

    \noindent \textbf{(DL) Supervised Learning:}
       In LLIE, supervised learning refers to the learning strategy with paired images. For example, supervised learning may use one dataset with 1,000 low-exposure images and another with 1,000 normal-exposure images that are different in only illuminations. The pioneering work LLNet~\cite{lore2017llnet} utilizes a stacked sparse denoising autoencoder to exploit the multi-scale information for image enhancement. The subsequent work MBLLEN~\cite{lv2018mbllen}, for the first time, applies low-light image enhancement techniques to videos. Later, KinD~\cite{zhang2019kindling} and KinD++~\cite{zhang2021beyond} combine model-based Retinex theory with data-driven image enhancement to cope with light adjustment and degradation removal. After that, Zhang et al.~\cite{zhang2021learning} exploit optical flow to promote stability in low-light video enhancement, whereas LLFlow leverages normalizing flow for illumination adjustment and noise suppression. The recent work IAT~\cite{cui2022illumination} proposes a lightweight illumination adaptive transformer for exposure correction and image enhancement. It is worth mentioning that the supervised learning method has achieved state-of-art-results on benchmark datasets.


    \noindent \textbf{(DL) Unsupervised Learning:}
        In LLIE, unsupervised learning refers to the learning strategy without paired images. For instance, unsupervised learning may use a dataset with low-exposure images and another with normal-exposure images that are different in more than illuminations. There are two unsupervised learning existing in LLIE literature. EnlightenGAN~\cite{jiang2021enlightengan} is an unsupervised generative adversarial network (GAN) that regularizes unpaired learning using a multi-scale discriminator,  a self-regularized perceptual loss, and the attention mechanism. SCI~\cite{ma2022toward} introduces a self-calibrated illumination framework that utilizes a cascaded illumination learning process with weight sharing. Compared with supervised learning methods, unsupervised learning approaches avoids the tedious work of collecting paired training images.

    
   \noindent \textbf{(DL) Semi-supervised learning:}
        In LLIE, semi-supervised learning is a learning strategy with a small quantity of paired images and many unpaired images. An example will be a dataset with low-exposure images and another with normal-exposure images where most images are different in more than illuminations, and few images are different in only illuminations. To our knowledge, semi-supervised learning has been used by only one representative LLIE method named DRBN~\cite{yang2020fidelity}, which is based upon a recursive neural network with band decomposition and recomposition. Compared with supervised and unsupervised learning techniques, the potential of semi-supervised learning remains to be excavated.

    
   \noindent \textbf{(DL) Zero-shot learning:}
        In LLIE, zero-shot learning is a learning strategy that requires neither paired data nor unpaired training datasets. Instead, zero-shot learning learns image enhancement at test time using data-free loss functions such as exposure loss or color loss. For example, ExCNet~\cite{zhang2019zero} introduces a zero-shot CNN based on estimating the "S-curve" that best fits the exposure of the back-lit images. Zero-DCE~\cite{guo2020zero} and Zero-DCE++~\cite{li2021learning} utilize zero-reference deep curve estimation and dynamic range adjustment. RUAS~\cite{liu2021retinex} develops a Retinex-inspired unrolling model with Neural Architecture Search (NAS). RetinexDIP~\cite{zhao2021retinexdip} proposes a Retinex-based zero-shot method using generative decomposition and latent component estimation. SGZ~\cite{zheng2022semantic} leverages pixel-wise light deficiency estimation and unsupervised semantic segmentation. Thanks to the zero-reference loss functions, zero-shot learning methods have outstanding generalization ability, require few parameters, and have fast inference speed.
    
    

    \noindent \textbf{Discussion:}
    The aforementioned traditional and deep learning strategies for LLIE have the following limitations.

    \begin{itemize}
        \item Traditional Learning methods' performances lag behind deep learning methods, even with their handcrafted priors and intricate optimization steps, which result in poor inference latency.
        \item Supervised Learning methods rely heavily on the paired training dataset, but none of the approaches for obtaining such a dataset is feasible. Specifically, it is difficult to capture image pairs that are only different in illuminations; it is hard to synthesize images that fit the complex real-world scenes; it is expensive and time-consuming to retouch large-scale low-light images. 
        \item Unsupervised Learning methods' dependencies on the unpaired training dataset induces data bias. Because of data bias, unsupervised learning methods like EnlightenGAN (EGAN)~\cite{jiang2021enlightengan} and SCI~\cite{ma2022toward} generalize poorly to testing datasets with significant domain gaps.
        \item Semi-supervised learning methods inherit the limitations of both supervised and unsupervised learning methods without fully utilizing their strengths. That's why semi-supervised learning has been used by only one representative LLIE method DRBN~\cite{yang2020fidelity}.
        \item Zero-shot learning methods require elaborate designs for the data-free loss functions. Still, they cannot cover all the necessary properties of real-world low-light images. Besides, their performances lag behind supervised learning methods like LLFlow~\cite{wang2022low} on benchmark datasets.
    \end{itemize}

\subsection{Network Structures}

    
    Many LLIE methods utilize a U-Net-like (37.2 \%) structure or multi-scale information (18.6 \%). Some methods use transformers  (7.0 \%) or unfolding networks (7.0 \%). A few methods (2.3 \% for each) use Neural Architecture Search (NAS), Normalizing Flow, Optical Flow, Recurrent Network, or Recursive Network. 
    
    \noindent \textbf{U-Net and Multi-Scale:}
   U-Net-like~\cite{ronneberger2015u} structure is the most popular in LLIE since it preserves high-resolution rich detail features and low-resolution rich semantic features, which are both essential for LLIE. Similarly, other structures that use multi-scale information are also welcomed in LLIE.
    
    \noindent \textbf{Transformers:}
    Recently, the transformers-based~\cite{dosovitskiy2020image} method has surged in computer vision, especially high-level vision tasks, due to its ability to track long-range dependencies and capture global information in an image. 
    

    \noindent \textbf{Unfolding and NAS:}
      The unfolding network (a.k.a. unrolling network)~\cite{zhang2020deep} has been used by several methods because it combines the wisdom of model-based and data-based approaches. NAS~\cite{zoph2016neural} is the automating design of a neural network that generates the optimal result with a given dataset. 
    
    
    
    \noindent \textbf{Normalizing Flow:}
    Normalizing flow-based~\cite{rezende2015variational} method transforms a simple probability distribution into a complex distribution with a sequence of invertible mappings.  

    \noindent \textbf{Optical Flow:}
    Optical flow-based~\cite{horn1981determining} methods estimate the pixel-level motions of adjacent video frames.
    
    
    \noindent \textbf{Recurrent and Recursive:}
    Recurrent network~\cite{rumelhart1985learning} is a type of neural network that repeatedly process the input in chain structures, whereas recursive network~\cite{goller1996learning} is a variant of the recurrent network that processes the input in hierarchical structures. 
    

    \noindent \textbf{Discussion:}
    The aforementioned network structures for LLIE have the following limitations. 

    \begin{itemize}
        \item Transformers is currently unpopular in low-level vision tasks like LLIE. Perhaps this is due to their impotence to integrate local and non-local attention and inefficiency at processing high-resolution images.
        \item The unfolding strategy requires elaborate network design, whereas NAS requires expensive parameter learning. 
        \item The normalizing flow and optical flow-based methods have poor computational efficiency. 
        \item The recurrent network have the vanishing gradient problem at large-scale data~\cite{lu2022introvae}, whereas the recursive network relies on the inductive bias of hierarchical distribution, which is unrealistic.
    \end{itemize}

\subsection{Loss Functions}
The choice of loss functions is highly diverse among LLIE methods. 55.2 \% of the LLIE methods use non-mainstream loss functions. Among mainstream loss functions, $L_1$ (11.5 \%) is the most popular, whereas $L_2$ (6.9 \%), Negative SSIM (6.9 \%), and Illumination Smoothness (6.9 \%) are also popular. A small quantity of methods use Total Variation (5.7 \%), Perceptual (3.4 \%), or Illumination Adjustment  (3.4 \%).

    \noindent \textbf{Full-Reference loss:}
    $L_1$ loss, $L_2$ loss, Negative SSIM loss, Perceptual loss~\cite{johnson2016perceptual} and Illumination adjustment loss~\cite{zhang2019kindling} are full-reference loss functions (i.e., loss requiring paired images). $L_1$ loss targets the absolute difference between image pairs, whereas $L_2$ loss targets the squared difference. Therefore, $L_2$ loss is rigid for large errors but tolerant for small errors, whereas $L_1$ loss does the opposite. Like other low-level vision tasks~\cite{zhao2016loss}, $L_1$ loss in LLIE is more popular than $L_2$ loss.  Negative SSIM loss is based upon the negative SSIM score. Essentially, it reflects the difference of image pairs in terms of luminance, contrast, and structure. However, Negative SSIM loss is uncommon in LLIE. That is different from other low-level vision tasks like image deraining, where it gains great popularity~\cite{ren2019progressive,wang2019spatial,zheng2022sapnet}. Perceptual loss is the $L_2$ difference of image pairs based on their features extracted from a pretrained convolutional neural network (e.g., VGG-16~\cite{simonyan2014very}). It is popular in low-level vision tasks like style transfer~\cite{johnson2016perceptual,liu2017depth} but is less explored in LLIE. Illumination adjustment loss is the $L_2$ difference for illumination and illumination gradients between image pairs. Due to its task-specific nature, it has only been applied in LLIE algorithms~\cite{zhang2019kindling,zhang2021beyond,wu2022uretinex}.

    \noindent \textbf{Non-Reference loss:}
     Total Variation (TV) loss~\cite{vogel1996iterative} and illumination smoothness loss~\cite{zhang2019kindling} are non-reference loss functions (i.e., losses that do not require paired images). TV loss measures the sum of the difference between adjacent pixels in vertical and horizontal directions for an image. Therefore, TV loss suppresses irregular patterns like noise and blur and promotes smoothness in the image. Illumination smoothness loss is similar to TV loss since it is written as the $L_1$ norm of illumination divided by the maximum variation. Despite their success at other low-level vision tasks like denoising~\cite{chen2010adaptive} and deblurring~\cite{oliveira2009adaptive}, the variation-based methods have been less explored in LLIE. 

\subsection{Evaluation Metrics}
Many LLIE methods choose PSNR (18.8 \%) or SSIM (18.1 \%) as the evaluation metrics. Apart from PSNR and SSIM, the User Study (US) (8.1 \%) is a popular method. Several methods use NIQE (5.4 \%), Inference Time (5.4 \%), \#Params (4.7 \%), FLOPs (4.0 \%), LOE (4.0 \%), or LPIPS (3.4 \%) as the evaluation metrics. 

    \noindent \textbf{Full-Reference Metrics:}
    Peak Signal-to-Noise Ratio (PSNR), Structure Similarity Index (SSIM), and Learned Perceptual Image Patch Similarity (LPIPS)~\cite{zhang2018unreasonable} are full-reference image quality evaluation metrics. PSNR measures the pixel-level similarity between image pairs, whereas SSIM measures the similarity according to luminance, contrast, and structure. LPIPS measures the patch-level difference between two images using a pretrained neural network. Higher PSNR and SSIM and lower LPIPS indicate better visual quality. 

    \noindent \textbf{Non-Reference Metrics:}
    Natural Image Quality Evaluator (NIQE)~\cite{mittal2012making} and Lightness order Error (LOE)~\cite{wang2013naturalness} are non-reference image quality evaluation metrics. Specifically, NIQE is based on the naturalness score for an image using a model trained with natural scenes, whereas LOE indicate the lightness-order errors for that image. A lower NIQE and LOE indicate better visual quality. 

    \noindent \textbf{Subjective Metrics:}
    User study is the only subjective metric used for representative LLIE methods. Typically, the user study score is the mean opinion score from a group of participants. A high user study score means better perceptual quality from human perspectives. 

    \noindent \textbf{Efficiency Metrics:}
    Efficiency metrics include inference time, Numbers of Parameters (\#Params), and Floating Point Operations (FLOPs). A shorter inference time and a smaller \#Params and FLOPs indicates better efficiency.

\subsection{Training and Testing Data}
Popular benchmark training data for LLIE include LOL (20.8 \%), SID (12.5 \%), and MIT-Adobe FiveK (8,3 \%). Alternative choices include SICE (6.2 \%), ACDC (4.2 \%), SDSD (4.2 \%), and SMID (4.2 \%). Meanwhile, many utilize their custom dataset (16.7 \%). Popular benchmark testing data for LLIE include LOL (11.0 \%) and LIME (8.0 \%). Some utilize MEF (7.0 \%), NPE (7.0 \%), and SID (7.0 \%), while others use DICM (6.0 \%), MIT-Adobe FiveK (5.0 \%), VV (5.0 \%), ExDark (4.0 \%), or SICE (4.0 \%). Similar to the case in training data, many methods utilize their custom dataset (7.0 \%) for testing. A detailed discussion for training and testing dataset is given in Section \ref{sec:evaluations}.

\begin{table}[t]
\centering
\caption{Table summary for several important topics. See Section \ref{others} for details. }
\begin{tabular}{l|cc}
\toprule
\textbf{Topics} & \textbf{Yes (\%)} & \textbf{No (\%)} \\ \hline

RGB         & \textbf{81.8}         & 18.2                \\ 

Limitations   & \textbf{64.7}            & 35.3            \\ 
 
New datasets   & 44.1            & \textbf{55.9}       \\ 

Retinex       & 41.2           & \textbf{58.8}           \\
 
Applications  & 20.6            & \textbf{79.4}            \\ 
 
Video         & 11.8            & \textbf{88.2}            \\ \bottomrule
               
\end{tabular}
\end{table}

\subsection{Others} \label{others}

\noindent \textbf{New Dataset:} The number of LLIE methods that introduce new datasets (55.9 \%) surpasses the number of LLIE methods that only use existing datasets (44.1 \%). This reflects the importance of data for LLIE.

\noindent \textbf{Limitations:} Most LLIE methods (64.7 \%) do not mention their limitations and future works. This makes it hard for future researchers to improve upon their work. 

\noindent \textbf{Applications:} Most LLIE methods (79.4 \%) do not relate low-level image enhancement to high-level applications like detection or segmentation. Therefore, the practical values of these methods remain a question.  

\noindent \textbf{Retinex:} Lots of methods (41.2 \%) utilize Retinex theory for LLIE enhancement. However, most LLIE methods (58.8 \%) do not utilize Retinex theory. Hence, the Retinex theory remains a popular but non-dominant choice for LLIE.

\noindent \textbf{Video:} A majority of LLIE methods (88.2 \%) do not consider Low-Light Video Enhancement (LLVE) tasks. Sadly, most real-world low-light visual data are stored in video format. 

\noindent \textbf{RGB:} A majority of LLIE methods (81.8 \%) uses RGB data for training. This is great since RGB is much more popular than RAW for modern digital devices like laptops or smartphones.

\begin{table}[t]
\setlength\tabcolsep{4pt}
\centering
\caption{Table summary of existing benchmark datasets versus proposed SICE\_Grad and SICE\_Mix. `Trains' indicates the inclusion of training datasets; `Paired' signifies the presence of paired images; `Task' denotes suitability for high-level vision tasks. Abbreviations `y', `n', and `b' represent `yes', `no', and `both', respectively. See Section \ref{sec:datasets} for details.  }
\begin{tabular}{l|r|c|c|c|c|c}
\toprule
\textbf{Dataset} & \textbf{Number} & \textbf{Resolutions} & \textbf{Type} & \textbf{Train} & \textbf{Paired} & \textbf{Task} \\ \hline
NPE~\cite{wang2013naturalness}             & 8               & Various              & Real          & N              & N               & N             \\ 
LIME~\cite{guo2016lime}             & 10              & Various              & Real          & N              & N               & N             \\ 
MEF~\cite{ma2015perceptual}             & 17              & Various              & Real          & N              & N               & N             \\ 
DICM~\cite{lee2013contrast}             & 64              & Various              & Real          & N              & N               & N             \\
VV                & 24              & Various              & Real          & N              & N               & N             \\
LOL~\cite{wei2018deep}              & 500             & 400 $\times$ 600          & Real          & Y              & Y               & N             \\ 
VE-LOL~\cite{liu2021benchmarking}           & 13,440          & Various              & Both          & B              & Y               & Y             \\
ACDC~\cite{sakaridis2021acdc}             & 4,006           & 1080 $\times$ 1920         & Real          & Y              & N               & Y             \\
DCS~\cite{zheng2022semantic}              & 150             & 1024 $\times$ 2048          & Syn           & N              & Y               & Y             \\
DarkFace~\cite{yang2020advancing}         & 6,000           & 720 $\times$ 1080        & Real          & Y              & N               & Y             \\ 
ExDark~\cite{loh2019getting}           & 7,363           & Various              & Real          & Y              & N               & Y             \\
MIT~\cite{bychkovsky2011learning}              & 5,000           & Various              & Both          & Y              & Y               & N             \\ 
MCR~\cite{dong2022abandoning}             & 3,944           & 1024 $\times$ 1280       & Both          & Y              & Y               & N             \\ 
LSRW~\cite{hai2023r2rnet}             & 5,650           & Various              & Real          & Y              & Y               & N \\
TYOL~\cite{hodan2018bop} & 5,991 & Various & Syn & Y & Y & N \\
SID~\cite{cheng2016learning}         & 5,094           & Various              & Real          & Y              & Y      & N  \\
SDSD~\cite{wang2021seeing}             & 37,500          & 1080 $\times$ 1920         & Real          & Y              & Y               & N             \\ 
SMID~\cite{chen2019seeing}            & 22,220          & 3672 $\times$ 5496          & Real          & Y              & Y               & N             \\ 
SICE~\cite{cai2018learning}     & 4,800   & Various         & Both  & Y    & Y  & N       \\ 
SICE\_Grad       & 589            & 600$\times$900 & Both  &  Y     &  Y  & N \\ 
SICE\_Mix        & 589             & 600$\times$900 & Both  &  Y    & Y  & N \\ \bottomrule
\end{tabular}
\label{tab:dataset}
\end{table}

\section{Datasets} \label{sec:datasets}

\subsection{Benchmark Datasets} \label{benchmark_datasets}

\noindent \textbf{NPE\cite{wang2013naturalness}/ LIME\cite{guo2016lime}/ MEF\cite{ma2015perceptual}/ DICM\cite{lee2013contrast}} carries 8/10/17/64 real low-light images of various resolutions. They contain indoor items and decorations, outdoor buildings, streetscapes, and natural landscapes, and they are all for testing. 

\noindent \textbf{VV}\footnote{https://sites.google.com/site/vonikakis/datasets} contains 24 real multi-exposure images of various resolutions. It contains traveling photos with indoor and outdoor persons and natural landscapes for testing. 

\noindent \textbf{LOL}~\cite{wei2018deep} contains 500 pairs of real low-light images of 400 $\times$ 600 resolutions. It only contains indoor items and divides into 485 training images and 15 testing images. 

\noindent \textbf{VE-LOL}~\cite{liu2021benchmarking} contains 13,440 real and synthetic low-light images and image pairs of various resolutions. It has diversified scenes, such as natural landscapes, streetscapes, buildings, human faces, etc. The paired portion VE-LOL-L has 2,100 pairs for training and 400 pairs for testing, whereas the unpaired portion VE-LOL-H has 6,940 images for training and 4,000 for testing. Additionally, the VE-LOL-H portion contains detection labels for high-level object detection tasks. 

\noindent \textbf{ACDC}~\cite{sakaridis2021acdc} contains 4,006 real low-light images of resolution 1,080 $\times$ 1,920. It includes autonomous driving scenes with adverse conditions (1,000 foggy, 1,000 snowy, 1,000 rainy, and 1,006 nighttime) and has 19 classes. In particular, the ACDC nighttime contains 400 training images, 106 validation images, and 500 test images. Besides, ACDC contains semantic segmentation labels which allow high-level semantic segmentation tasks. 

\noindent \textbf{DCS}~\cite{zheng2022semantic} contains 150 synthetic low-light images of resolution 1,024 $\times$ 2,048. Specifically, it is synthesized with gamma correction upon the original CityScape~\cite{cordts2016cityscapes} dataset, and it contains urban scenes with fine segmentation labels (30 classes). Therefore, it permits high-level instance segmentation, semantic segmentation, and panoptic segmentation tasks. The Dark CityScape (DCS) dataset is intended for testing only. 

\noindent \textbf{DarkFace}~\cite{yang2020advancing} contains 10,000 real low-light images of resolution 720 $\times$ 1,080. It contains nighttime streetscapes with many human faces in each image. It consists of 6,000 training and validation images and 4,000 testing images. With object detection labels, it can be applied to high-level object detection tasks. 

\noindent \textbf{ExDark}~\cite{loh2019getting} contains 7,363 real low-light images of Various resolutions. It contains images with diversified indoor and outdoor scenes under 10 illumination conditions with 12 object classes. It is split into 4,800 training images and 2,563 testing images. It contains object detection labels and can be applied to high-level object detection tasks. 

\noindent \textbf{SICE}~\cite{cai2018learning} contains 4,800 real and synthetic multi-exposure images of various resolutions. It contains images with diversified indoor and outdoor scenes with different exposure levels. The train/val/test follows a 7:1:2 ratio. In particular, SICE contains both normal-exposed and ill-exposed images. Therefore, it can be used for supervised, unsupervised, and zero-shot learning. 


\noindent \textbf{SID}~\cite{cheng2016learning} contains 5,094 real short-exposure images, each with a matched long-exposure reference image. The resolution is 4,240 $\times$ 2,832 for Sony and 6,000 $\times$ 4,000 for Fuji images. The train/val/test follows a 7:1:2 ratio. It contains indoor and outdoor images, where the illuminance of the outdoor scene is 0.2lux~5lux, and the illuminance of the indoor scene is 0.03lux~0.3lux.

\noindent \textbf{SDSD}~\cite{wang2021seeing} contains 37,500 real images of resolution 1,080 $\times$ 1,920. It is the first high-quality paired video dataset for dynamic scenarios, containing identical scenes and motion in high-resolution video pairs in both low- and normal-light conditions.

\noindent \textbf{SMID}~\cite{chen2019seeing} contains 22,220 real low-light images of resolution 3672 $\times$ 5496. The dataset was randomly divided into 3 groups: training (64 \%), validation (12 \%), and testing (24 \%). Some scenarios include various lighting setups, including light sources with various color temperatures, levels of illumination, and placements.

\noindent \textbf{MIT Adobe FiveK}~\cite{bychkovsky2011learning} contains 5,000 real and synthetic images in diverse light conditions in various resolutions, including the RAW images taken directly from the camera and the edited versions created by 5 professional photographers. The dataset is divided into 80 \% for training and 20 \% for testing.

\noindent \textbf{MCR}~\cite{dong2022abandoning} contains 3,944 real and synthetic short-exposure and long-exposure images of 1,024 $\times$ 1,280 resolutions with monochrome and color raw pairs. The dataset is divided into train and test sets with a 9:1 ratio. The image are collected in both indoor fixed positions, and indoor/outdoor sliding platforms conditions.


\noindent \textbf{LSRW}~\cite{hai2023r2rnet} contains 5,650 real low-light paired images of various resolutions with indoor and outdoor scenes. 5,600 paired images are selected for training and the remaining 50 pairs are for testing.

 \noindent \textbf{TYOL}~\cite{hodan2018bop} contains 5,991 synthetic images and most images conform to VGA resolution. It splits into 2,562 training images, 1,680 test images, and 1,669 test targets. The subset of TYOL, TYO-L (Toyota Light) contains texture-mapped 3D models of 12 objects with a wide range of sizes.


\noindent \textbf{Discussion:} The current benchmark datasets for LLIE have the following limitations. 


\begin{itemize}
    \item Many datasets use synthetic images to meet the paired image requirement for supervised learning methods. These image synthesis techniques often follow simple gamma correction or exposure adjustment, which does not fit the diverse illuminations in the real-world. Consequently, methods trained with these synthetic images generalize poorly to the real-world images. 
    \item The existing datasets consider single images rather than both images and videos. That is because that high-quality low-light videos are hard to capture and that many methods cannot process high-resolution video frames in real-time. However, the real-world applications (e.g., visual surveillance, autonomous driving, and UAV) are heavily dependent on videos rather than single images. Hence, the lack of low-light video dataset greatly undermines the benefits of LLIE for these fields.
    \item The existing datasets either consider underexposure or overexposure only, or consider underexposure and overexposure in separate images in a dataset. There is no dataset that contains mixed under-/overexposure in single images. See the detailed discussions in Subsection \ref{exposure_analysis}. 
    
\end{itemize}














\begin{figure}[t]
    \centering
    \subfloat[SICE (Under)~\cite{cai2018learning}]{
    \includegraphics[width=0.17\columnwidth]{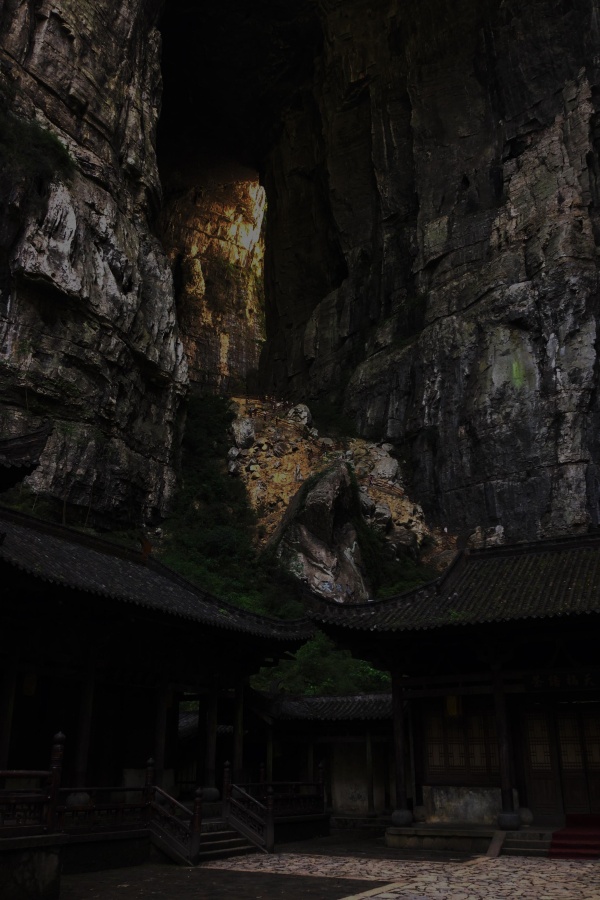}
    \includegraphics[width=0.38\columnwidth]{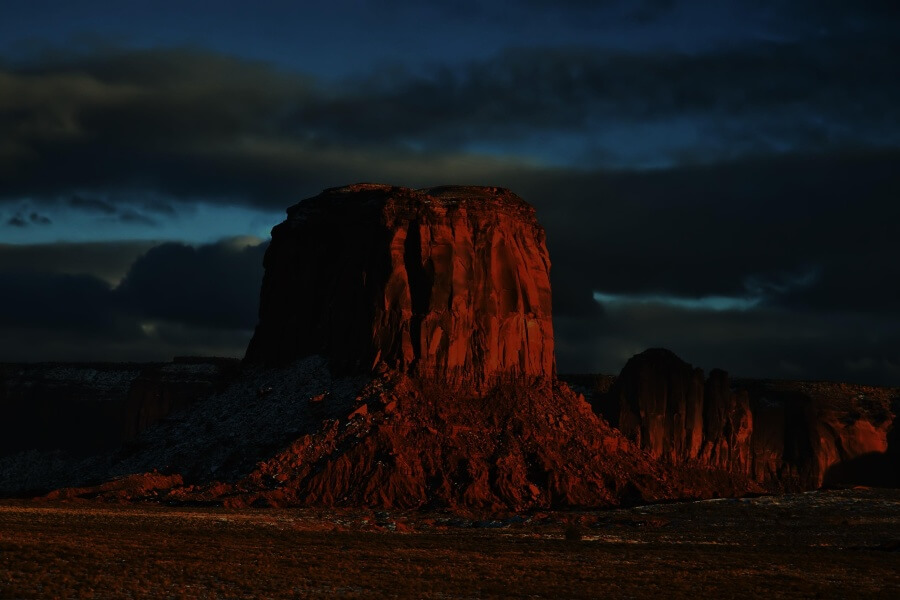} 
    \includegraphics[width=0.38\columnwidth]{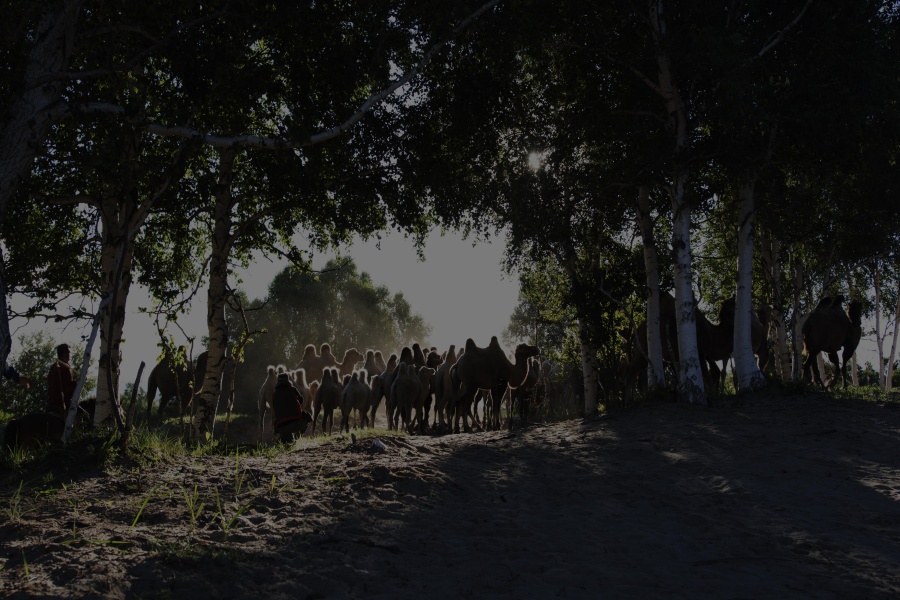}
    }
     \\
    \subfloat[SICE (Normal)~\cite{cai2018learning}]{
    \includegraphics[width=0.17\columnwidth]{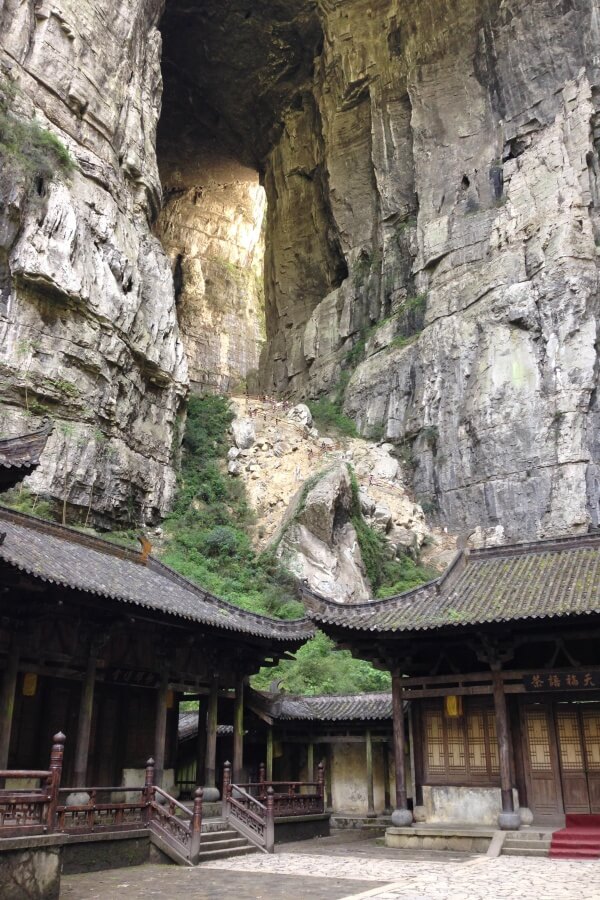}
    \includegraphics[width=0.38\columnwidth]{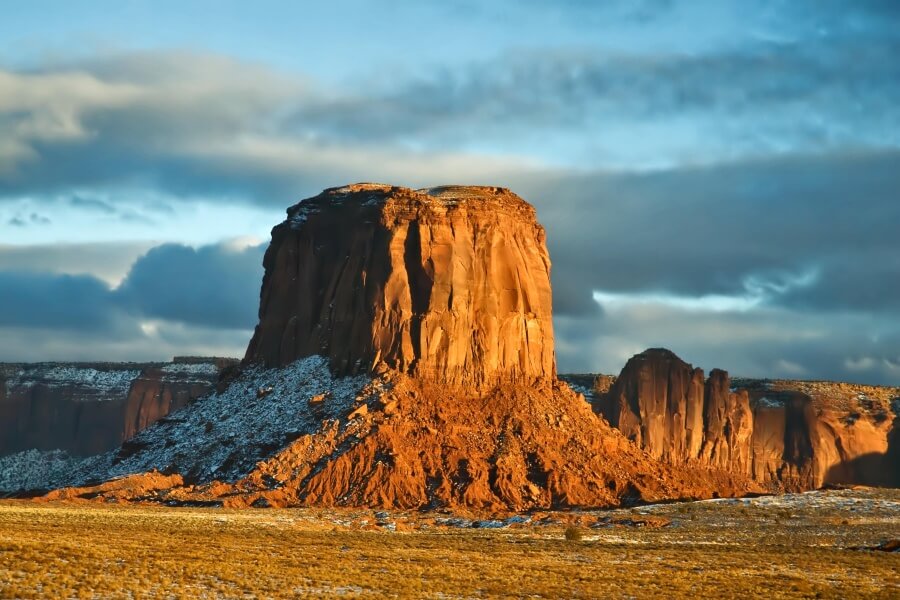}  
    \includegraphics[width=0.38\columnwidth]{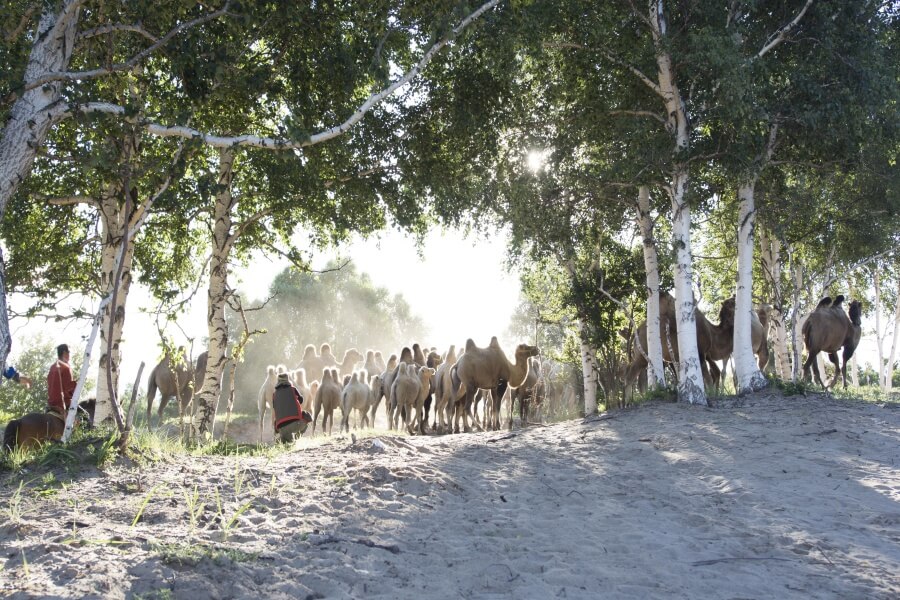}
    }
 \\
     \subfloat[SICE (Over)~\cite{cai2018learning}]{
    \includegraphics[width=0.17\columnwidth]{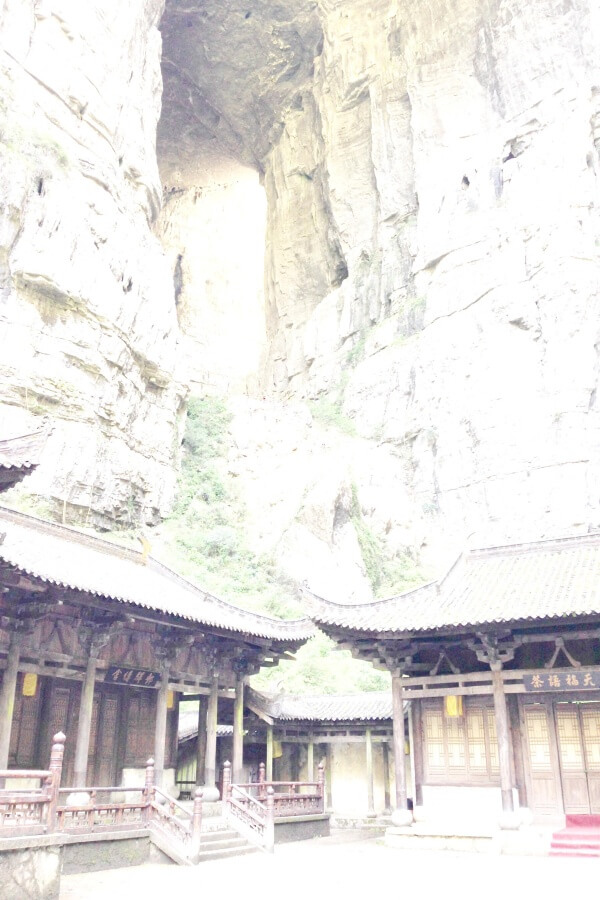}
    \includegraphics[width=0.38\columnwidth]{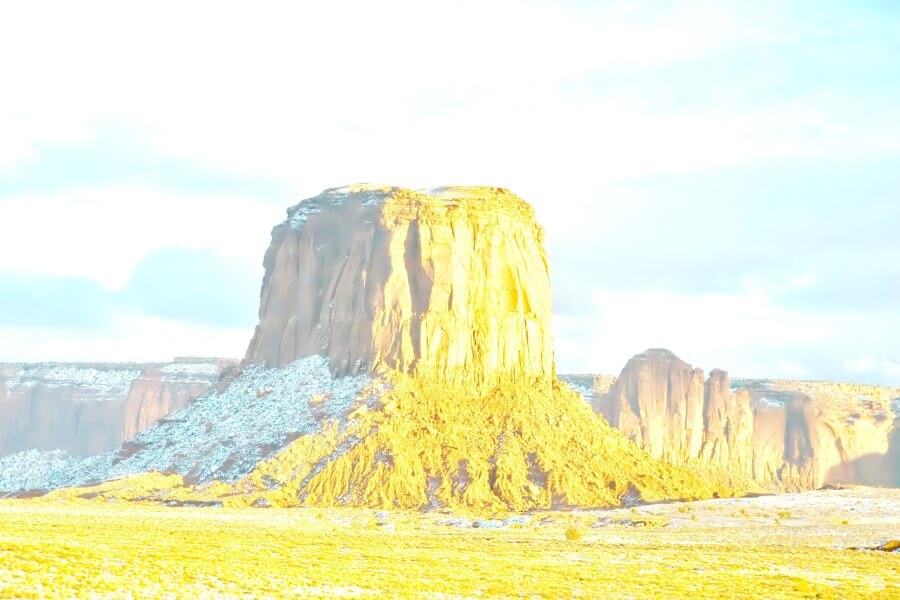}  
    \includegraphics[width=0.38\columnwidth]{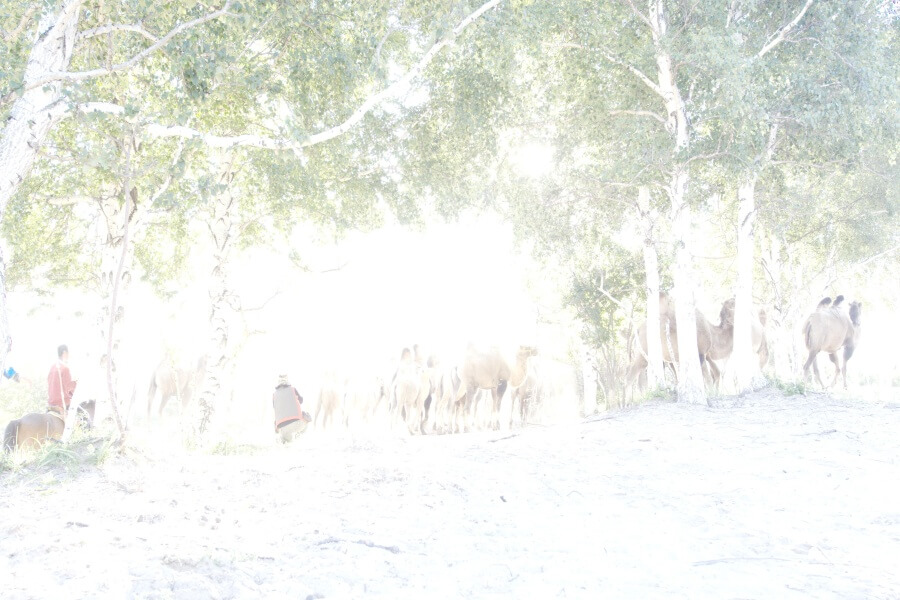}
     }
 \\       
    \subfloat[SICE\_Grad]{
    \includegraphics[width=0.17\columnwidth]{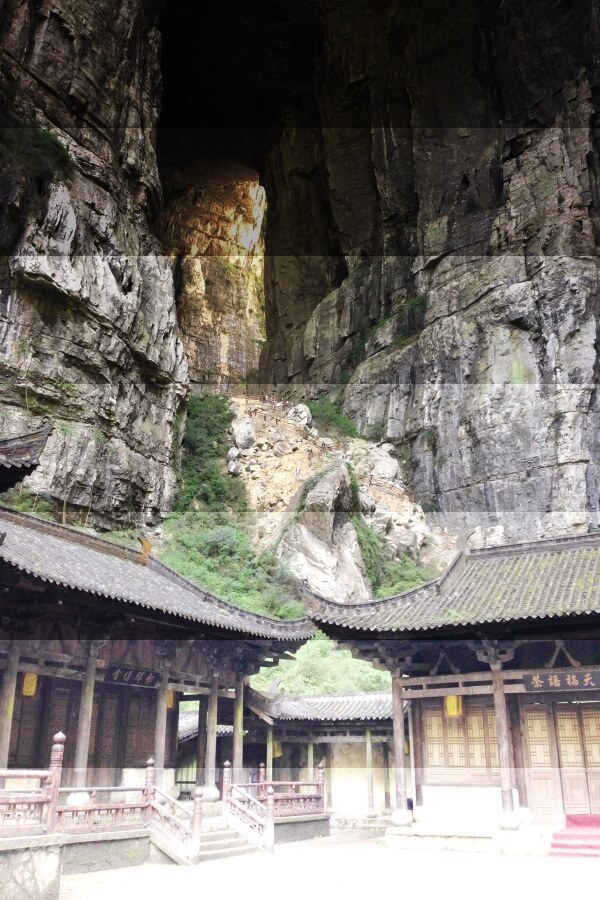}
    \includegraphics[width=0.38\columnwidth]{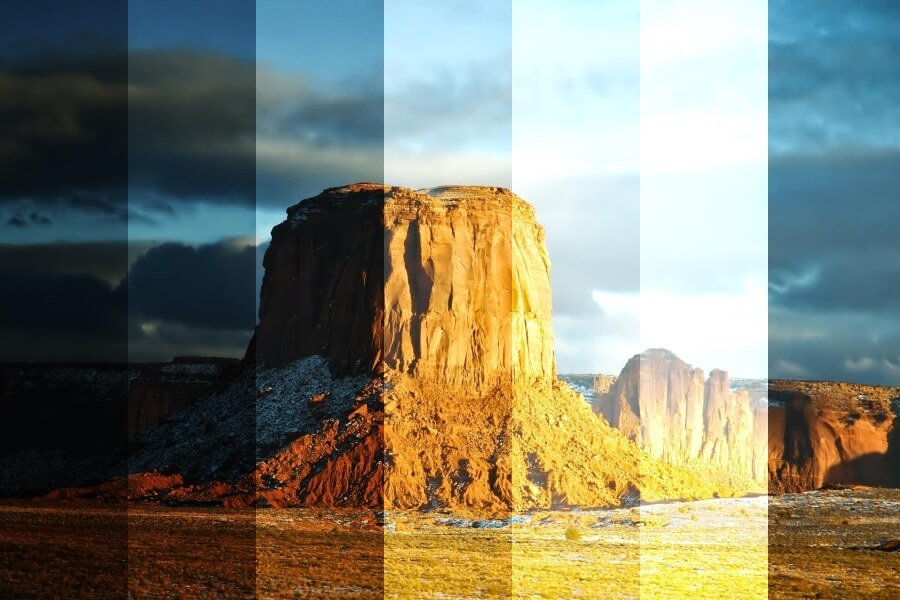}  
    \includegraphics[width=0.38\columnwidth]{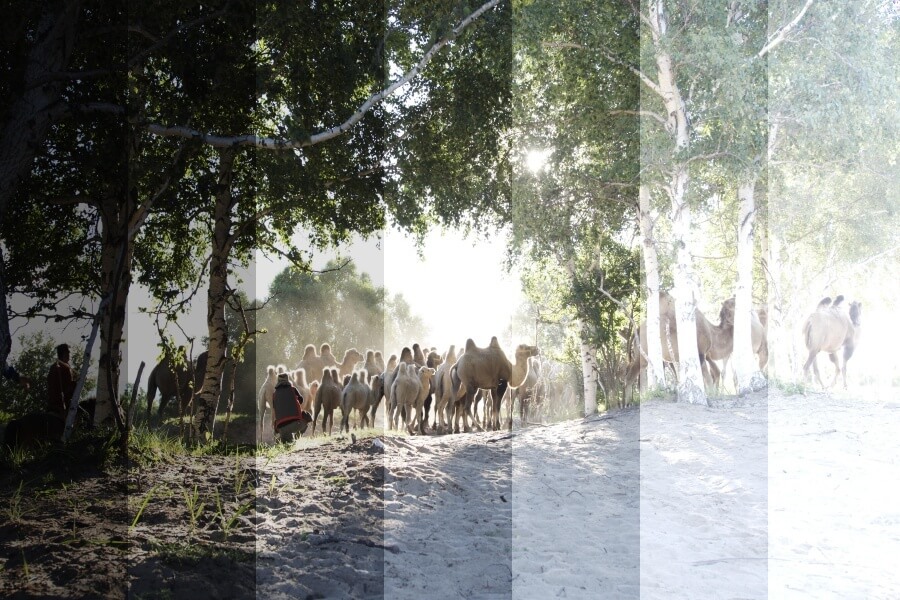}
}
 \\
     \subfloat[SICE\_Mix]{
    \includegraphics[width=0.17\columnwidth]{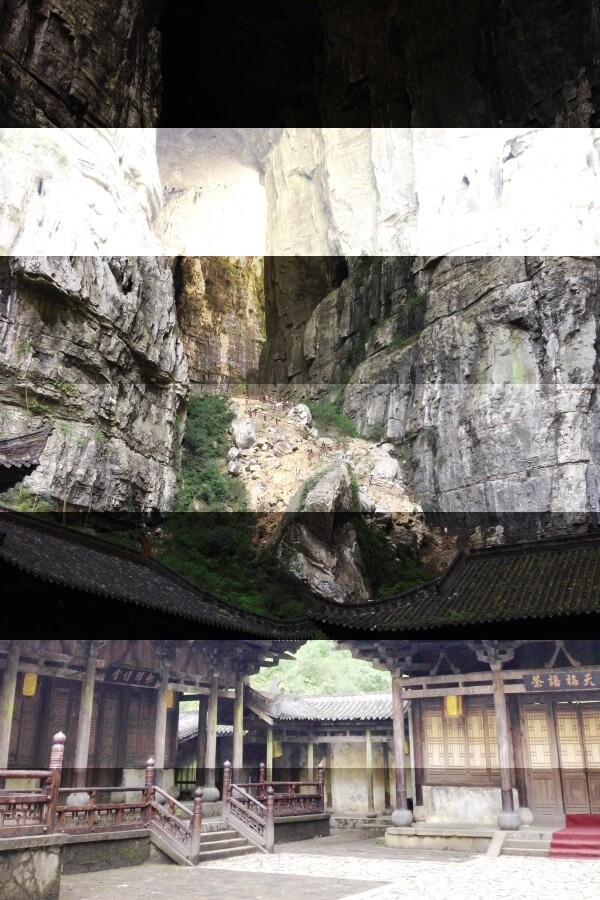}
    \includegraphics[width=0.38\columnwidth]{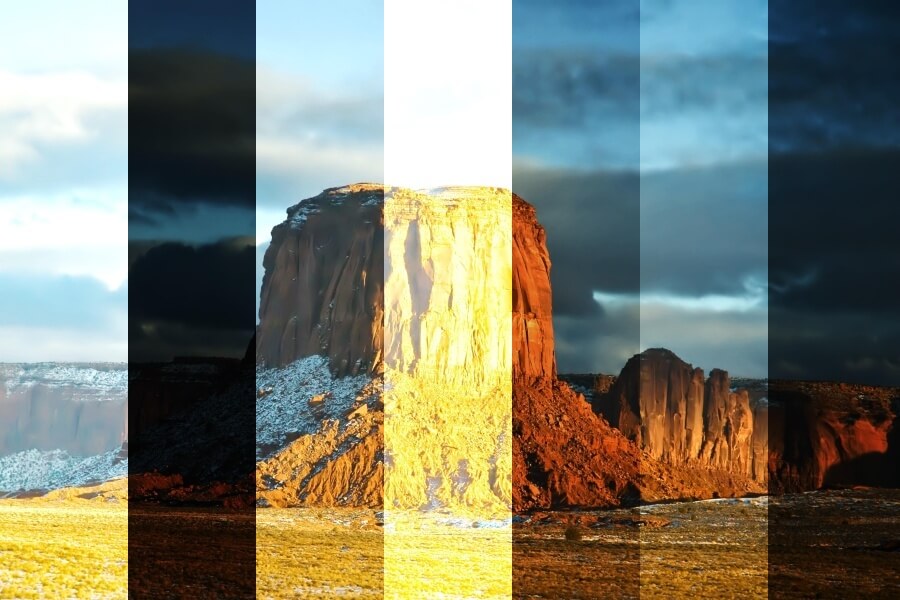} 
    \includegraphics[width=0.38\columnwidth]{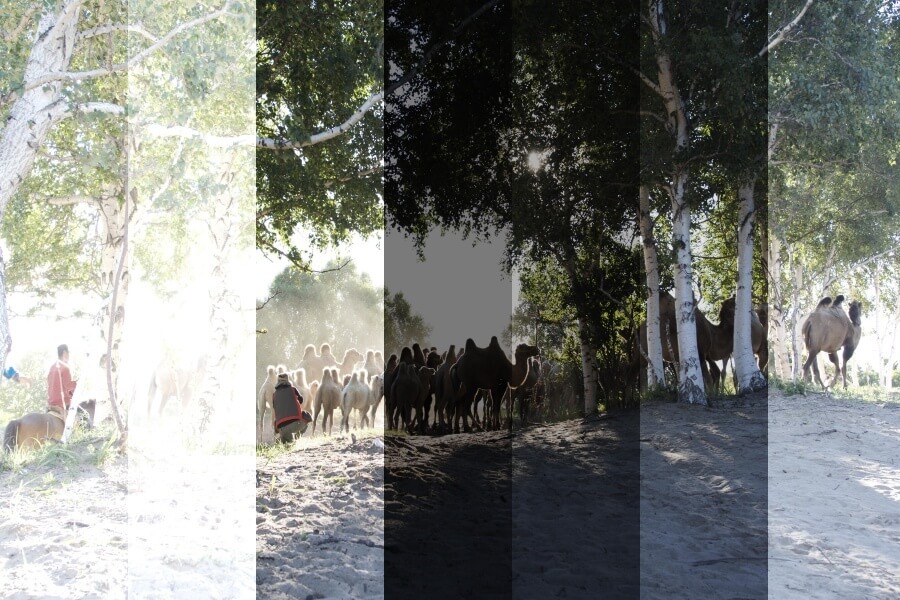}
     }
 \\    
    \caption{\textbf{Example Images from SICE (Under-Exposure, Normal-Exposure, Over-Exposure) and our SICE\_Grad and SICE\_Mix.} SICE\_Grad is created by permuting panels from low to high exposure, with the exception of the rightmost panel, which is randomized. SICE\_Mix is created by randomly permuting panels. See Section \ref{new_image_dataset} for details. }
    \label{fig:SICE_Example}
    
\end{figure}

\begin{figure}[t]
    \centering
    \subfloat[DCS~\cite{zheng2022semantic}]{\includegraphics[width=0.5\columnwidth]{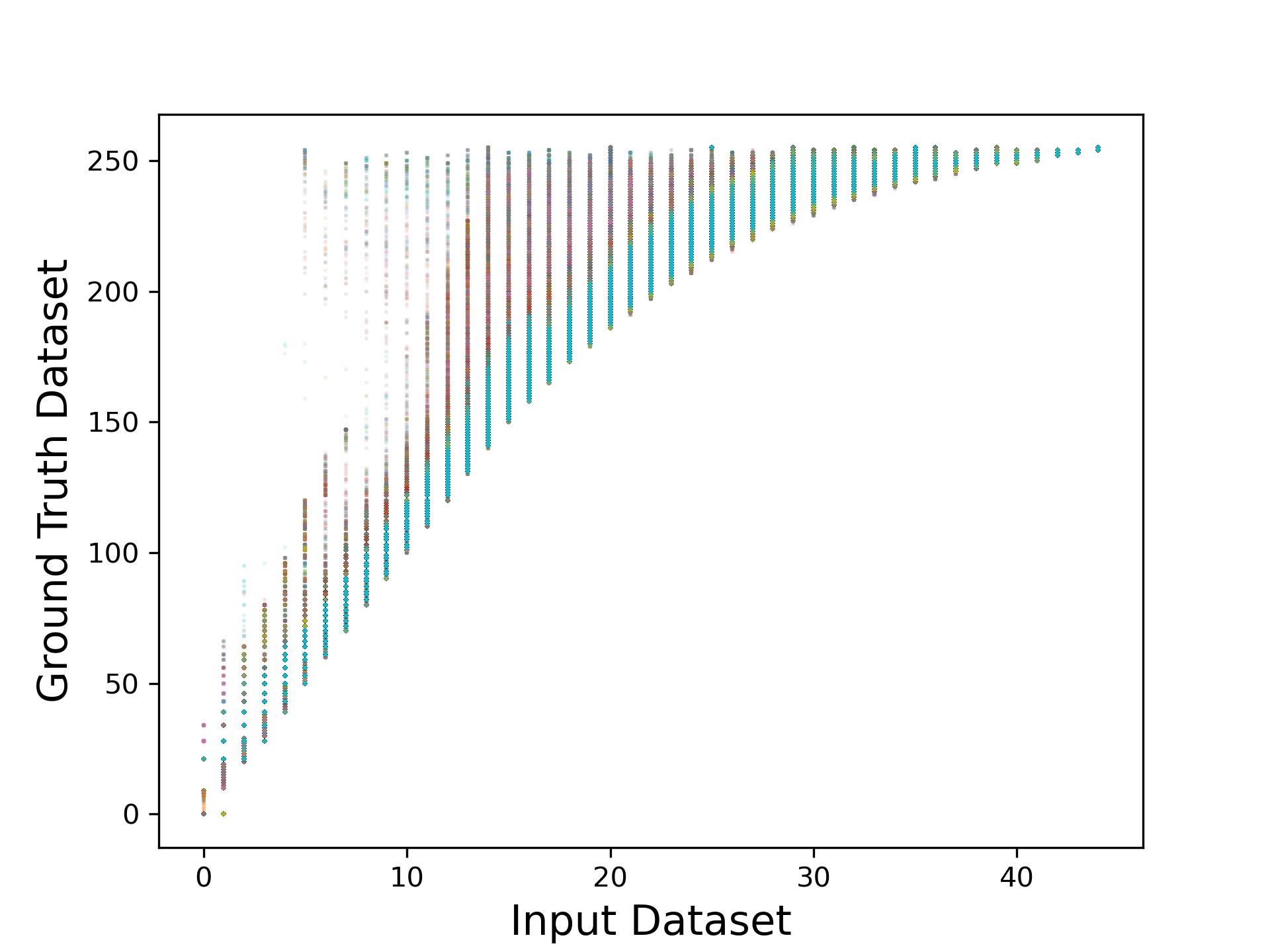}}
    \subfloat[LOL~\cite{wei2018deep}]{\includegraphics[width=0.5\columnwidth]{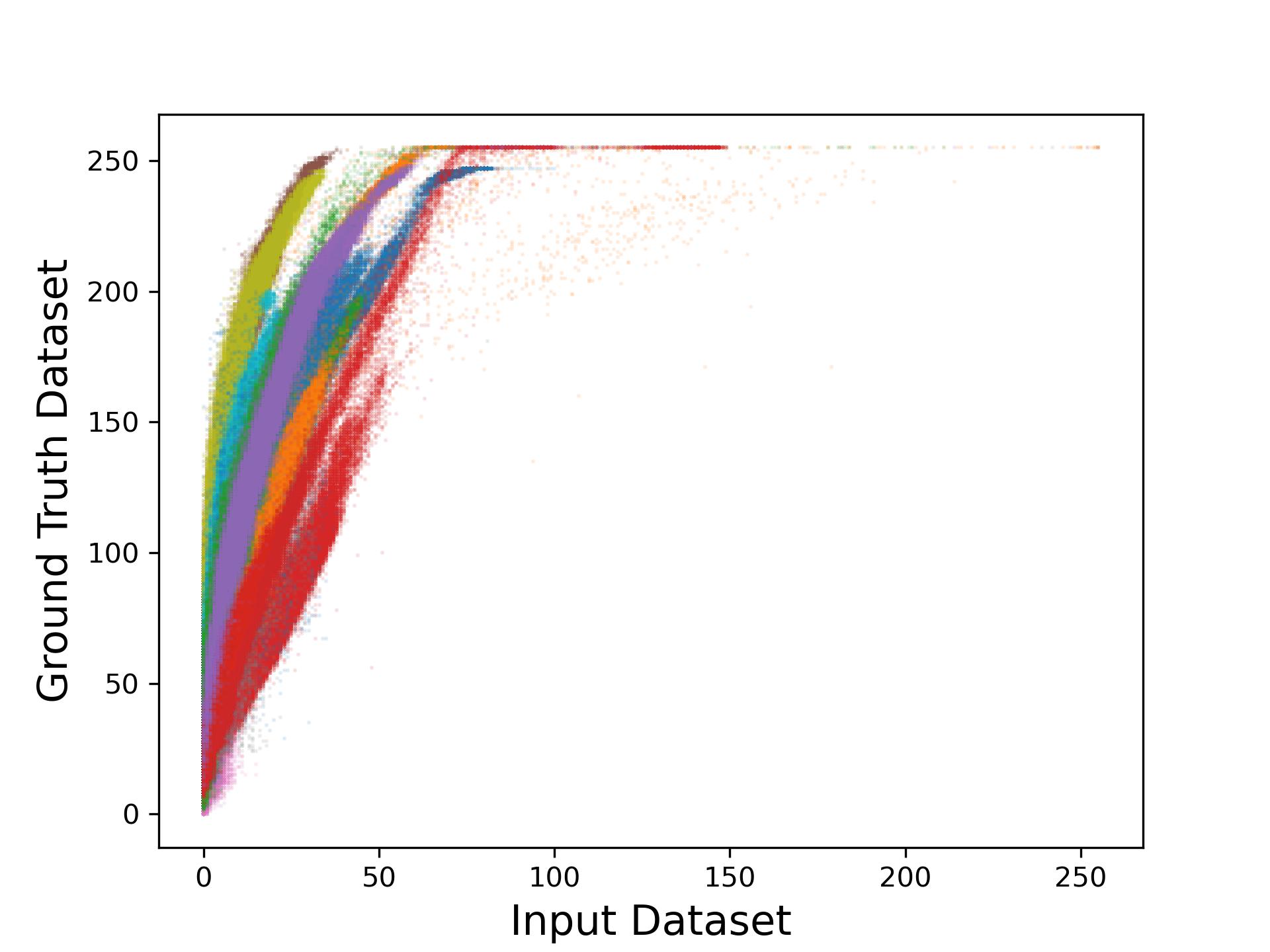}}
    \\
    \subfloat[VELOL\_Syn~\cite{liu2021benchmarking}]{\includegraphics[width=0.5\columnwidth]{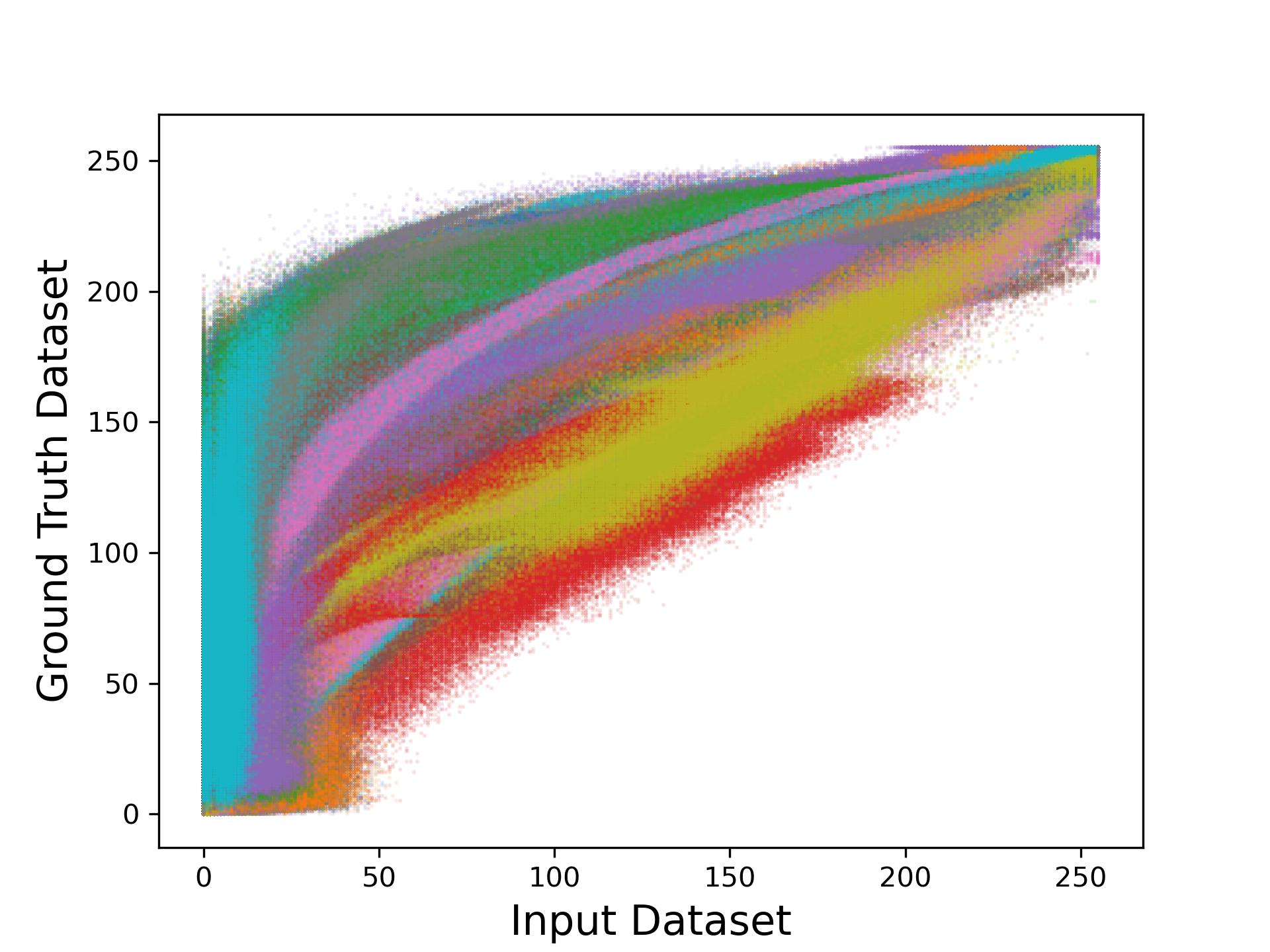}} 
    \subfloat[VELOL\_Real~\cite{liu2021benchmarking}]{\includegraphics[width=0.5\columnwidth]{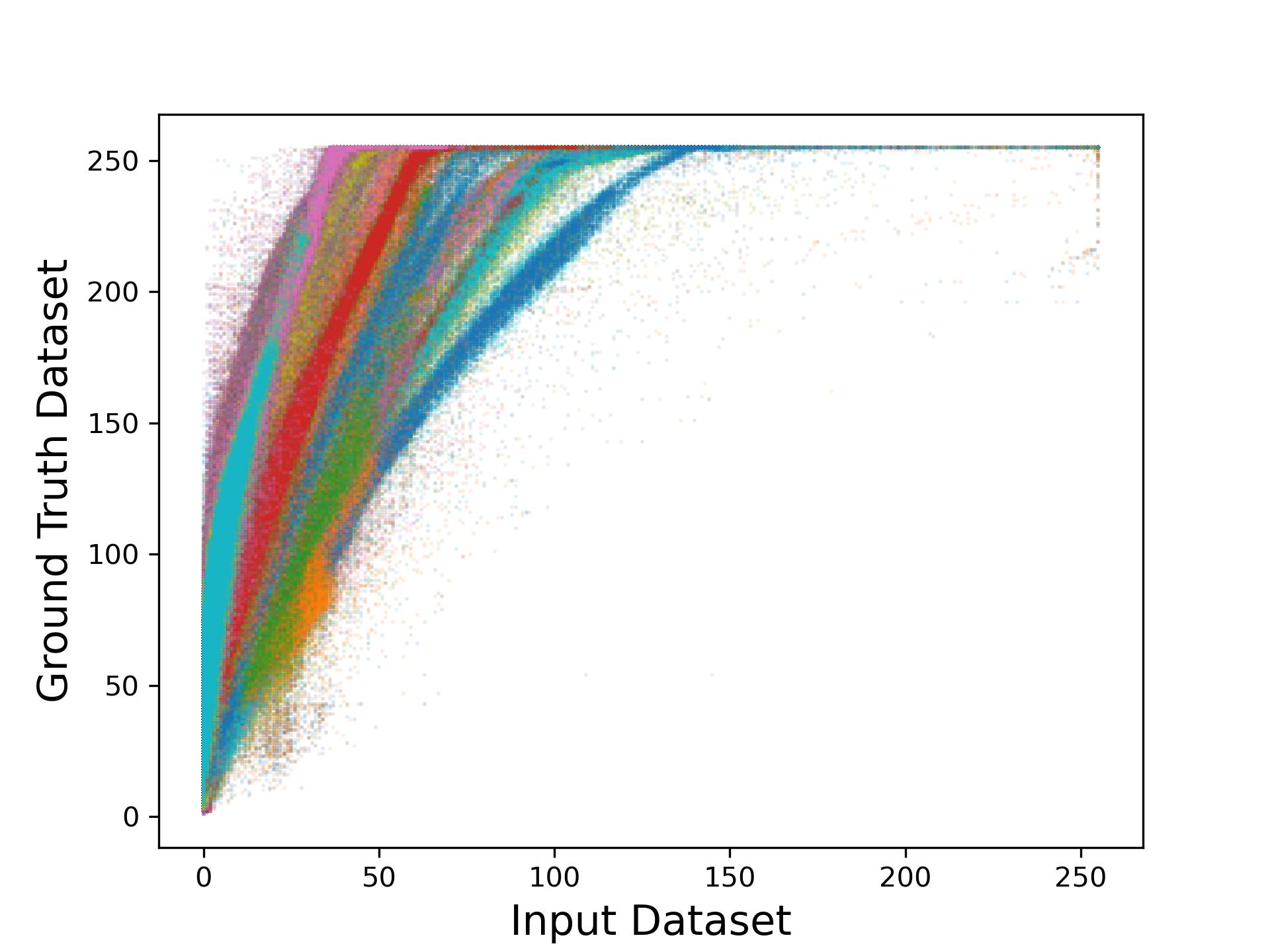}}  
    \\
    \subfloat[SICE (low)~\cite{cai2018learning}]{\includegraphics[width=0.5\columnwidth]{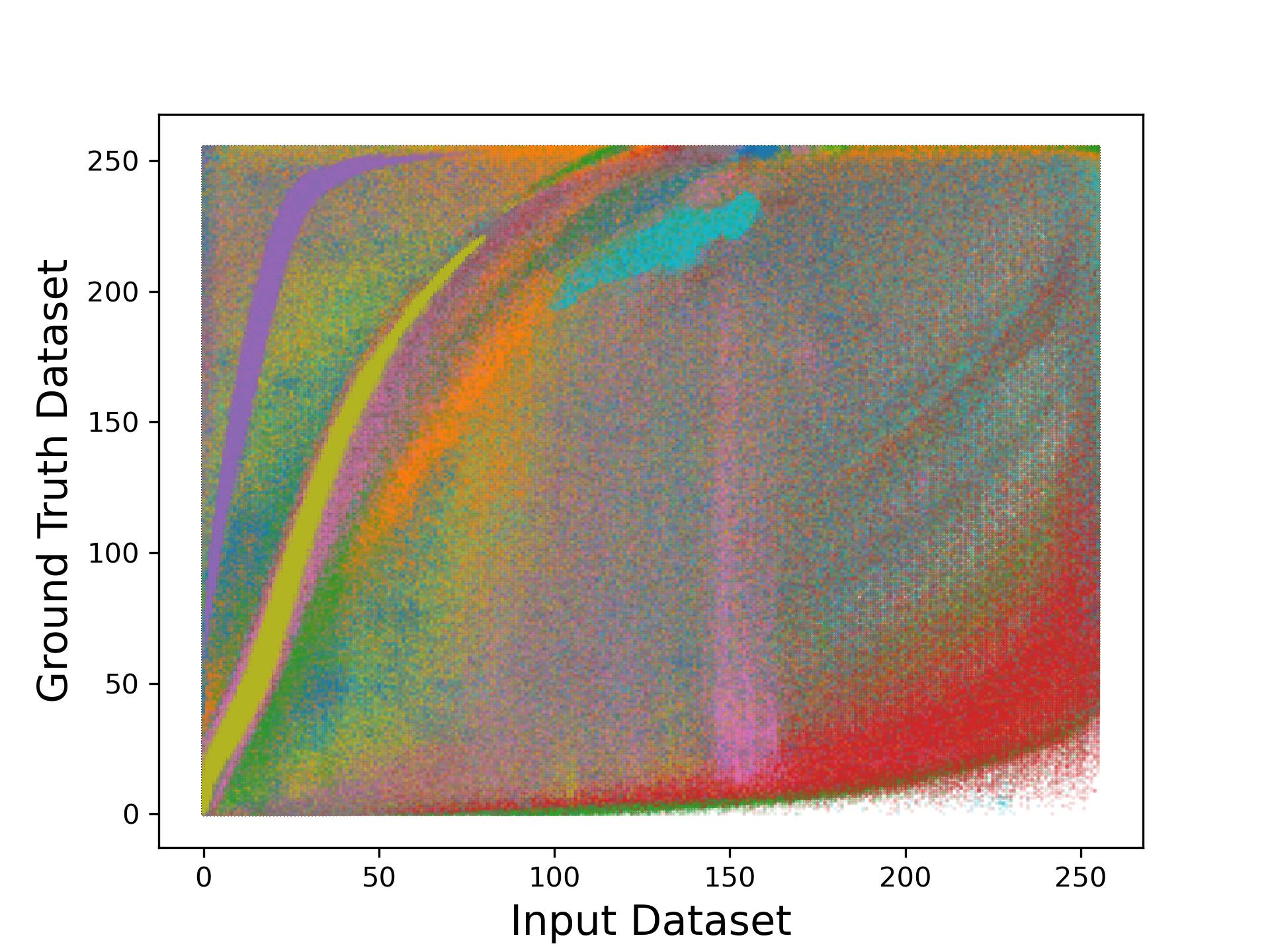}}
    \subfloat[SICE (high)~\cite{cai2018learning}]{\includegraphics[width=0.5\columnwidth]{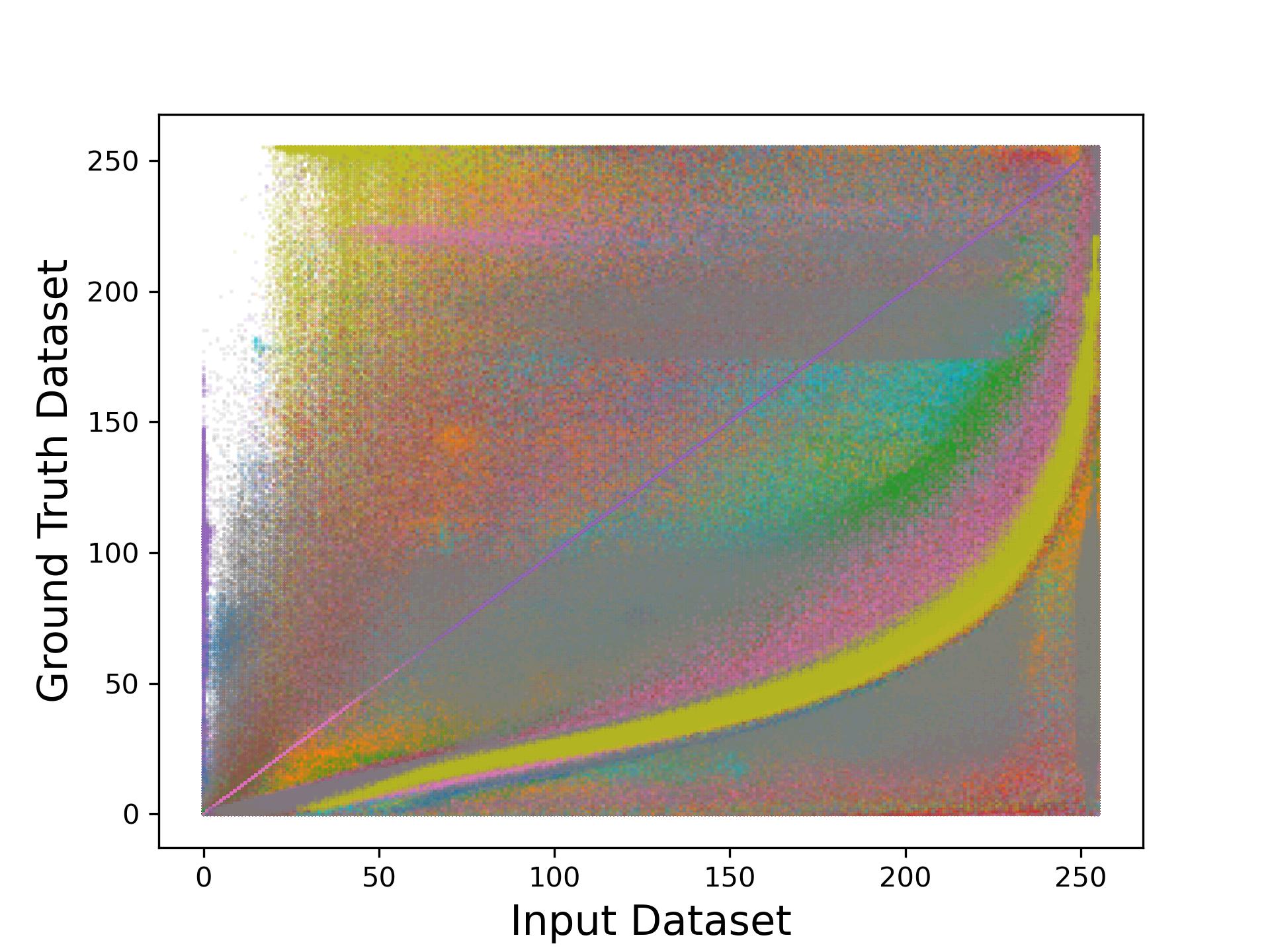}} 
    \\
    \subfloat[SICE\_Grad]{\includegraphics[width=0.5\columnwidth]{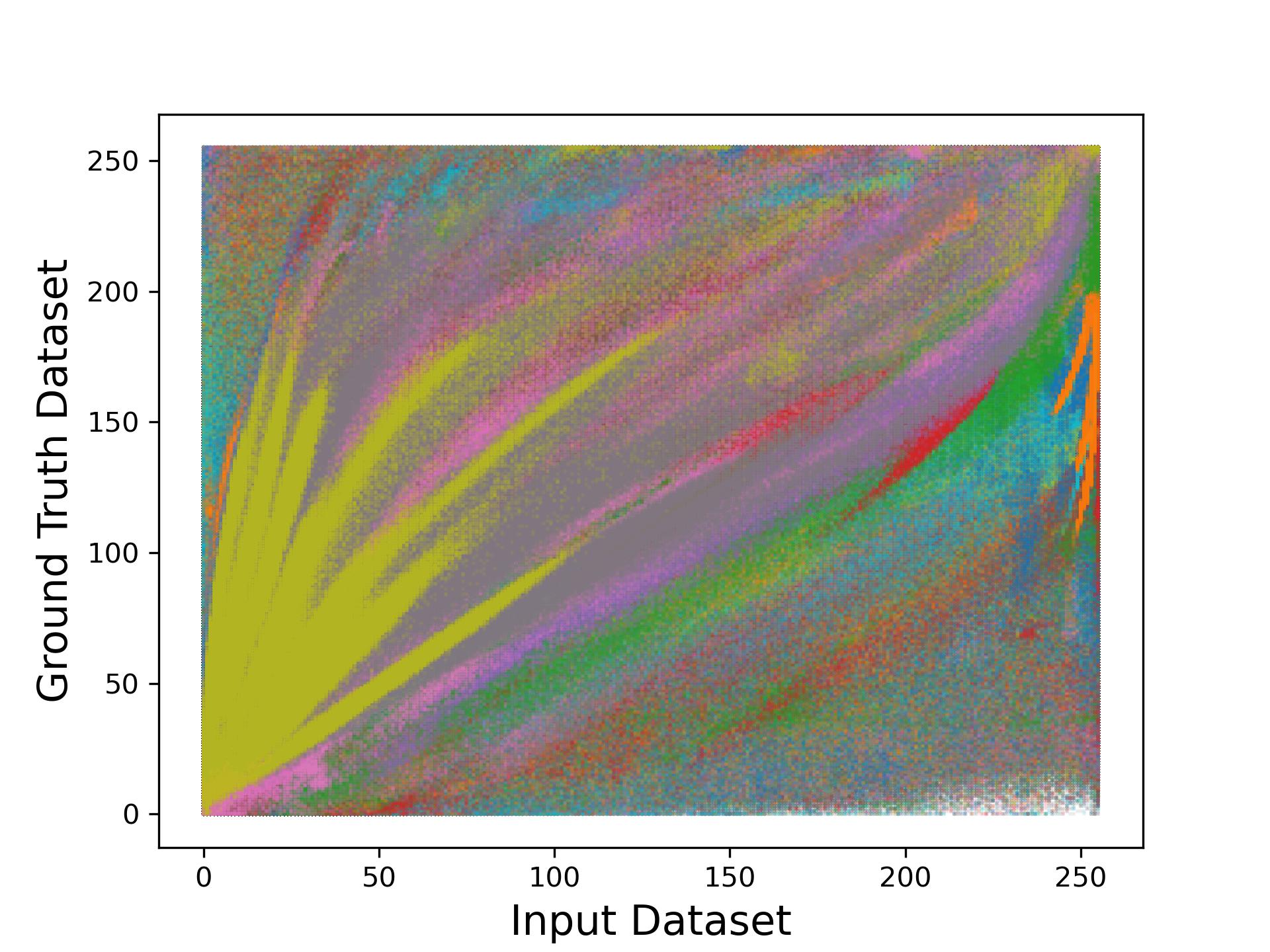}}
    \subfloat[SICE\_Mix]{\includegraphics[width=0.5\columnwidth]{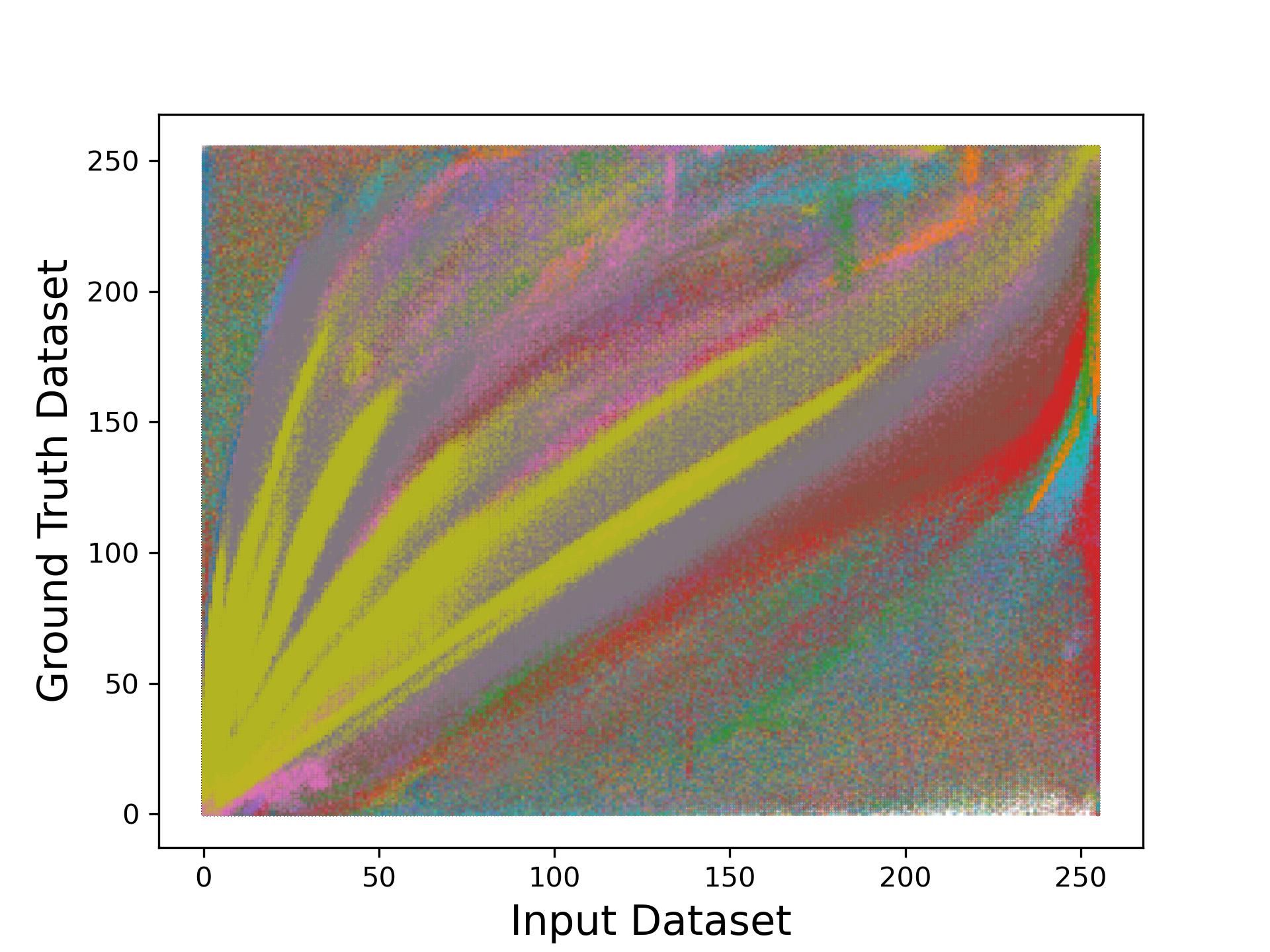}}
    \caption{\textbf{Over-/Underexposure Analysis on Paired Image Datasets using Pixel-to-Pixel Scatter Plots.} For each subfigure, the horizontal axis refers to the pixel value of the image from the input dataset, whereas the vertical axis refers to the ground truth dataset. A individual curve with the same color represents pixels from the same image. SICE\_Grad and SICE\_Mix are the only datasets that contain mixed overexposure and underexposure in a single image, since they feature individual curves with both concave and convex segments. See Section \ref{exposure_analysis} for details. }
    \label{fig:exposure}
\end{figure}

\subsection{New Image Dataset} \label{new_image_dataset}

We synthesize two new datasets, dubbed \textbf{SICE\_Grad} and \textbf{SICE\_Mix}, based on the SICE~\cite{cai2018learning} dataset. To obtain these two datasets, we first reshape the original SICE dataset to a resolution of 600 $\times$ 900. After that, we obtain panels from images in SICE with the same background but different exposures. The next step is different for SICE\_Grad and SICE\_Mix. For SICE\_Grad, we arrange the panels from low exposure to high exposure. To make it more challenging, we randomly placed some normally exposed panels at the end instead of at the mid. For SICE\_Mix, we permute all panels at random. Example images for the original SICE and the proposal SICE\_Grad and SICE\_Mix are in Fig. \ref{fig:SICE_Example}. The panel width is $\frac{1}{7}$ of the image width. The table summary for SICE\_Grad and SICE\_Mix is shown in Tab. \ref{tab:dataset}.

Our SICE\_Grad and SICE\_Mix dataset has two distinct advantages over existing datasets. (1)  SICE\_Grad and SICE\_Mix are synthesized through permuting the panels within the SICE dataset.   which enables their use in conjunction with the SICE dataset for training supervised learning methods. Meanwhile, they are suitable as testing datasets. (2) SICE\_Grad and SICE\_Mix exhibit extremely uneven exposure within individual images, a common occurrence in real-world scenarios like UAV.


\subsection{Exposure Analysis} \label{exposure_analysis}



    %

We conduct an exposure analysis to study the overexposure and underexposure for benchmark datasets with paired images. Specifically, we pick DCS~\cite{zheng2022semantic}, LOL~\cite{wei2018deep}, VE-LOL (Syn)~\cite{liu2021benchmarking}, VE-LOL (Real)~\cite{liu2021benchmarking}, and SICE~\cite{cai2018learning} to compare with the proposed SICE\_Grad and SICE\_Mix. 



Fig. \ref{fig:exposure} presents our exposure analysis using pixel-to-pixel scatter plots. The horizontal axis shows pixel values of input images, and the vertical axis represents ground truth images. Key characteristics include: (1) Each color-coded curve corresponds to an image pair. (2) Uniformly concave or convex plots indicate exclusively under or over-exposed input images. (3) Plots featuring both curve types signify a mix of under and over-exposure in the dataset. (4) Individual curves with both concave and convex segments suggest mixed exposure within single images.

We notice that DCS~\cite{zheng2022semantic}, LOL~\cite{wei2018deep}, VE-LOL (Syn)~\cite{liu2021benchmarking}, and VE-LOL (Real)~\cite{liu2021benchmarking} contain under-exposed images only. SICE (low-exposure) has a majority of under-exposed images and a minority of over-exposed images, whereas SICE (high-exposure) has a majority of over-exposed images and a minority of under-exposed images. SICE\_Grad and SICE\_Mix are unique because they not only have over-exposed and under-exposed images across the whole dataset but also mixed overexposure and underexposure in single images. These characteristics make SICE\_Grad and SICE\_Mix particularly challenging for image enhancement. Our experiments in Section \ref{sec:evaluations} show that no current representative LLIE method shows satisfactory result on SICE\_Grad or SICE\_Mix.

\subsection{New Video Dataset} \label{new_video_dataset}

We collect a large-scale dataset named \textbf{Night Wenzhou} to comprehensively analyze the performance of existing methods in real-world low-light conditions. In particular, our dataset contains aerial videos captured with DJI Mini 2 and streetscapes captured with GoPro HERO7 Silver. All videos are taken nighttime in Wenzhou, Zhejiang, China, and have an FPS of 30. The table summary for Night Wenzhou is displayed in Tab. \ref{tab:night_wenzhou}. As we can see, Night Wenzhou contains videos of 2 hours and 3 minutes and has a size of 26.144 GB.

\begin{figure}[t]
    \centering
    \includegraphics[width=0.32\columnwidth]{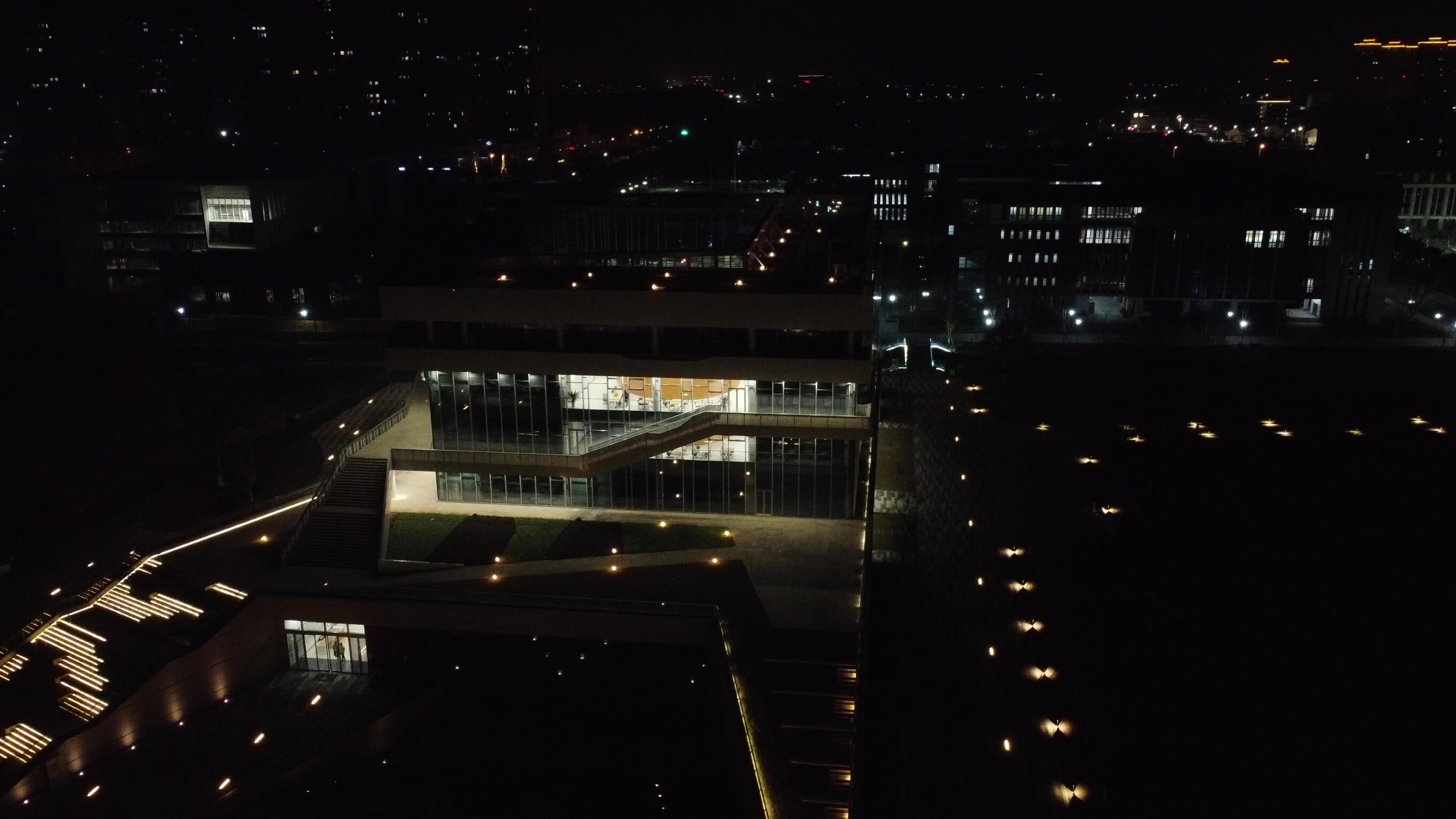}
    \includegraphics[width=0.32\columnwidth]{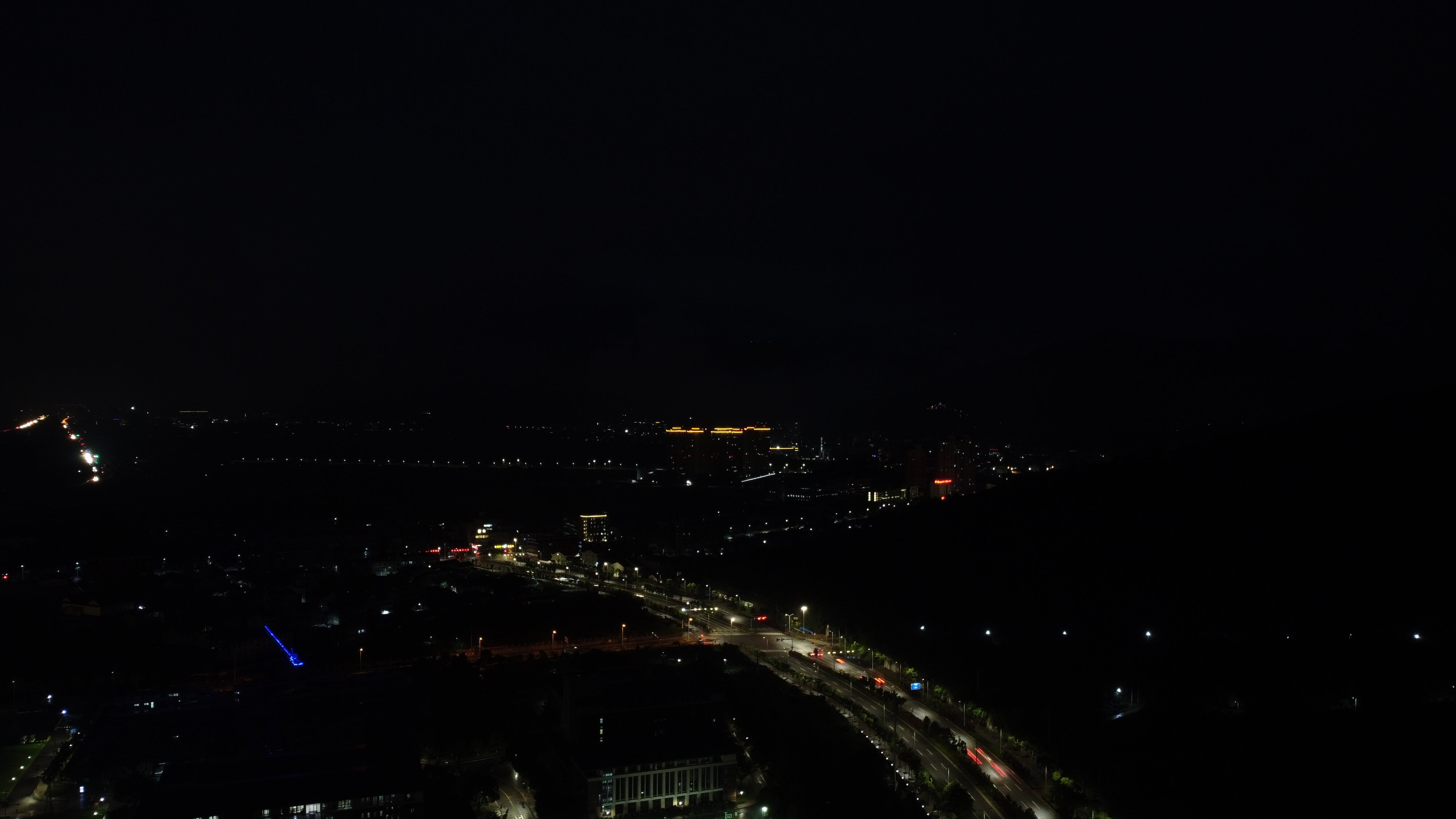}
    \includegraphics[width=0.32\columnwidth]{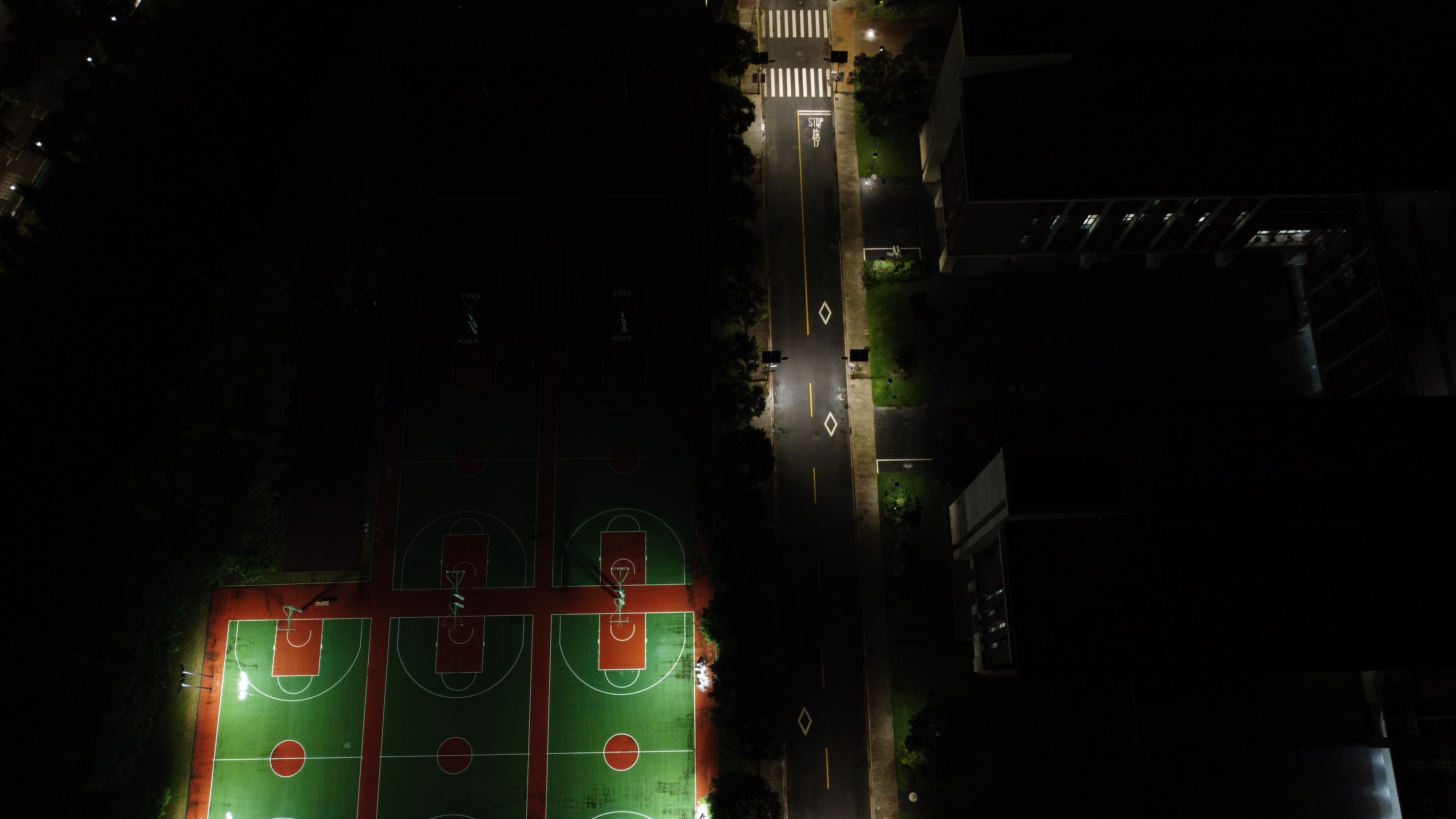} 
    \includegraphics[width=0.32\columnwidth]{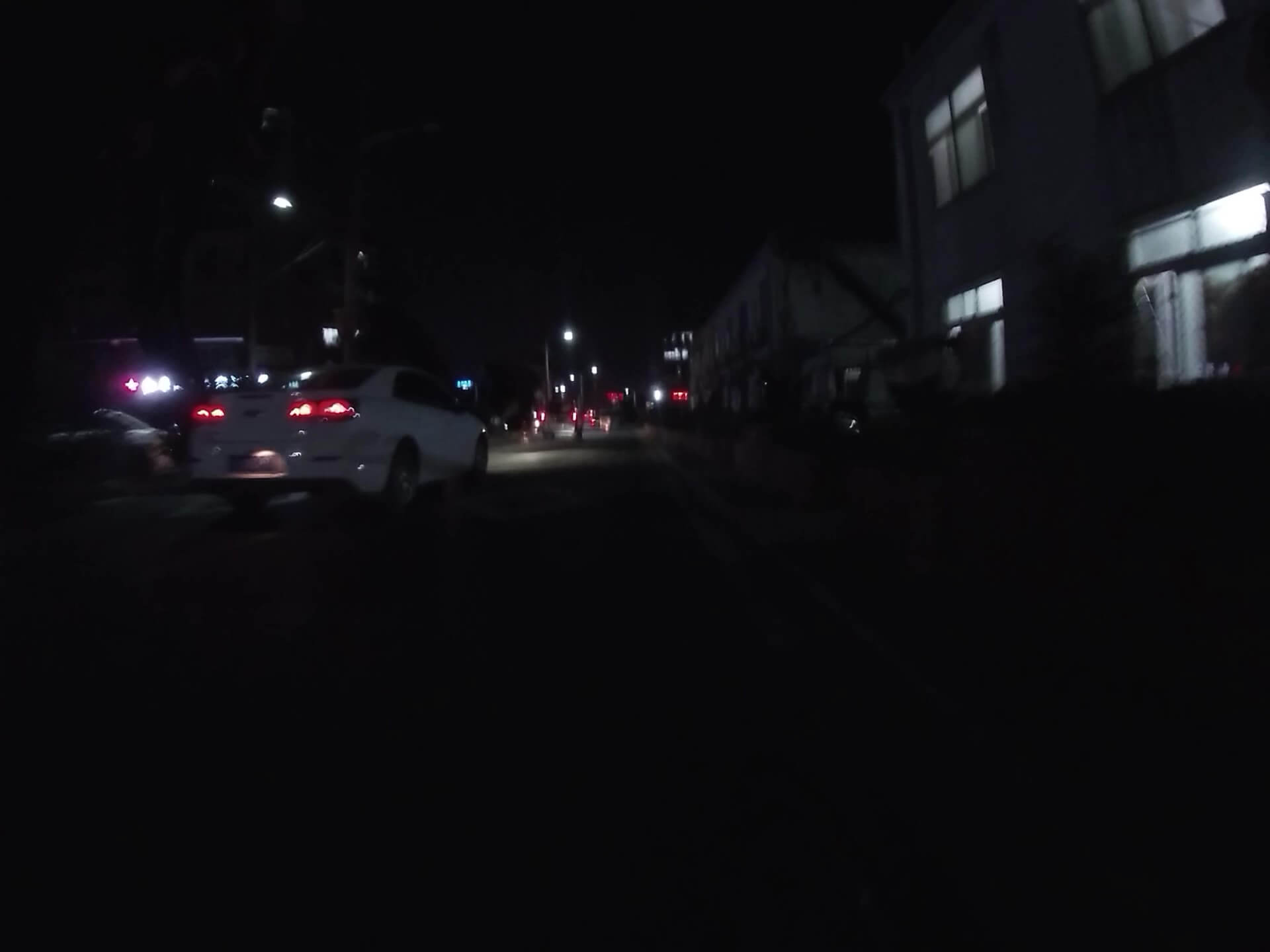}
    \includegraphics[width=0.32\columnwidth]{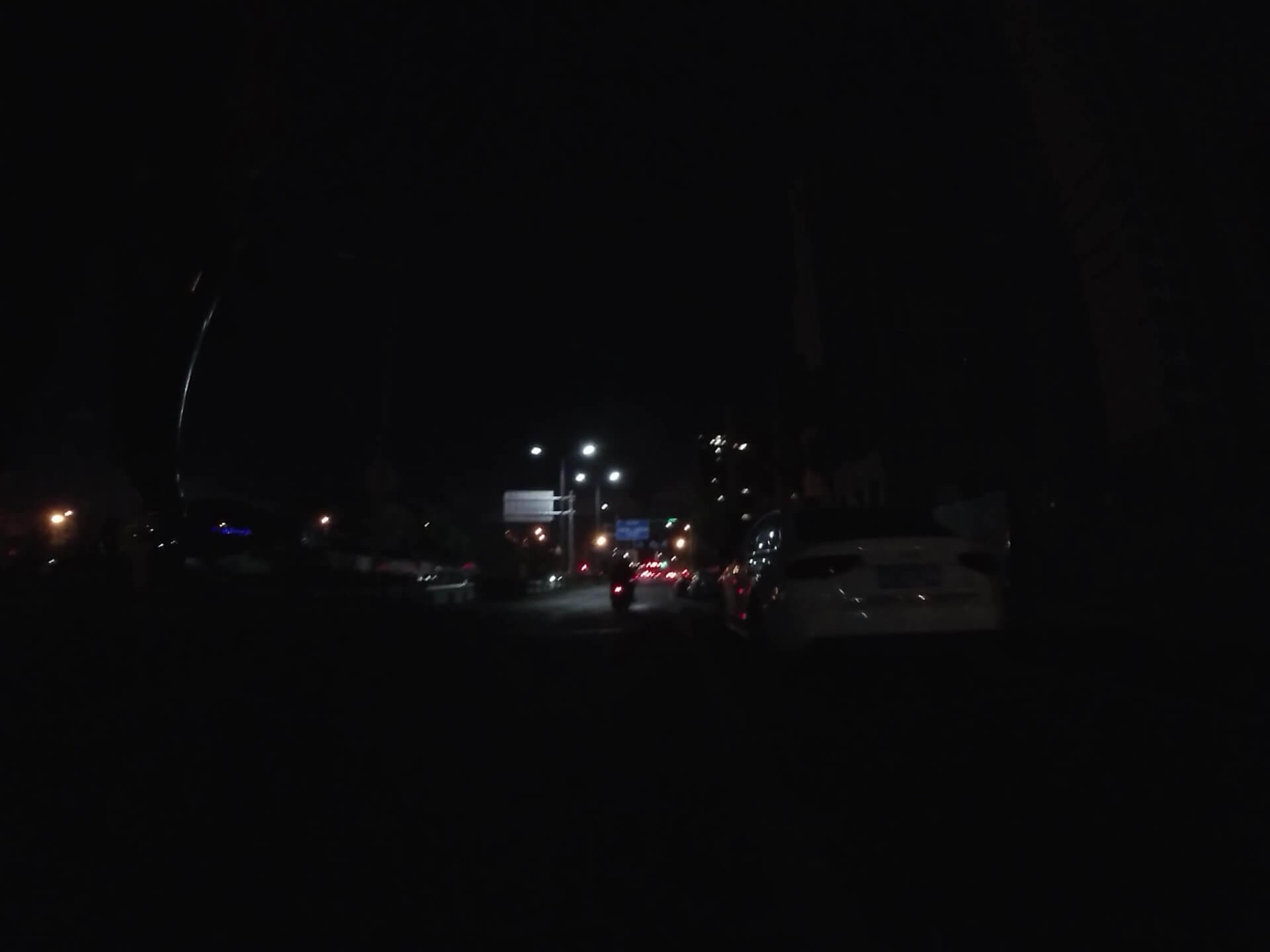}
    \includegraphics[width=0.32\columnwidth]{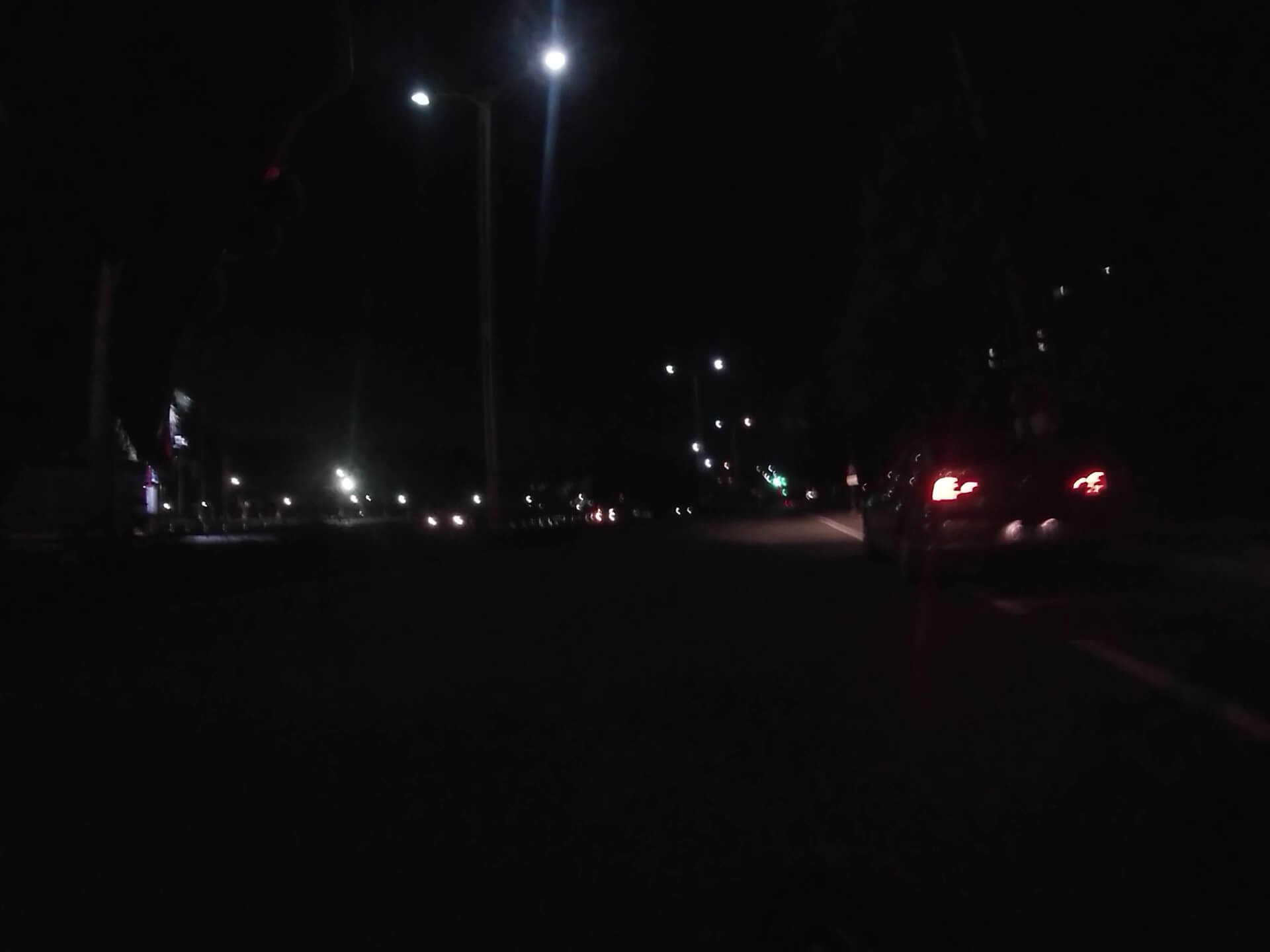} 
    \includegraphics[width=0.32\columnwidth]{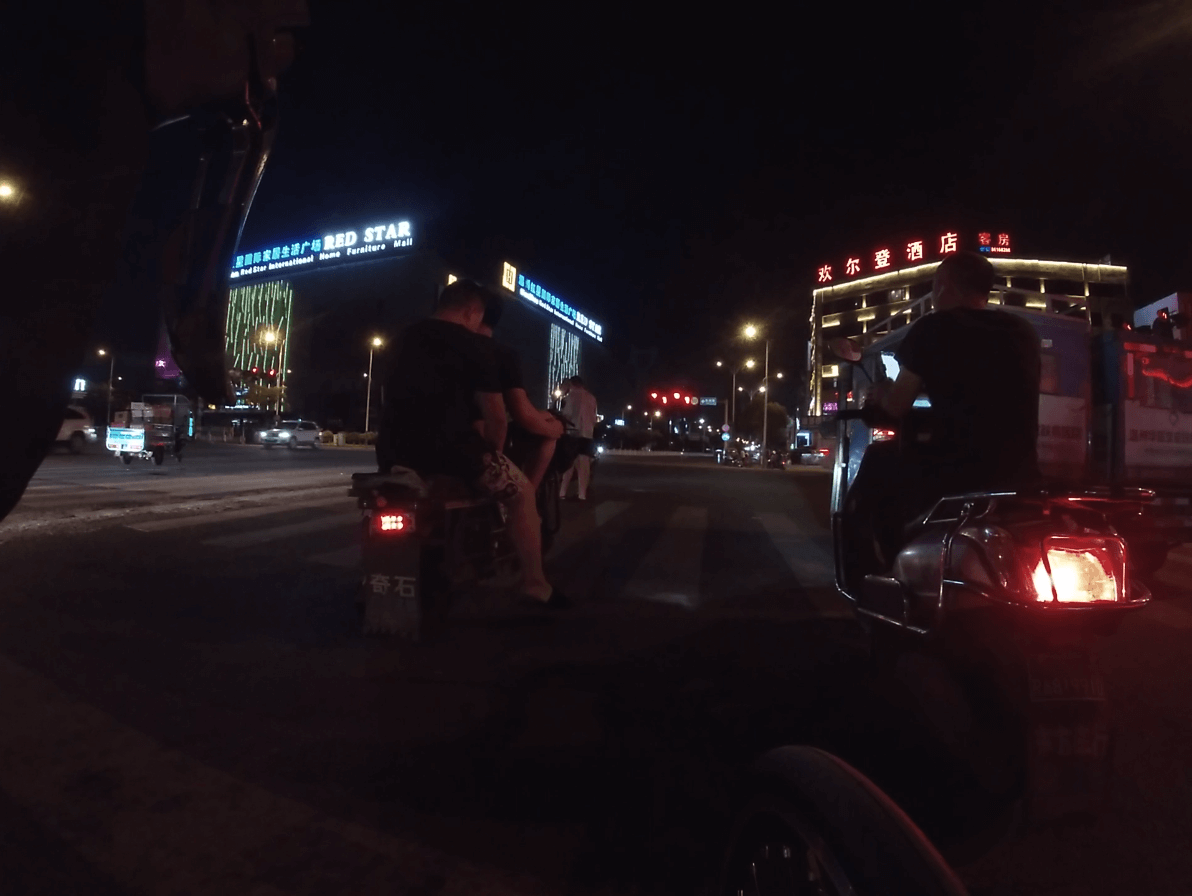}
    \includegraphics[width=0.32\columnwidth]{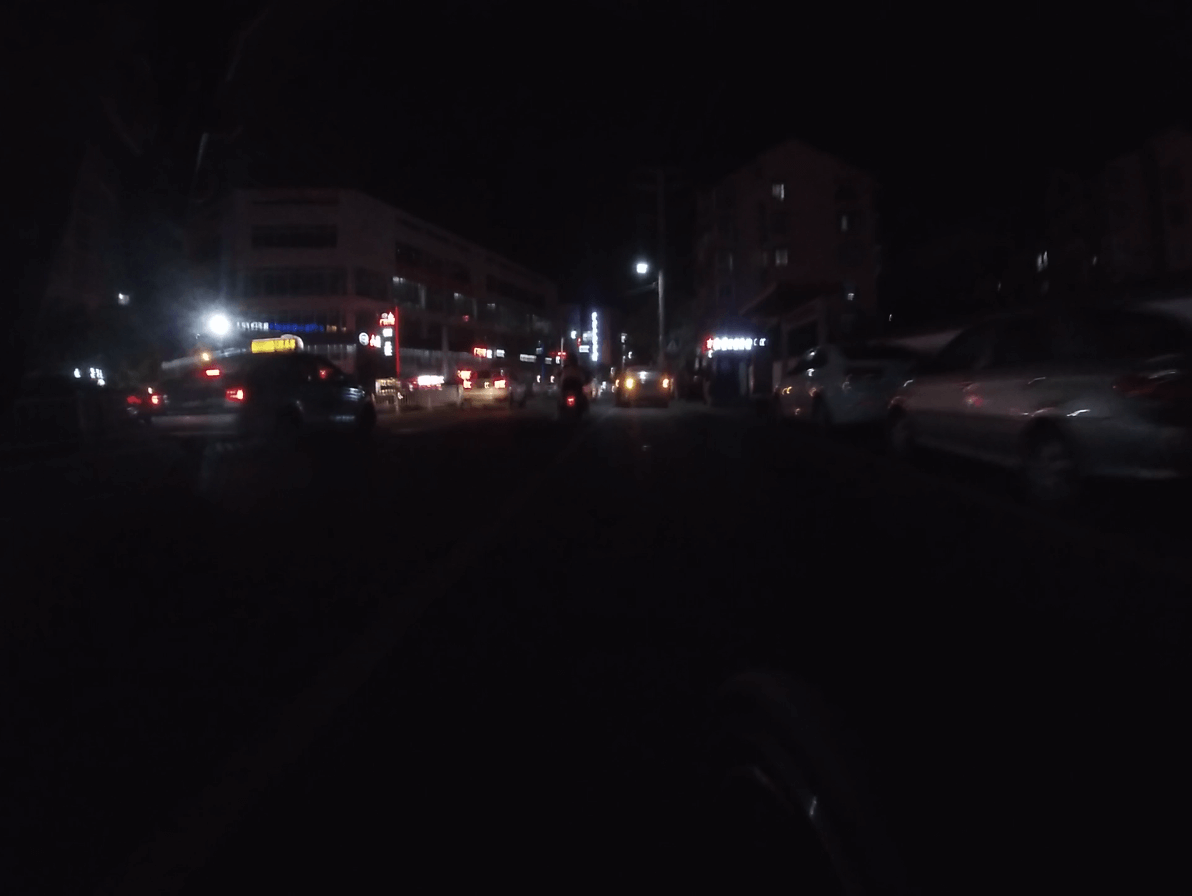}
    \includegraphics[width=0.32\columnwidth]{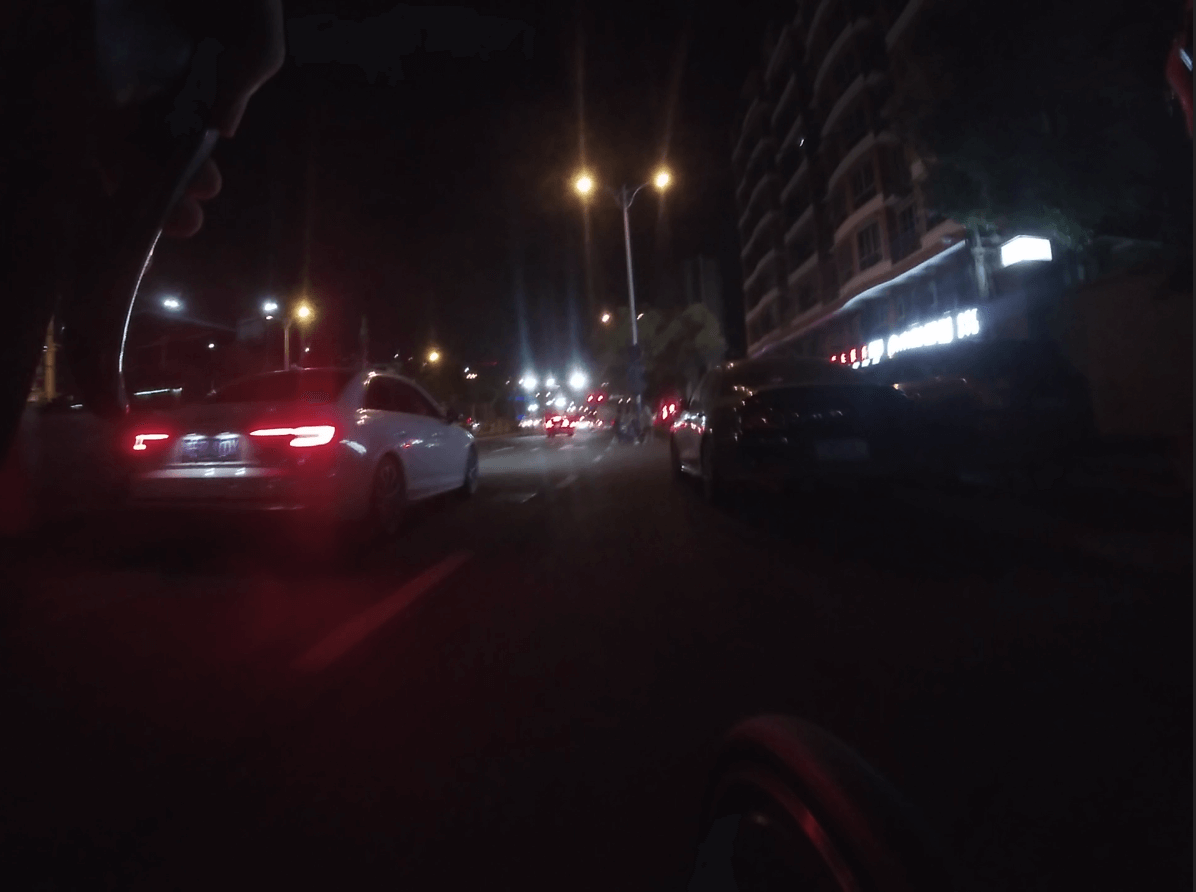}
   \caption{\textbf{Example Video Frames from the Night Wenzhou Dataset.} The Night Wenzhou Dataset features high-resolution videos captured under a wide range of night-time illumination conditions, including extreme darkness, underexposure, moonlit scenes, and uneven lighting. It also encompasses various types of degradation such as noise, blur, shadows, and artifacts. See Section \ref{new_video_dataset} for details. }
    \label{fig:night_wenzhou}
\end{figure}

Our Night Wenzhou dataset is challenging since (1) it contains large-scale high-resolution videos of diverse illumination conditions (e.g., extremely dark, underexposure, moonlight, uneven illumination, etc.). (2) it features various degradation (e.g., noise, blur, shadows, artifacts) commonly seen in real-world applications like autonomous driving. Our Night Wenzhou dataset can be used to train unsupervised and zero-shot learning methods and to test LLIE methods with any learning strategy. Sample images for our Night Wenzhou dataset are in Fig. \ref{fig:night_wenzhou}.

\begin{table}[t]
\centering
\caption{Table Summary for Night Wenzhou dataset. See Section \ref{new_video_dataset} for details. }
\setlength\tabcolsep{4pt}
\begin{tabular}{l|c|r|r}
\toprule
\textbf{Device}      & \textbf{Resolution (H $\times$ W)} & \textbf{Duration (h:m:s)} & \textbf{Size (GB)} \\ \hline
\multirow{9}{*}{GoPro}       & \multirow{9}{*}{1440 $\times$ 1920}     & 0:09:08           & 1.871     \\ 
       &      & 0:13:04          & 2.675     \\ 
       &      & 0:17:41          & 3.727     \\ 
       &     & 0:17:40          & 3.727     \\ 
       &      & 0:17:43          & 3.727     \\ 
       &     & 0:11:15          & 2.316     \\ 
       &     & 0:05:14           & 1.029     \\ 
       &     & 0:17:57          & 3.727     \\ 
       &     & 0:02:25           & 0.526     \\  \hline
\textbf{GoPro Total} &                 & \textbf{1:52:07}          & \textbf{23.325}    \\ \hline\hline
\multirow{7}{*}{DJI}          & \multirow{7}{*}{1530 $\times$ 2720}       & 0:01:47           & 0.500     \\ 
         &     & 0:00:27           & 0.126     \\ 
         &      & 0:00:42           & 0.198     \\ 
         &     & 0:00:42           & 0.177     \\ 
         &    & 0:00:27           & 0.127     \\ 
         &     & 0:06:21           & 1.604     \\ 
         &     & 0:00:27           & 0.087     \\  \hline
\textbf{DJI Total}   &                 &  \textbf{0:10:53}         & \textbf{2.819}     \\ \hline\hline
\textbf{Total }      &                 &  \textbf{2:03:00}         & \textbf{26.144}     \\ \bottomrule 
\end{tabular}
\label{tab:night_wenzhou}
\end{table}

\section{Evaluations} \label{sec:evaluations}

\subsection{Quantitative Comparisons}





In this subsection, we leverage full-reference metrics including PSNR, SSIM, and LPIPS~\cite{dee2007teachers}, and non-reference metrics including UNIQUE~\cite{zhang2021uncertainty}, BRISQUE~\cite{mittal2012no}, and SPAQ~\cite{fang2020perceptual}. Note that some entries are `-' due to the out-of-memory error during inference. 

\noindent \textbf{Results on Synthetic Datasets:}
Tab. \ref{tab:quant1} shows the quantitative comparison on LOL~\cite{wei2018deep}, DCS~\cite{zheng2022semantic}, VE-LOL~\cite{liu2021benchmarking}, SICE\_Mix, and SICE\_Grad. We observe that LLFlow~\cite{wang2022low} demonstrates superior performance: it achieves the best PSNR, SSIM, and LPIPS on LOL~\cite{wei2018deep} and VE-LOL (Real)~\cite{liu2021benchmarking}, and the best PSNR and SSIM on DCS~\cite{zheng2022semantic}. KinD++~\cite{zhang2021beyond} exhibits excellent performance with the best PSNR, SSIM, and LPIPS on VE-LOL (Syn) and SICE\_Mix, and the best PSNR and LPIPS on SICE\_Grad. Furthermore, Zero-DCE~\cite{guo2020zero} has the best LPIPS on DCS~\cite{zheng2022semantic}, whereas KinD~\cite{zhang2019kindling} has the best SSIM on SICE\_Grad. No other methods achieve the best score at any metrics. 

\noindent \textbf{Results on Simple Real-World Datasets:}
Tab. \ref{tab:quant2} displays the quantitative comparison on NPE~\cite{wang2013naturalness}, LIME~\cite{guo2016lime}, MEF~\cite{ma2015perceptual}, DICM~\cite{lee2013contrast}, and VV. We note that the competition for Tab. \ref{tab:quant2} is much more intense than Tab. \ref{tab:quant1}. The only method with 4 best scores is Zero-DCE~\cite{guo2020zero}, whereas the only method with 3 best scores is KinD++~\cite{zhang2021beyond}. Besides, the methods with 2 best scores are RetinexNet~\cite{wei2018deep} and SGZ~\cite{zheng2022semantic}. Furthermore, the methods with only 1 best score include KinD~\cite{zhang2019kindling}, LLFlow~\cite{wang2022low}, URetinexNet~\cite{wu2022uretinex}, and SCI~\cite{ma2022toward}. RUAS~\cite{liu2021retinex} is the only method that does not perform the best at any metric. 

\noindent \textbf{Results on Complex Real-World Datasets:}
Tab. \ref{tab:quant3} presents the quantitative comparison on DarkFace~\cite{yang2020advancing} and ExDark~\cite{loh2019getting}. We notice that RUAS~\cite{liu2021retinex} earns the best UNIQUE and BRISQUE on DarkFace~\cite{yang2020advancing}, whereas RetinexNet~\cite{wei2018deep} scores the best SPAQ on DarkFace~\cite{yang2020advancing} and ExDark~\cite{loh2019getting}. Additionally, LLFlow~\cite{wang2022low} obtains the best UNIQUE on ExDark~\cite{loh2019getting}. 

\noindent \textbf{Results for Model Efficiency:}
 Tab. \ref{tab:eff} presents the quantitative comparison of model efficiency. We choose ACDC~\cite{sakaridis2021acdc} as the benchmark for efficiency comparison since it contains images of 2K resolution (i.e., 1080 $\times$ 1920), which is closer to real-world applications such as autonomous driving, UAV, and photography. It is shown that SGZ~\cite{zheng2022semantic} obtains the best FLOPs and Inference Time, whereas SCI~\cite{ma2022toward} gains the best \#Params. Besides, it is worth mentioning Zero-DCE~\cite{guo2020zero}, RUAS~\cite{liu2021retinex}, SGZ~\cite{zheng2022semantic}, and SCI~\cite{ma2022toward} achieve real-time processing on a single GPU. 

\noindent \textbf{Results for Semantic Segmentation:}
 Tab. \ref{tab:sem} shows the quantitative comparison of semantic segmentation. For both ACDC~\cite{sakaridis2021acdc} and DCS~\cite{zheng2022semantic}, we feed the enhanced image into a semantic segmentation model named PSPNet~\cite{zhao2017pyramid} and calculate the mPA and mIoU score with the default thresholds. On ACDC~\cite{sakaridis2021acdc}, LLFlow~\cite{wang2022low} leads to the best mPA, whereas SCI~\cite{ma2022toward} results in the best mIoU. On DCS~\cite{zheng2022semantic}, SGZ~\cite{zheng2022semantic} contributes to the best mPA and mIoU. 

\noindent \textbf{Results for Object Detection:}
 Tab. \ref{tab:det} displays the quantitative comparison of object detection. In particular, we feed the enhanced image into a face detection model named DSFD~\cite{li2019dsfd} and calculate the IoU score with different IoU thresholds (0.5, 0.6, and 0.7). It is shown that LLFlow~\cite{wang2022low} yields the best IoU with all given thresholds. 

\begin{table*}[t]
\centering
\setlength\tabcolsep{1.0pt}
\scriptsize
\caption{Quantitative Comparison on LOL~\cite{wei2018deep}, DCS~\cite{zheng2022semantic}, VE-LOL (Synthetic \& Real)~\cite{liu2021benchmarking}, SICE\_Mix, and SICE\_Grad with PSNR, SSIM, and LPIPS.}
\begin{tabular}{l|ccc|ccc|ccc|ccc|ccc|ccc}
\toprule
\multirow{2}{*}{\textbf{Method}} & \multicolumn{3}{c|}{\textbf{LOL}~\cite{wei2018deep}}                                                                    & \multicolumn{3}{c|}{\textbf{DCS}~\cite{zheng2022semantic}}                                                                    & \multicolumn{3}{c|}{\textbf{VE-LOL (Syn)}~\cite{liu2021benchmarking}}                                                           & \multicolumn{3}{c|}{\textbf{VE-LOL (Real)}~\cite{liu2021benchmarking}}                                                          & \multicolumn{3}{c|}{\textbf{SICE\_Mix}}                                                               & \multicolumn{3}{c}{\textbf{SICE\_Grad}}                                                              \\ \cline{2-19} 
                         & \multicolumn{1}{c}{\textbf{PSNR$\uparrow$}}            & \multicolumn{1}{c}{\textbf{SSIM$\uparrow$}}           & \textbf{LPIPS$\downarrow$}          & \multicolumn{1}{c}{\textbf{PSNR$\uparrow$}}            & \multicolumn{1}{c}{\textbf{SSIM$\uparrow$}}           & \textbf{LPIPS$\downarrow$}          & \multicolumn{1}{c}{\textbf{PSNR$\uparrow$}}            & \multicolumn{1}{c}{\textbf{SSIM$\uparrow$}}           & \textbf{LPIPS$\downarrow$}          & \multicolumn{1}{c}{\textbf{PSNR$\uparrow$}}            & \multicolumn{1}{c}{\textbf{SSIM$\uparrow$}}           & \textbf{LPIPS$\downarrow$}          & \multicolumn{1}{c}{\textbf{PSNR$\uparrow$}}            & \multicolumn{1}{c}{\textbf{SSIM$\uparrow$}}           & \textbf{LPIPS$\downarrow$}          & \multicolumn{1}{c}{\textbf{PSNR$\uparrow$}}            & \multicolumn{1}{c}{\textbf{SSIM$\uparrow$}}           & \textbf{LPIPS$\downarrow$}          \\ \hline
RetinexNet \cite{wei2018deep}              & \multicolumn{1}{c}{17.559}          & \multicolumn{1}{c}{0.645}          & 0.381          & \multicolumn{1}{c}{-}             & \multicolumn{1}{c}{-}            & -            & \multicolumn{1}{c}{15.606}          & \multicolumn{1}{c}{0.449}          & 0.769          & \multicolumn{1}{c}{17.676}          & \multicolumn{1}{c}{0.642}          & 0.441          & \multicolumn{1}{c}{12.397}          & \multicolumn{1}{c}{0.606}          & 0.407          & \multicolumn{1}{c}{12.450}          & \multicolumn{1}{c}{0.619}          & 0.364          \\ 
KinD \cite{zhang2019kindling}                    & \multicolumn{1}{c}{15.867}          & \multicolumn{1}{c}{0.637}          & 0.341          & \multicolumn{1}{c}{13.145}          & \multicolumn{1}{c}{0.720}          & 0.304          & \multicolumn{1}{c}{16.259}          & \multicolumn{1}{c}{0.591}          & 0.432          & \multicolumn{1}{c}{20.588}          & \multicolumn{1}{c}{0.818}          & 0.143          & \multicolumn{1}{c}{12.986}          & \multicolumn{1}{c}{0.656}          & 0.346          & \multicolumn{1}{c}{13.144}          & \multicolumn{1}{c}{\textbf{0.668}} & 0.302          \\
Zero-DCE \cite{guo2020zero}                  & \multicolumn{1}{c}{14.861}          & \multicolumn{1}{c}{0.562}          & 0.330          & \multicolumn{1}{c}{16.224}          & \multicolumn{1}{c}{0.849}          & \textbf{0.172} & \multicolumn{1}{c}{14.071}          & \multicolumn{1}{c}{0.369}          & 0.652          & \multicolumn{1}{c}{18.059}          & \multicolumn{1}{c}{0.580}          & 0.308          & \multicolumn{1}{c}{12.428}          & \multicolumn{1}{c}{0.633}          & 0.362          & \multicolumn{1}{c}{12.475}          & \multicolumn{1}{c}{0.644}          & 0.314          \\ 
KinD++ \cite{zhang2021beyond}                  & \multicolumn{1}{c}{15.724}          & \multicolumn{1}{c}{0.621}          & 0.363          & \multicolumn{1}{c}{-}             & \multicolumn{1}{c}{-}            & -           & \multicolumn{1}{c}{\textbf{16.523}} & \multicolumn{1}{c}{\textbf{0.613}} & \textbf{0.411} & \multicolumn{1}{c}{17.660}          & \multicolumn{1}{c}{0.761}          & 0.218          & \multicolumn{1}{c}{\textbf{13.196}} & \multicolumn{1}{c}{\textbf{0.657}} & \textbf{0.334} & \multicolumn{1}{c}{\textbf{13.235}} & \multicolumn{1}{c}{0.666}          & \textbf{0.295} \\ 
RUAS \cite{liu2021retinex}                    & \multicolumn{1}{c}{11.309}          & \multicolumn{1}{c}{0.435}          & 0.377          & \multicolumn{1}{c}{11.601}          & \multicolumn{1}{c}{0.412}          & 0.449          & \multicolumn{1}{c}{12.386}          & \multicolumn{1}{c}{0.357}          & 0.642          & \multicolumn{1}{c}{13.975}          & \multicolumn{1}{c}{0.469}          & 0.329          & \multicolumn{1}{c}{8.684}           & \multicolumn{1}{c}{0.493}          & 0.525          & \multicolumn{1}{c}{8.628}           & \multicolumn{1}{c}{0.494}          & 0.499          \\ 
SGZ \cite{zheng2022semantic}                     & \multicolumn{1}{c}{15.345}          & \multicolumn{1}{c}{0.573}          & 0.334          & \multicolumn{1}{c}{16.369}          & \multicolumn{1}{c}{0.854}          & 0.204          & \multicolumn{1}{c}{13.830}          & \multicolumn{1}{c}{0.385}          & 0.664          & \multicolumn{1}{c}{18.582}          & \multicolumn{1}{c}{0.584}          & 0.309          & \multicolumn{1}{c}{10.866}          & \multicolumn{1}{c}{0.607}          & 0.415          & \multicolumn{1}{c}{10.987}          & \multicolumn{1}{c}{0.621}          & 0.364          \\ 
LLFlow \cite{wang2022low}                  & \multicolumn{1}{c}{\textbf{19.341}} & \multicolumn{1}{c}{\textbf{0.839}} & \textbf{0.142} & \multicolumn{1}{c}{\textbf{20.385}} & \multicolumn{1}{c}{\textbf{0.897}} & 0.240          & \multicolumn{1}{c}{15.440}          & \multicolumn{1}{c}{0.476}          & 0.517          & \multicolumn{1}{c}{\textbf{24.152}} & \multicolumn{1}{c}{\textbf{0.895}} & \textbf{0.098} & \multicolumn{1}{c}{12.737}          & \multicolumn{1}{c}{0.617}          & 0.388          & \multicolumn{1}{c}{12.737}          & \multicolumn{1}{c}{0.617}          & 0.388          \\ 
URetinexNet \cite{wu2022uretinex}             & \multicolumn{1}{c}{17.278}          & \multicolumn{1}{c}{0.688}          & 0.302          & \multicolumn{1}{c}{16.009}          & \multicolumn{1}{c}{0.755}          & 0.369          & \multicolumn{1}{c}{15.273}          & \multicolumn{1}{c}{0.466}          & 0.591          & \multicolumn{1}{c}{21.093}          & \multicolumn{1}{c}{0.858}          & 0.103          & \multicolumn{1}{c}{10.903}          & \multicolumn{1}{c}{0.600}          & 0.402          & \multicolumn{1}{c}{10.894}          & \multicolumn{1}{c}{0.610}          & 0.356          \\ 
SCI\cite{ma2022toward}                     & \multicolumn{1}{c}{14.784}          & \multicolumn{1}{c}{0.525}          & 0.333          & \multicolumn{1}{c}{14.264}          & \multicolumn{1}{c}{0.689}          & 0.249          & \multicolumn{1}{c}{12.542}          & \multicolumn{1}{c}{0.373}          & 0.681          & \multicolumn{1}{c}{17.304}          & \multicolumn{1}{c}{0.540}          & 0.307          & \multicolumn{1}{c}{8.644}           & \multicolumn{1}{c}{0.529}          & 0.511          & \multicolumn{1}{c}{8.559}           & \multicolumn{1}{c}{0.532}          & 0.484          \\ \bottomrule
\end{tabular}
\label{tab:quant1}
\end{table*}

\begin{table*}[t]
\setlength\tabcolsep{3.5pt}
\centering
\caption{Quantitative Comparison on NPE~\cite{wang2013naturalness}, LIME~\cite{guo2016lime}, MEF~\cite{ma2015perceptual}, DICM~\cite{lee2013contrast}, and VV with UNIQUE (UNI), BRISQUE (BRI), and SPAQ.}
\begin{tabular}{l|ccc|ccc|ccc|ccc|ccc}
\toprule
\multirow{2}{*}{\textbf{Method}} & \multicolumn{3}{c|}{\textbf{NPE}~\cite{wang2013naturalness}}                                                                     & \multicolumn{3}{c|}{\textbf{LIME}~\cite{guo2016lime}}                                                                    & \multicolumn{3}{c|}{\textbf{MEF}~\cite{ma2015perceptual}}                                                                     & \multicolumn{3}{c|}{\textbf{DICM}~\cite{lee2013contrast}}                                                                    & \multicolumn{3}{c}{\textbf{VV} }                                                                      \\ \cline{2-16} 
                         & \multicolumn{1}{c}{\textbf{UNI$\uparrow$}}         & \multicolumn{1}{c}{\textbf{BRI$\downarrow$}}         & \textbf{SPAQ$\uparrow$}            & \multicolumn{1}{c}{\textbf{UNI$\uparrow$}}         & \multicolumn{1}{c}{\textbf{BRI$\downarrow$}}         & \textbf{SPAQ$\uparrow$}            & \multicolumn{1}{c}{\textbf{UNI$\uparrow$}}         & \multicolumn{1}{c}{\textbf{BRI$\downarrow$}}         & \textbf{SPAQ$\uparrow$}            & \multicolumn{1}{c}{\textbf{UNI$\uparrow$}}         & \multicolumn{1}{c}{\textbf{BRI$\downarrow$}}         & \textbf{SPAQ$\uparrow$}            & \multicolumn{1}{c}{\textbf{UNI$\uparrow$}}         & \multicolumn{1}{c}{\textbf{BRI$\downarrow$}}         & \textbf{SPAQ$\uparrow$}            \\ \hline
RetinexNet~\cite{wei2018deep}               & \multicolumn{1}{c}{0.801}          & \multicolumn{1}{c}{16.533}          & 71.264          & \multicolumn{1}{c}{0.787}          & \multicolumn{1}{c}{24.310}           & 70.468          & \multicolumn{1}{c}{0.742}          & \multicolumn{1}{c}{14.583}          & \textbf{69.333} & \multicolumn{1}{c}{0.778}          & \multicolumn{1}{c}{\textbf{22.877}} & 62.550          & \multicolumn{1}{c}{-}            & \multicolumn{1}{c}{-}             & -             \\ 
KinD~\cite{zhang2019kindling}                     & \multicolumn{1}{c}{0.792}          & \multicolumn{1}{c}{20.239}          & 70.444          & \multicolumn{1}{c}{0.766}          & \multicolumn{1}{c}{39.783} & 67.180          & \multicolumn{1}{c}{0.747}          & \multicolumn{1}{c}{32.019}          & 63.266          & \multicolumn{1}{c}{0.776}          & \multicolumn{1}{c}{33.092}          & 59.946          & \multicolumn{1}{c}{0.814}          & \multicolumn{1}{c}{29.439}          & \textbf{61.453} \\ 
Zero-DCE~\cite{guo2020zero}                  & \multicolumn{1}{c}{\textbf{0.814}} & \multicolumn{1}{c}{17.456}          & \textbf{72.945} & \multicolumn{1}{c}{0.811}          & \multicolumn{1}{c}{20.437}          & 67.736 & \multicolumn{1}{c}{\textbf{0.762}} & \multicolumn{1}{c}{17.321}          & 66.864          & \multicolumn{1}{c}{0.777}          & \multicolumn{1}{c}{27.560}          & 57.402          & \multicolumn{1}{c}{\textbf{0.836}} & \multicolumn{1}{c}{34.656}          & 60.716          \\ 
KinD++~\cite{zhang2021beyond}                   & \multicolumn{1}{c}{0.801} & \multicolumn{1}{c}{19.507} & 71.742 & \multicolumn{1}{c}{0.748} & \multicolumn{1}{c}{\textbf{19.954}} & \textbf{73.414} & \multicolumn{1}{c}{0.732}          & \multicolumn{1}{c}{27.781}          & 67.831          & \multicolumn{1}{c}{0.774}          & \multicolumn{1}{c}{27.573}          & \textbf{62.744} & \multicolumn{1}{c}{-}            & \multicolumn{1}{c}{-}             & -             \\ 
RUAS~\cite{liu2021retinex}                    & \multicolumn{1}{c}{0.706}          & \multicolumn{1}{c}{47.852}          & 61.598          & \multicolumn{1}{c}{0.783}          & \multicolumn{1}{c}{27.589}          & 62.076          & \multicolumn{1}{c}{0.713}          & \multicolumn{1}{c}{23.677}          & 60.701          & \multicolumn{1}{c}{0.710}          & \multicolumn{1}{c}{38.747}          & 47.781          & \multicolumn{1}{c}{0.770}          & \multicolumn{1}{c}{38.370}          & 47.443          \\ 
SGZ~\cite{zheng2022semantic}                     & \multicolumn{1}{c}{0.783}          & \multicolumn{1}{c}{\textbf{14.615}} & 72.367          & \multicolumn{1}{c}{0.789}          & \multicolumn{1}{c}{20.046}          & 67.735          & \multicolumn{1}{c}{0.755}          & \multicolumn{1}{c}{\textbf{14.463}} & 66.134          & \multicolumn{1}{c}{0.777}          & \multicolumn{1}{c}{25.646}          & 55.934          & \multicolumn{1}{c}{0.824}          & \multicolumn{1}{c}{31.402}          & 58.789          \\ 
LLFlow~\cite{wang2022low}                   & \multicolumn{1}{c}{0.791} & \multicolumn{1}{c}{28.861} & 67.926 & \multicolumn{1}{c}{0.805} & \multicolumn{1}{c}{27.060} & 66.816 & \multicolumn{1}{c}{0.710}          & \multicolumn{1}{c}{30.267}          & 67.019          & \multicolumn{1}{c}{\textbf{0.807}} & \multicolumn{1}{c}{26.361}          & 61.132          & \multicolumn{1}{c}{0.800}          & \multicolumn{1}{c}{31.673}          & 61.252          \\ 
URetinexNet~\cite{wu2022uretinex}              & \multicolumn{1}{c}{0.737}          & \multicolumn{1}{c}{25.570}           & 70.066          & \multicolumn{1}{c}{\textbf{0.816}} & \multicolumn{1}{c}{24.222}          & 67.423          & \multicolumn{1}{c}{0.715}          & \multicolumn{1}{c}{22.346}          & 66.310           & \multicolumn{1}{c}{0.765}          & \multicolumn{1}{c}{26.453}          & 59.856          & \multicolumn{1}{c}{0.801}          & \multicolumn{1}{c}{30.085}          & 55.399          \\ 
SCI                      & \multicolumn{1}{c}{0.702}          & \multicolumn{1}{c}{28.948}          & 64.054          & \multicolumn{1}{c}{0.747}          & \multicolumn{1}{c}{23.344}          & 64.574          & \multicolumn{1}{c}{0.733}          & \multicolumn{1}{c}{15.335}          & 64.616          & \multicolumn{1}{c}{0.720}          & \multicolumn{1}{c}{31.263}          & 48.506          & \multicolumn{1}{c}{0.779}          & \multicolumn{1}{c}{\textbf{26.132}} & 48.667          \\ \bottomrule
\end{tabular}
\label{tab:quant2}
\end{table*}

\begin{table}[t]
\centering
\caption{Quantitative Comparison on DarkFace~\cite{yang2020advancing} and ExDark~\cite{loh2019getting} with UNIQUE (UNI), BRISQUE (BRI), and SPAQ.}
\begin{tabular}{l|ccc|cc}
\toprule
\multirow{2}{*}{\textbf{Method}} & \multicolumn{3}{c|}{\textbf{DarkFace}~\cite{yang2020advancing}}                                                                & \multicolumn{2}{c}{\textbf{ExDark}~\cite{loh2019getting}}                           \\ \cline{2-6} 
                         & \multicolumn{1}{c}{\textbf{UNI$\uparrow$}}         & \multicolumn{1}{c}{\textbf{BRI$\downarrow$}}         & \textbf{SPAQ$\uparrow$}            & \multicolumn{1}{c}{\textbf{UNI$\uparrow$}}         & \textbf{SPAQ$\uparrow$}            \\  \hline
RetinexNet~\cite{wei2018deep}               & \multicolumn{1}{c}{0.737}          & \multicolumn{1}{c}{18.574}          & \textbf{54.966} & \multicolumn{1}{c}{0.708}          & \textbf{66.330}  \\ 
KinD~\cite{zhang2019kindling}                    & \multicolumn{1}{c}{0.737}          & \multicolumn{1}{c}{48.311}          & 41.070           & \multicolumn{1}{c}{0.728}          & 55.690           \\ 
Zero-DCE~\cite{guo2020zero}                  & \multicolumn{1}{c}{0.720}  & \multicolumn{1}{c}{26.194}          & 47.868& \multicolumn{1}{c}{0.729}          & 52.700   \\ 
KinD++~\cite{zhang2021beyond}                  & \multicolumn{1}{c}{0.719} & \multicolumn{1}{c}{32.492} & 52.905& \multicolumn{1}{c}{0.723} & 61.036 \\ 
RUAS~\cite{liu2021retinex}                    & \multicolumn{1}{c}{\textbf{0.740}}  & \multicolumn{1}{c}{\textbf{13.770}}  & 42.329          & \multicolumn{1}{c}{0.712}          & 47.785          \\ 
SGZ~\cite{zheng2022semantic}                    & \multicolumn{1}{c}{0.713}          & \multicolumn{1}{c}{24.647} & 47.392          & \multicolumn{1}{c}{0.729}          & 51.236          \\ 
LLFlow~\cite{wang2022low}                  & \multicolumn{1}{c}{0.708} & \multicolumn{1}{c}{22.284} & 51.544 & \multicolumn{1}{c}{\textbf{0.735}} & 56.116 \\ 
URetinexNet~\cite{wu2022uretinex}             & \multicolumn{1}{c}{0.739}          & \multicolumn{1}{c}{15.148}          & 51.290           & \multicolumn{1}{c}{0.722} & 57.291          \\ 
SCI~\cite{ma2022toward}                     & \multicolumn{1}{c}{0.719}          & \multicolumn{1}{c}{19.511}          & 46.046          & \multicolumn{1}{c}{0.709}          & 50.618          \\ \bottomrule
\end{tabular}
\label{tab:quant3}
\end{table}


\begin{table}[t]
\centering
\caption{Efficiency Comparison on ACDC~\cite{sakaridis2021acdc} (resolution of 1080 $\times$ 1920) with FLOPs, \#Params, and Inference Time using a single NVIDIA GeForce RTX 3090 GPU. \textcolor{blue}{Blue} indicates real-time capability. }
\begin{tabular}{l|r|r|r}
\toprule
\textbf{Method}       & \textbf{FLOPs$\downarrow$}       & \textbf{\#Params$\downarrow$}      & \textbf{Time$\downarrow$}       \\ \hline
RetinexNet~\cite{wei2018deep}             & -             & 0.5550           & -             \\ 
KinD~\cite{zhang2019kindling}              & 1103.9117       & 8.1600          & 3.5288          \\ 
Zero-DCE~\cite{guo2020zero}         & 164.2291        & 0.0794          & \textcolor{blue}{0.0281}         \\ 
KinD++~\cite{zhang2021beyond}           & -             & 8.2750          & -             \\ 
RUAS~\cite{liu2021retinex}         & 6.7745          & 0.0034          & \textcolor{blue}{0.0280}          \\ 
SGZ~\cite{zheng2022semantic}             & \textbf{0.2135} & 0.0106          & \textbf{\textcolor{blue}{0.0026}} \\ 
LLFlow~\cite{wang2022low}            & 892.7097        & 1.7014          & 0.3926          \\ 
URetinexNet~\cite{wu2022uretinex}       & 1801.4110       & 0.3401          & 0.2934          \\ 
SCI~\cite{ma2022toward}        & 0.7465          & \textbf{0.0003} & \textcolor{blue}{0.0058}          \\ \bottomrule
\end{tabular}
\label{tab:eff}
\end{table}

\begin{table}[t]
\centering
\caption{Semantic Segmentation Result Comparison on ACDC~\cite{sakaridis2021acdc} and DCS~\cite{zheng2022semantic} with mPA (\%)  and mIoU (\%).}
\setlength\tabcolsep{2.5pt}
\begin{tabular}{ccc|ccc}
\toprule
\multicolumn{3}{c|}{\textbf{ACDC}~\cite{sakaridis2021acdc}}                                                               & \multicolumn{3}{c}{\textbf{DCS}~\cite{zheng2022semantic}}                                                               \\ \hline
\multicolumn{1}{l|}{\textbf{Method}}     & \multicolumn{1}{c}{\textbf{mPA$\uparrow$}}            & \textbf{mIoU$\uparrow$}           & \multicolumn{1}{l|}{\textbf{Method}}    & \multicolumn{1}{c}{\textbf{mPA$\uparrow$}}            & \textbf{mIoU$\uparrow$}           \\ \hline 
\multicolumn{1}{l|}{KinD~\cite{zhang2019kindling}}        & \multicolumn{1}{c}{60.79}          & 49.18          & \multicolumn{1}{l|}{PIE~\cite{fu2015probabilistic}}        & \multicolumn{1}{c}{68.89}          & 61.97          \\ 
\multicolumn{1}{l|}{Zero-DCE~\cite{guo2020zero}}     & \multicolumn{1}{c}{59.00}          & 49.51          & \multicolumn{1}{l|}{RetinexNet~\cite{wei2018deep}} & \multicolumn{1}{c}{66.76}          & 57.96          \\ 
\multicolumn{1}{l|}{RUAS~\cite{liu2021retinex}}        & \multicolumn{1}{c}{50.42}          & 44.48          & \multicolumn{1}{l|}{MBLLEN~\cite{lv2018mbllen}}     & \multicolumn{1}{c}{59.06}          & 51.98          \\ 
\multicolumn{1}{l|}{SGZ~\cite{zheng2022semantic}}         & \multicolumn{1}{c}{61.65}          & 49.50          & \multicolumn{1}{l|}{KinD~\cite{zhang2019kindling}}       & \multicolumn{1}{c}{71.69}          & 63.42          \\ 
\multicolumn{1}{l|}{LLFlow~\cite{wang2022low}}      & \multicolumn{1}{c}{\textbf{62.68}} & 49.30          & \multicolumn{1}{l|}{Zero-DCE~\cite{guo2020zero}}    & \multicolumn{1}{c}{74.20}          & 64.36          \\ 
\multicolumn{1}{l|}{URetinexNet~\cite{wu2022uretinex}} & \multicolumn{1}{c}{62.32}          & 48.71          & \multicolumn{1}{l|}{Zero-DCE++~\cite{li2021learning}}  & \multicolumn{1}{c}{74.43}          & 65.51          \\ 
\multicolumn{1}{l|}{SCI~\cite{ma2022toward}}         & \multicolumn{1}{c}{57.52}          & \textbf{49.66} & \multicolumn{1}{l|}{SGZ~\cite{zheng2022semantic}}        & \multicolumn{1}{c}{\textbf{74.50}} & \textbf{65.87} \\ \bottomrule
\end{tabular}
\label{tab:sem}
\end{table}

\begin{table}[t]
\centering
\caption{Object Detection Result Comparison on DarkFace with different IoU thresholds. Each numerical entry is a mAP$\uparrow$ value. }
\setlength\tabcolsep{3.5pt}
\begin{tabular}{c|l|c|c|c}
\toprule
\textbf{Learning}            & \textbf{Method}    & \textbf{IoU@0.5$\uparrow$} & \textbf{IoU@0.6$\uparrow$} & \textbf{IoU@0.7$\uparrow$} \\ \hline
TL                   & LIME~\cite{guo2016lime}      & 0.244   & 0.083   & 0.010   \\ \hline
\multirow{6}{*}{SL}  & LLNet~\cite{lore2017llnet}      & 0.228   & 0.063   & 0.006   \\ 
                     & LightenNet~\cite{li2018lightennet} & 0.270   & 0.085   & 0.011   \\ 
                     & MBLLEN~\cite{lv2018mbllen}    & 0.269   & 0.092   & 0.012   \\ 
                     & KinD~\cite{zhang2019kindling}     & 0.255   & 0.081   & 0.010   \\ 
                     & KinD++~\cite{zhang2021beyond}    & 0.271   & 0.090   & 0.011   \\
                      & URetinexNet~\cite{wu2022uretinex} & 0.283   & 0.101  & 0.015  \\ 
                      & LLFlow~\cite{wang2022low} & \textbf{0.290} & \textbf{0.103} & \textbf{0.016}           \\ \hline
UL                   & EnlightenGAN~\cite{jiang2021enlightengan}      & 0.261   & 0.088   & 0.012   \\  \hline
\multirow{4}{*}{ZSL} & ExCNet~\cite{zhang2019zero}     & 0.276   & 0.092   & 0.010   \\ 
                     & Zero-DCE~\cite{guo2020zero}    & 0.281   & 0.092   & 0.013   \\ 
                     & Zero-DCE++~\cite{li2021learning}  & 0.278   & 0.090   & 0.012   \\ 
                     & SGZ~\cite{zheng2022semantic}       & 0.279   & 0.092   & 0.012   \\ \bottomrule
\end{tabular}
\label{tab:det}
\end{table}

\subsection{User Studies} 

Since there are few effective metrics to evaluate the visual quality of low-light video enhancement, we conducted a user study to assess the performances of different methods on the proposed Night Wenzhou dataset. Specifically, we ask 100 adult participants to watch the enhancement results of 7 models, including EnlightenGAN~\cite{jiang2021enlightengan}, KinD~\cite{zhang2019kindling}, KinD+~\cite{zhang2021beyond}, MBLLEN~\cite{lv2018mbllen}, RetinexNet~\cite{wei2018deep}, SGZ~\cite{zheng2022semantic}, and Zero-DCE~\cite{guo2020zero}. They are asked to vote `1' to `5' for each method, where `1' indicates the worst performance, and `5' indicates the best.



\begin{figure}[t]
    \centering
    \includegraphics[width=\columnwidth]{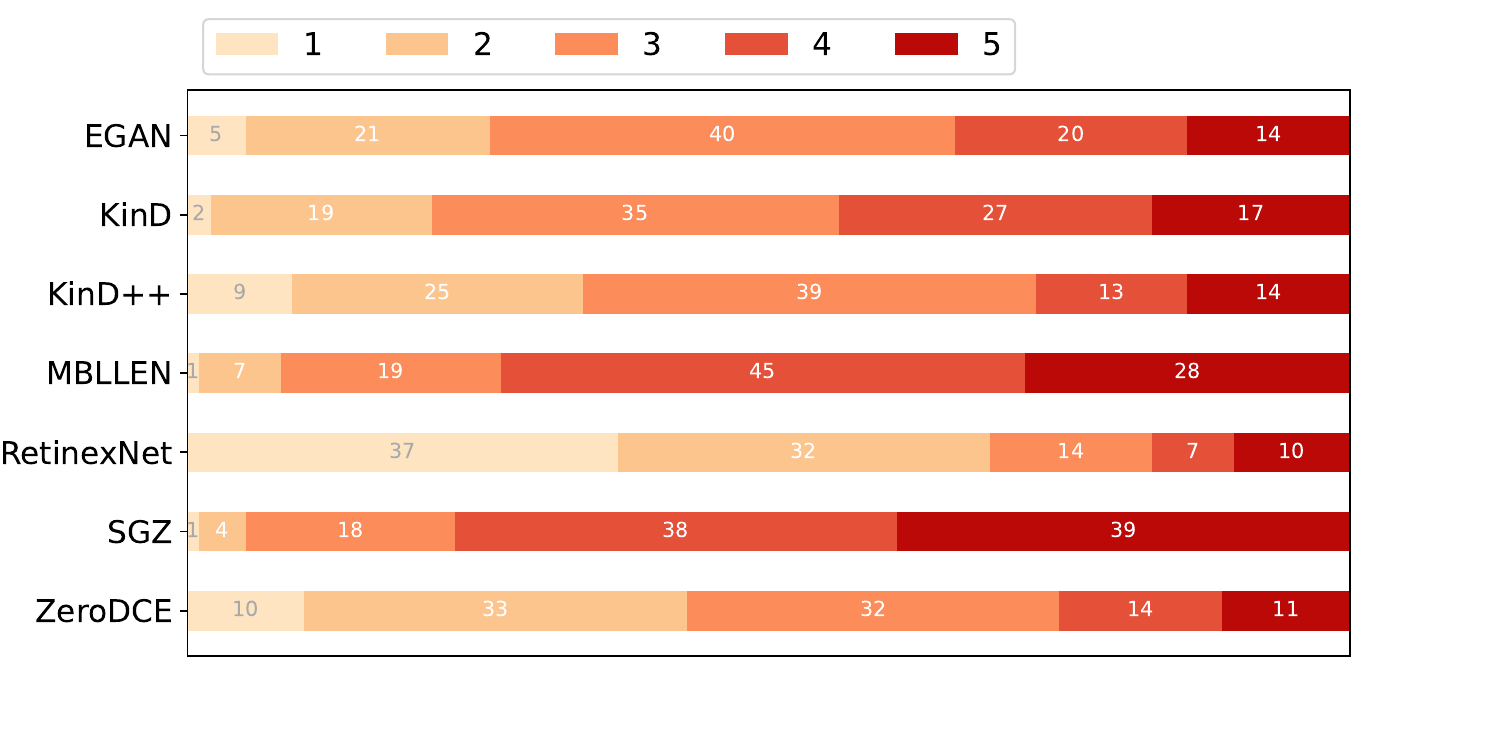}
    \caption{\textbf{User Study Results for Low-Light Image Enhancement in Stacked Bar Plot.} With exactly 100 participants, each count in the plot directly corresponds to a percentage.}
    \label{fig:barplot}
\end{figure}

We make a stacked bar graph in Fig. \ref{fig:barplot} to show the category-wise information for different methods. It can be seen that RetinexNet~\cite{wei2018deep} has the most (37 \%) of `1's; Zero-DCE~\cite{guo2020zero} has the most (33 \%) of `2's; EnlightenGAN~\cite{jiang2021enlightengan} has the most (40 \%) of `3's; MBLLEN has the most (45 \%) of `4's; SGZ~\cite{zheng2022semantic} has the most (39 \%) of `5's. Therefore, RetinexNet~\cite{wei2018deep} is voted to have the worst performance, whereas SGZ~\cite{zheng2022semantic} is voted to have the best performance.


\subsection{Qualitative Comparisons}

\begin{figure*}[t]
    \centering
    \subfloat[Dark]{
    \includegraphics[width=0.153\linewidth]{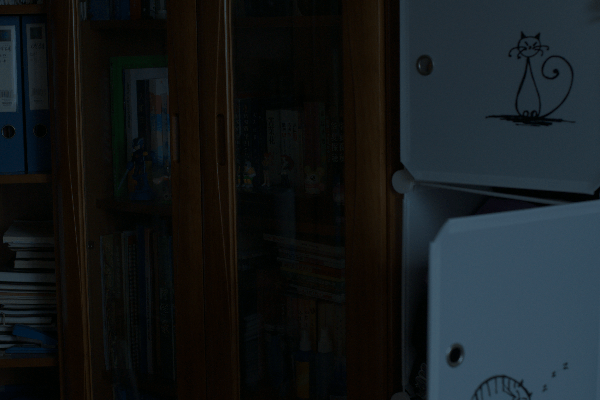}
    }
    \subfloat[PIE~\cite{fu2015probabilistic}]{
    \includegraphics[width=0.153\linewidth]{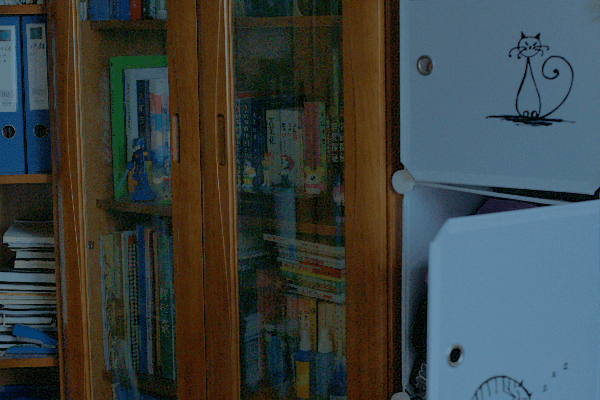}
    }
    \subfloat[LIME~\cite{guo2016lime}]{
    \includegraphics[width=0.153\linewidth]{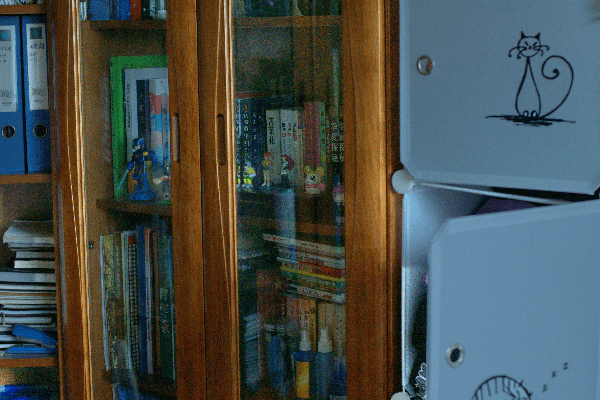}
    }
    \subfloat[MBLLEN~\cite{lv2018mbllen}]{
    \includegraphics[width=0.153\linewidth]{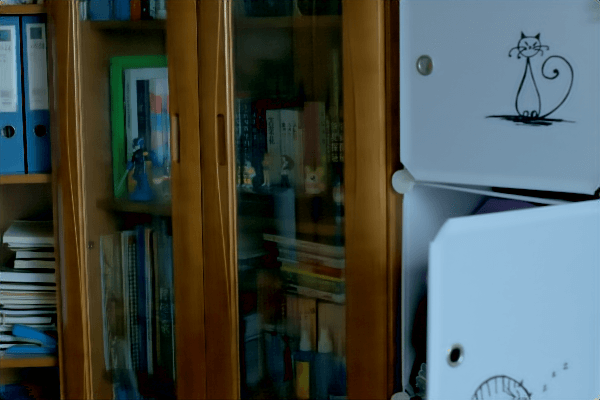}
    }
    \subfloat[KinD~\cite{zhang2019kindling}]{
    \includegraphics[width=0.153\linewidth]{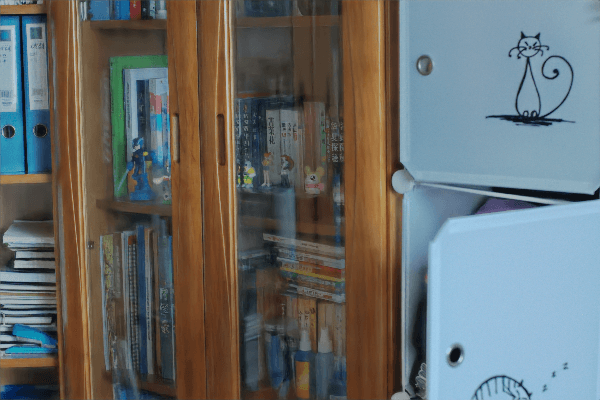}
    }
    \subfloat[Zero-DCE~\cite{guo2020zero}]{
    \includegraphics[width=0.153\linewidth]{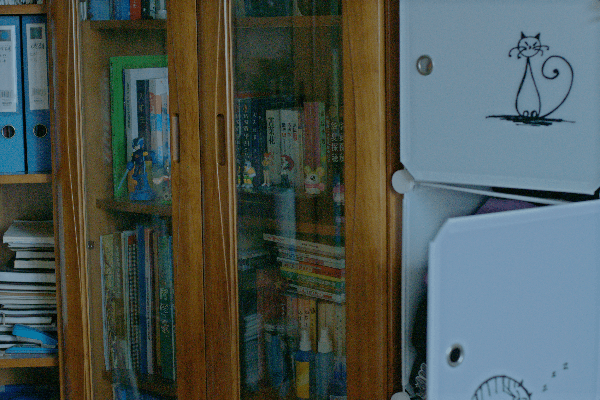}
    }
    \\
    \subfloat[RUAS~\cite{liu2021retinex}]{
    \includegraphics[width=0.153\linewidth]{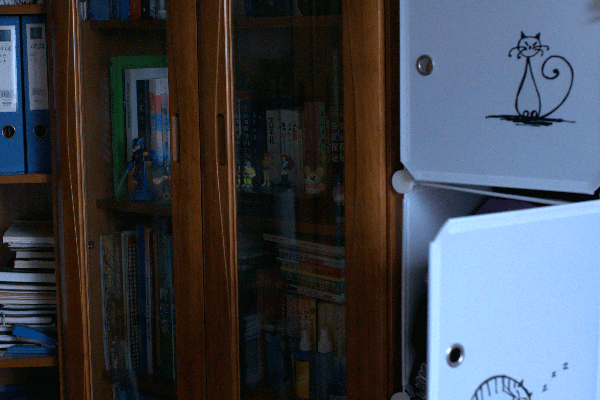}
    }
    \subfloat[LLFlow~\cite{wang2022low}]{
    \includegraphics[width=0.153\linewidth]{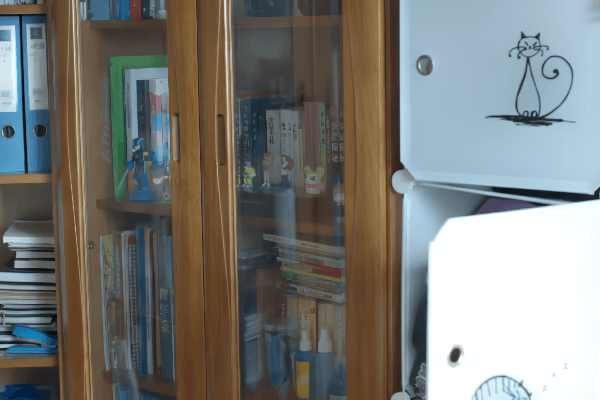}
    }
    \subfloat[SGZ~\cite{zheng2022semantic}]{
    \includegraphics[width=0.153\linewidth]{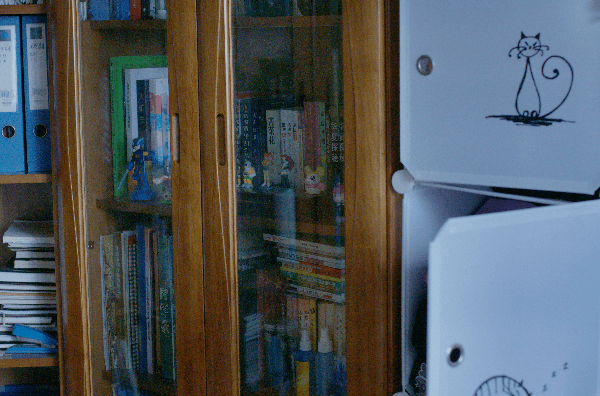}
    }
    \subfloat[SCI~\cite{ma2022toward}]{
    \includegraphics[width=0.153\linewidth]{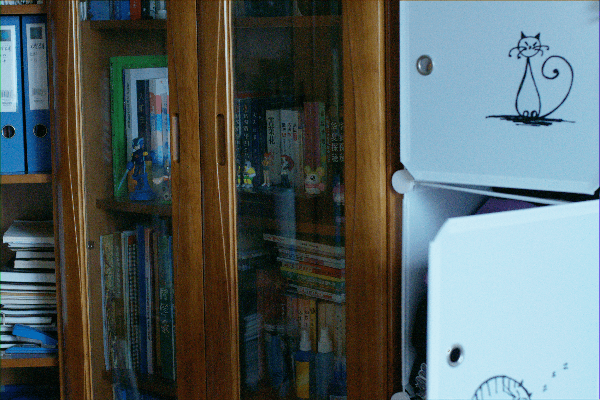}
    }
    \subfloat[URetinexNet~\cite{wu2022uretinex}]{
    \includegraphics[width=0.153\linewidth]{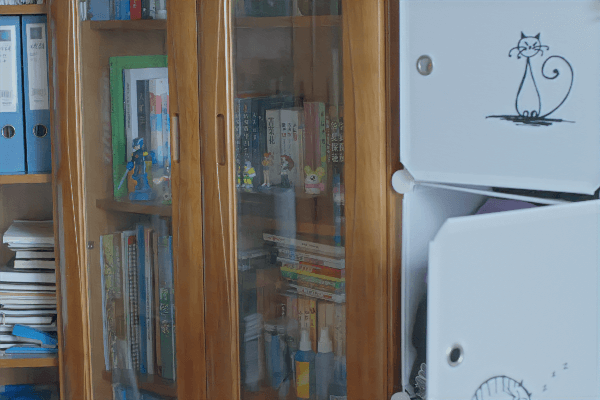}
    }
    \subfloat[Ground Truth]{
    \includegraphics[width=0.153\linewidth]{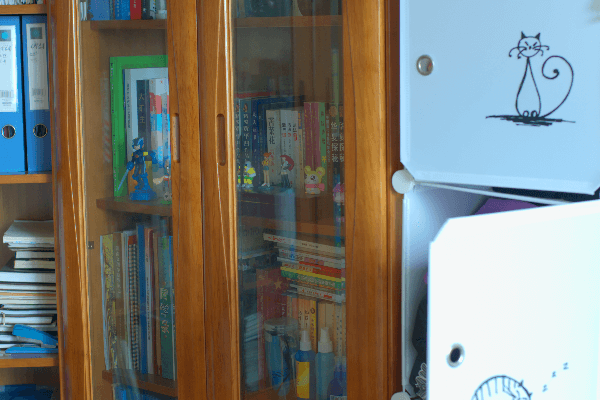}
    }
    \caption{\textbf{Visual Comparison on the LOL Dataset.}}
    \label{fig:LOL}
\end{figure*}

\noindent \textbf{Results on the LOL Dataset:}
Fig. \ref{fig:LOL} presents the qualitative comparison on an image from the LOL~\cite{wei2018deep} dataset. Our finding are as follows: 1) RUAS~\cite{liu2021retinex} produce over-exposed result 2) MBLLEN~\cite{lv2018mbllen} over-smooth the image. 3) PIE~\cite{fu2015probabilistic}, LIME~\cite{guo2016lime}, Zero-DCE~\cite{guo2020zero}, SGZ~\cite{zheng2022semantic}, and SCI~\cite{ma2022toward} yield noise. 3) KinD~\cite{zhang2019kindling}, LLFlow~\cite{wang2022low}, and URetinexNet~\cite{wu2022uretinex} are close to GT. 

\begin{figure*}[t]
    \centering
    \subfloat[PIE~\cite{fu2015probabilistic}]{
    \includegraphics[width=0.187\linewidth]{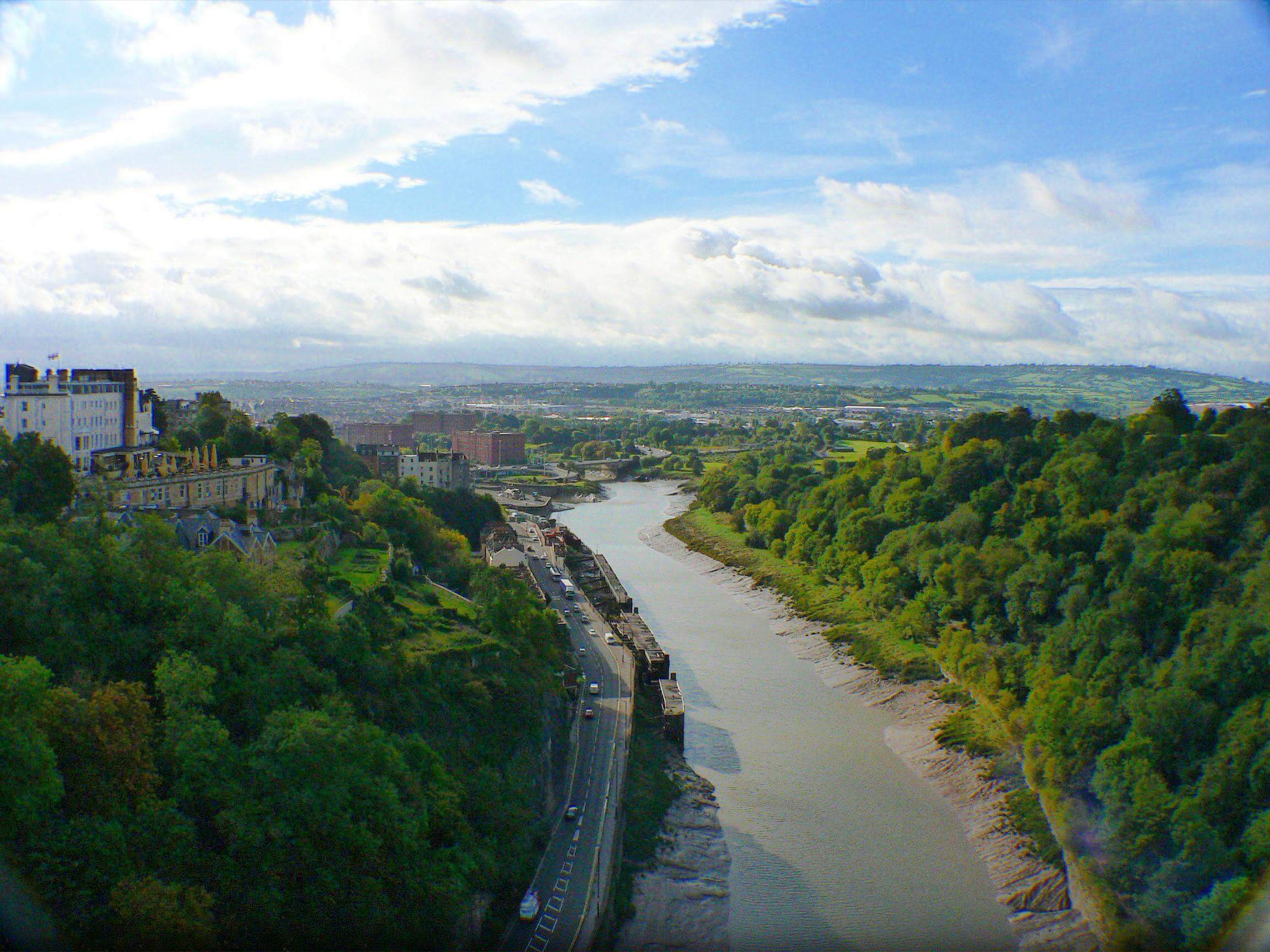}
    }
    \subfloat[LIME~\cite{guo2016lime}]{
    \includegraphics[width=0.187\linewidth]{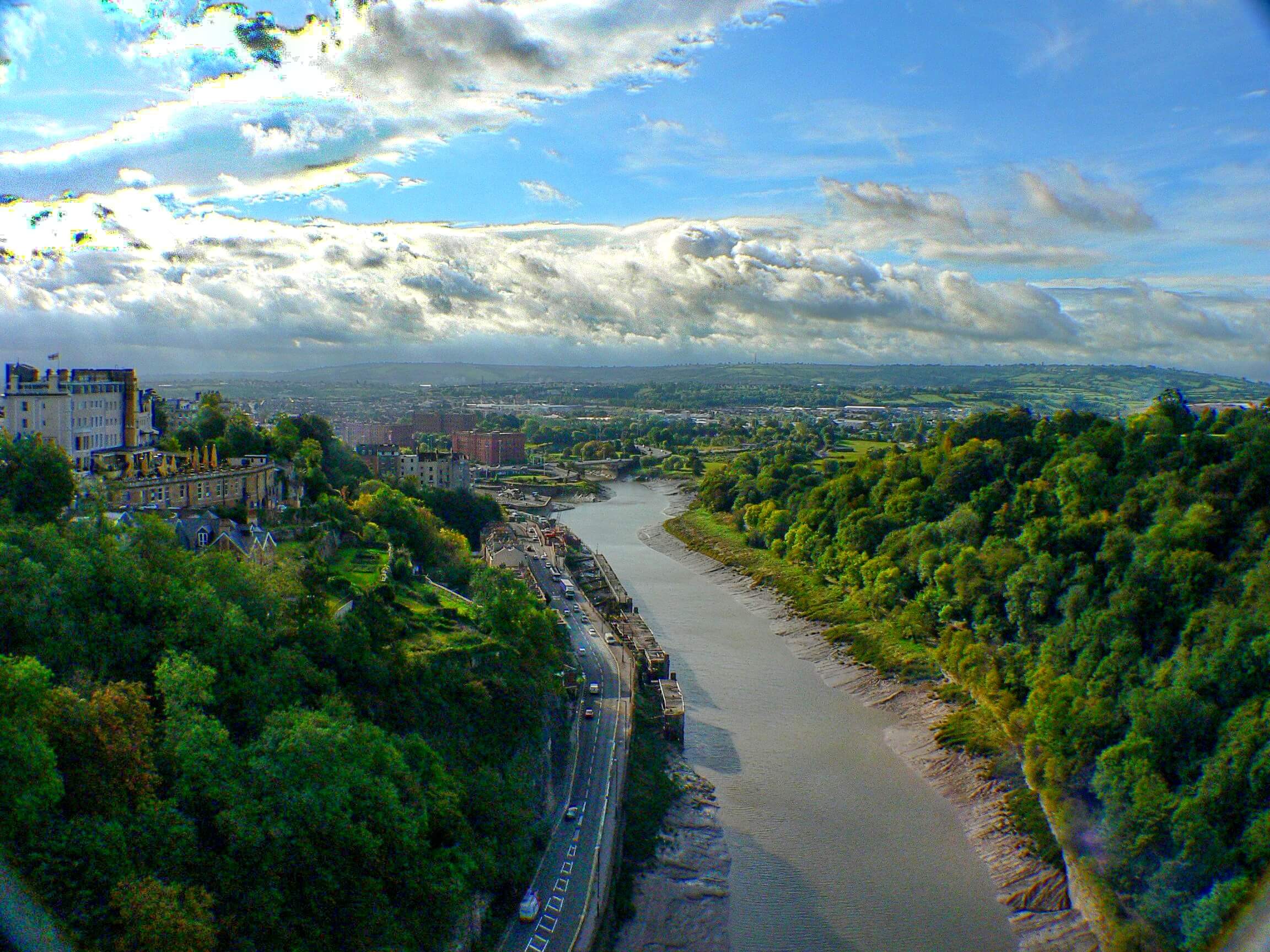}
    }
    \subfloat[MBLLEN~\cite{lv2018mbllen}]{
     \includegraphics[width=0.187\linewidth]{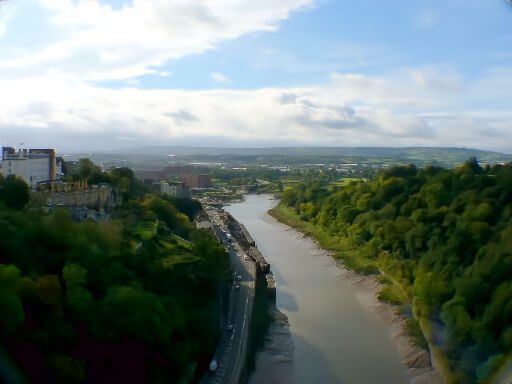}
    }
   \subfloat[KinD~\cite{zhang2019kindling}]{
    \includegraphics[width=0.187\linewidth]{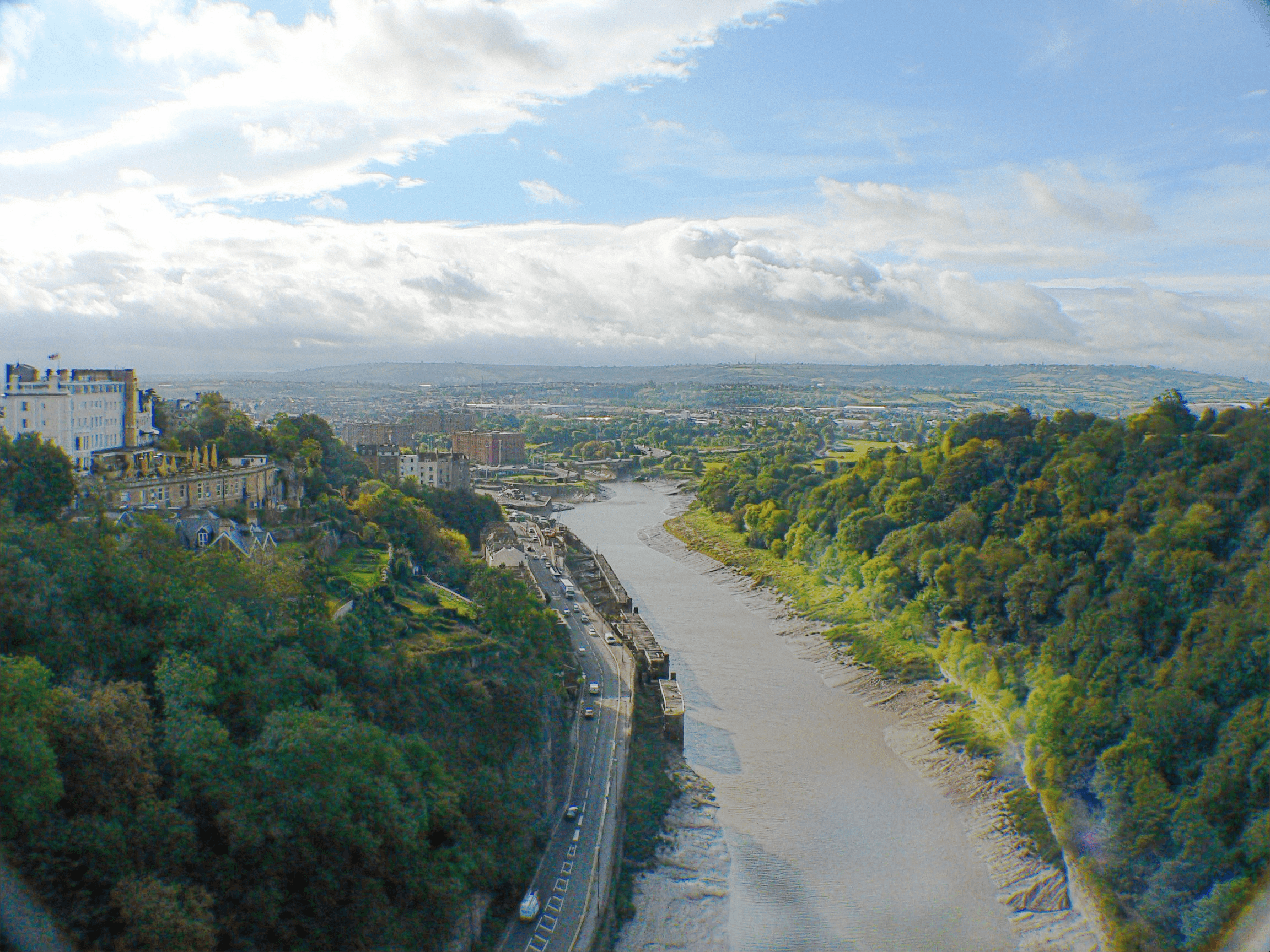}
    }
    \subfloat[Zero-DCE~\cite{guo2020zero}]{
    \includegraphics[width=0.187\linewidth]{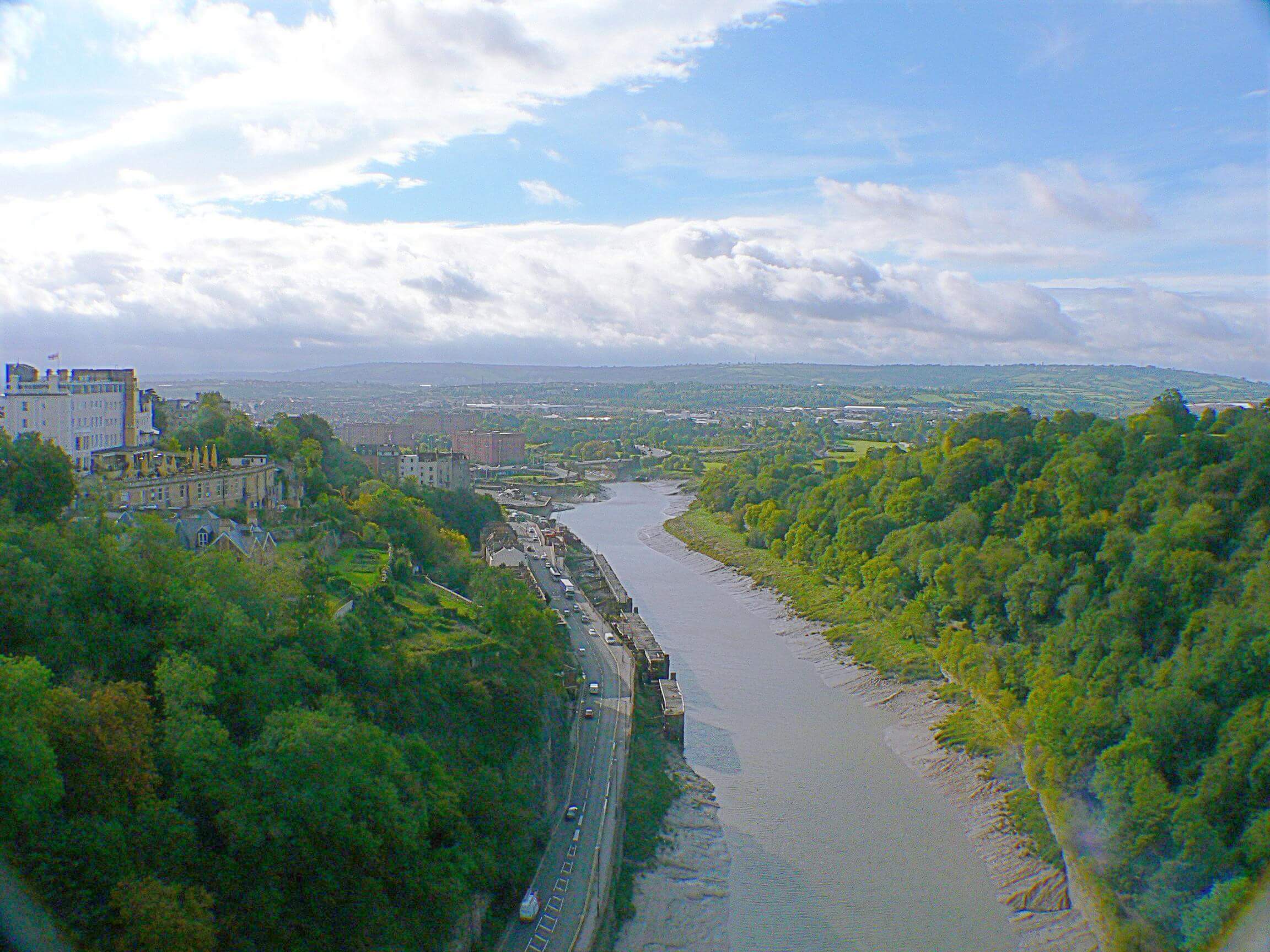}
    } 
    \\
    \subfloat[RUAS~\cite{liu2021retinex}]{
    \includegraphics[width=0.187\linewidth]{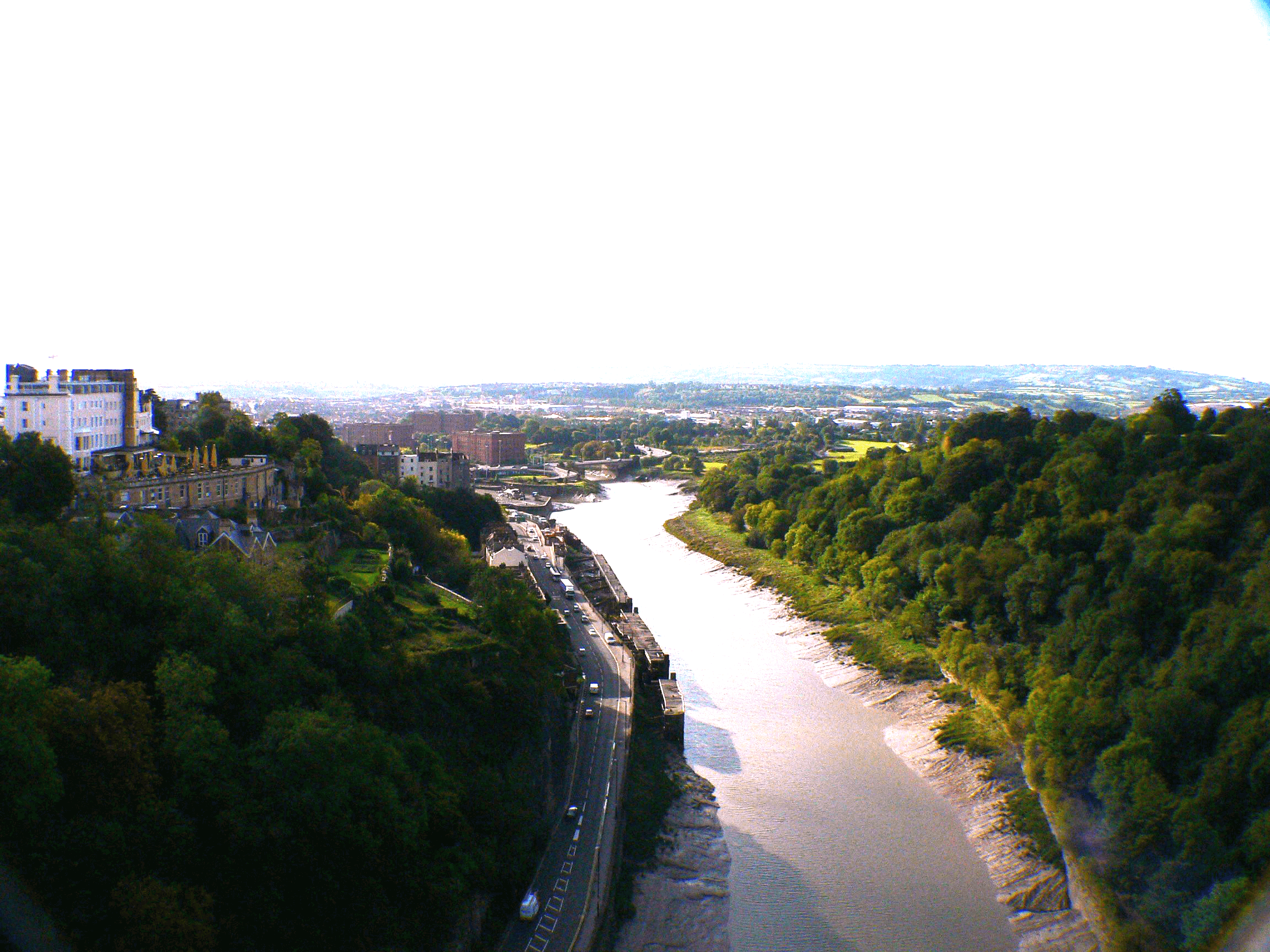}
    }
    \subfloat[LLFlow~\cite{wang2022low}]{
    \includegraphics[width=0.187\linewidth]{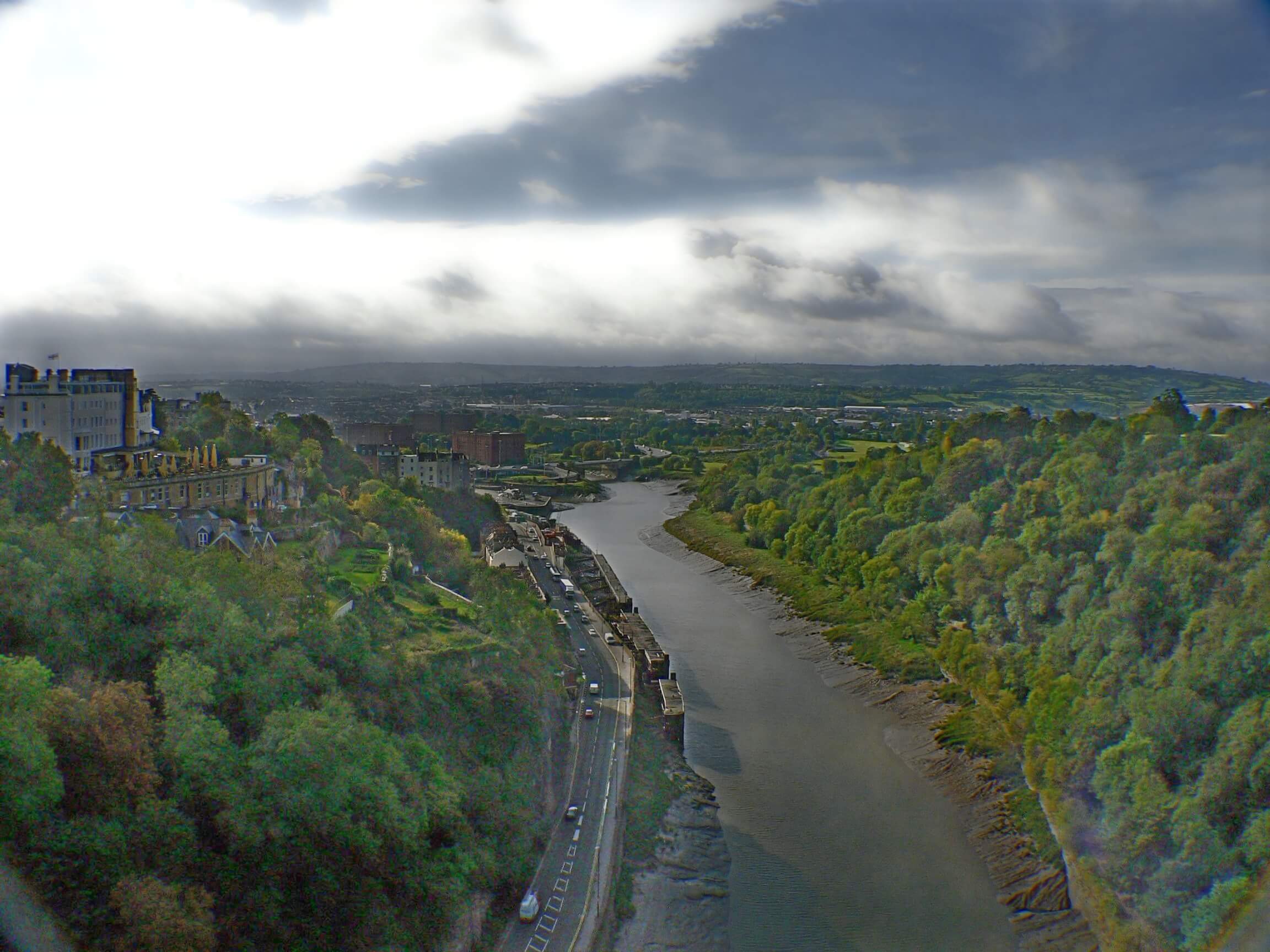}
    }
    \subfloat[SGZ~\cite{zheng2022semantic}]{
    \includegraphics[width=0.187\linewidth]{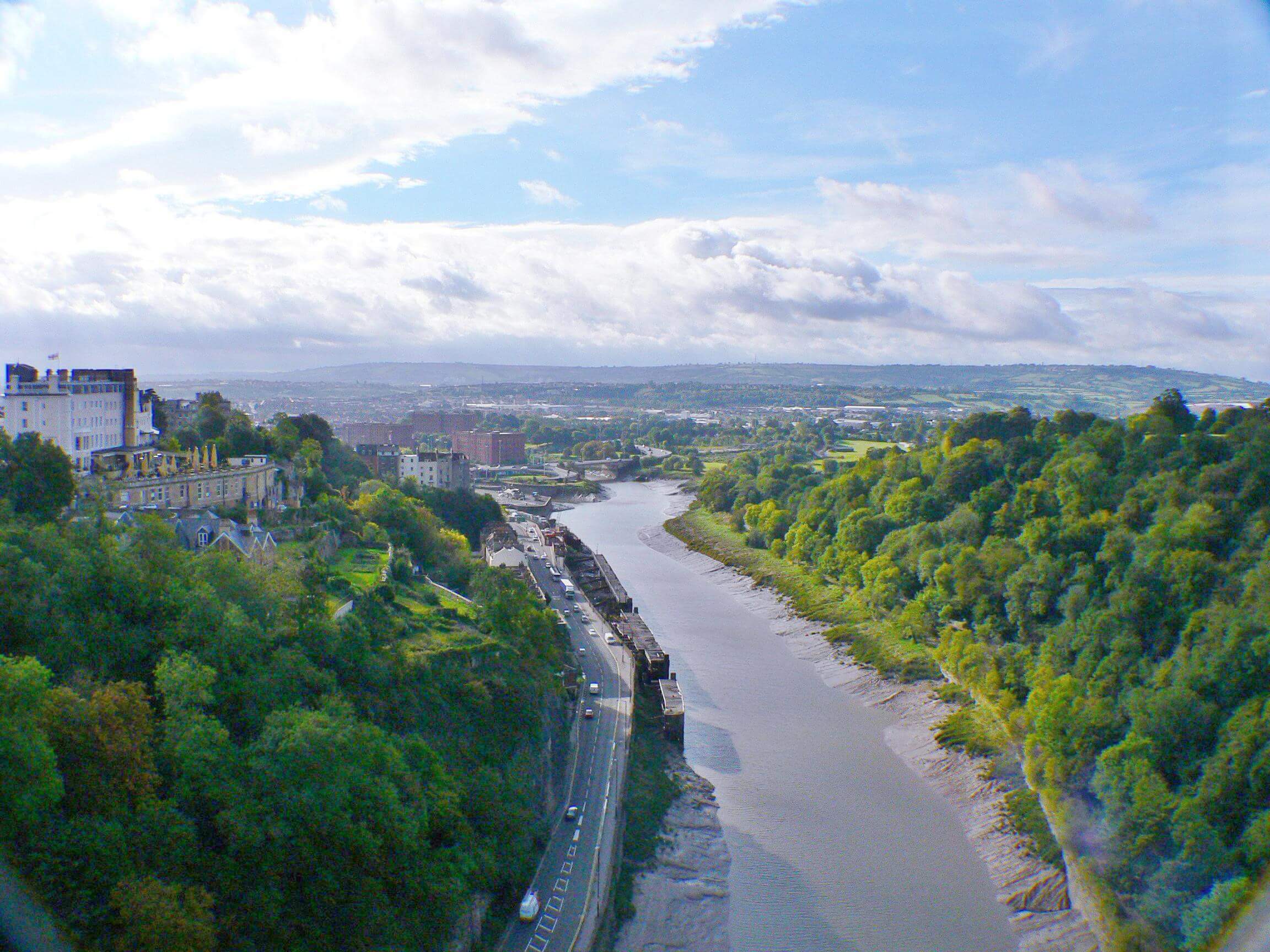}
    }
    \subfloat[SCI~\cite{ma2022toward}]{
    \includegraphics[width=0.187\linewidth]{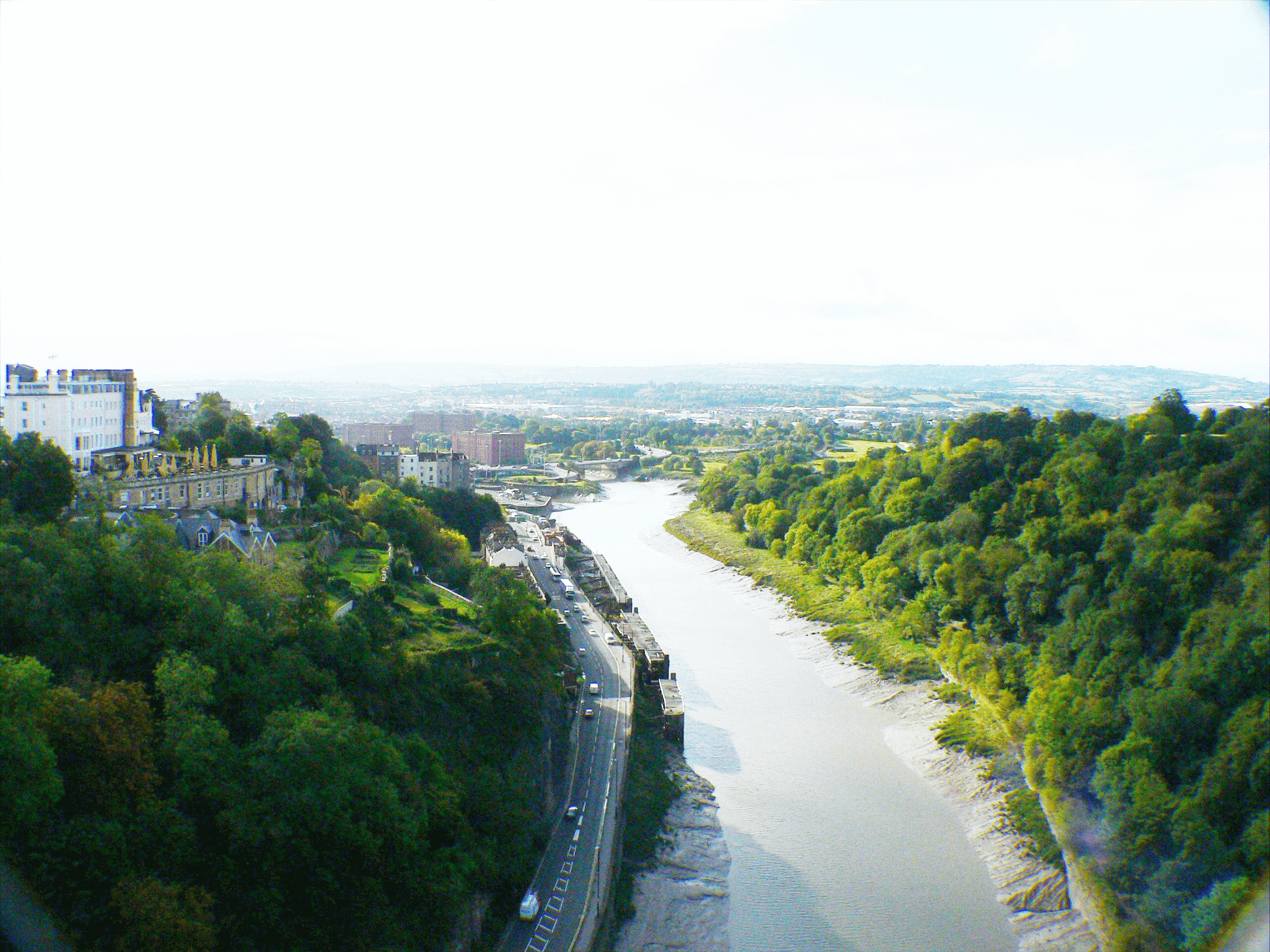}
    }
    \subfloat[URetinexNet~\cite{wu2022uretinex}]{
    \includegraphics[width=0.187\linewidth]{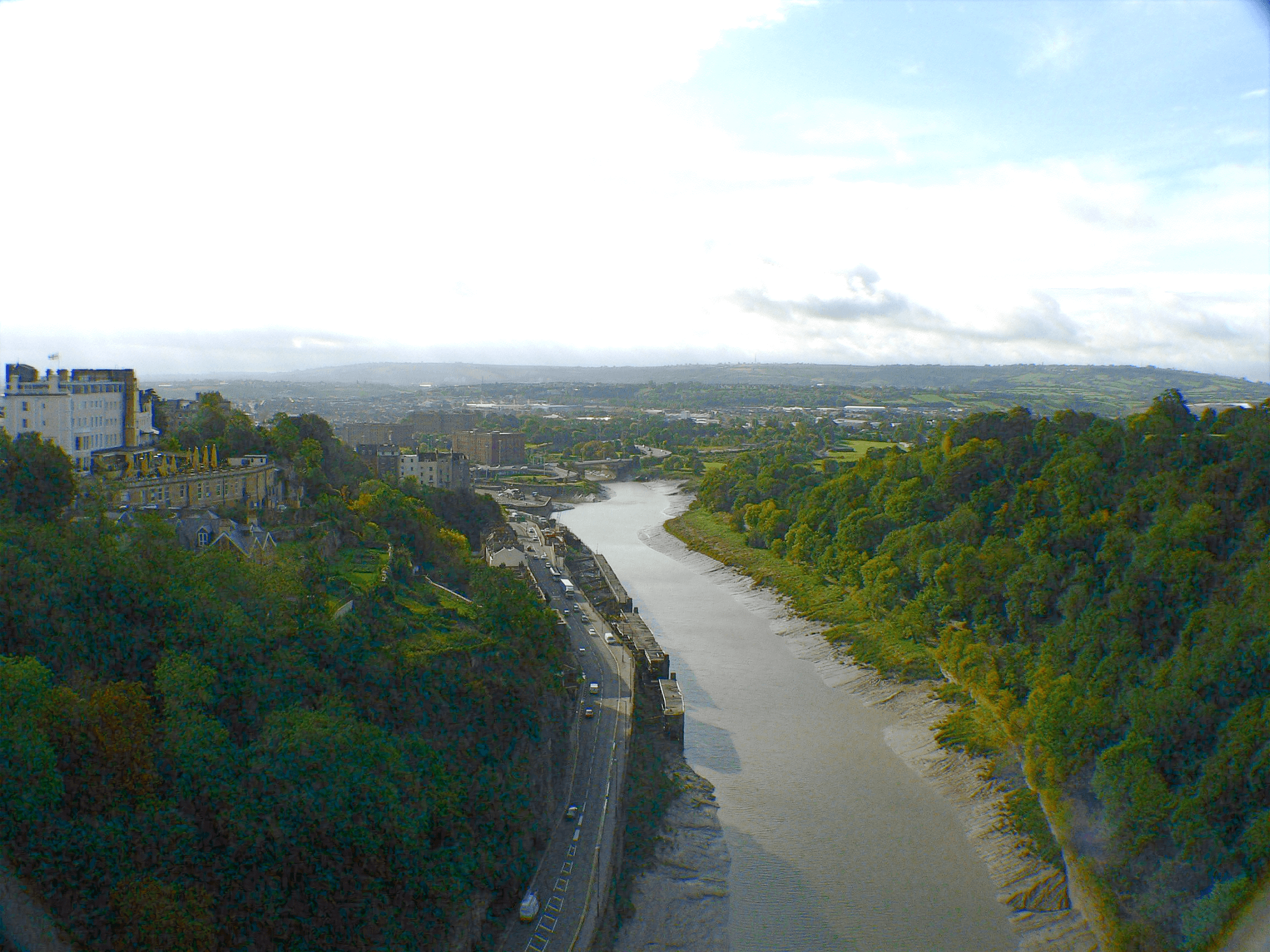}
    }
    \caption{\textbf{Visual Comparison on the VV Dataset.}}
    \label{fig:VV}
\end{figure*}

\noindent \textbf{Results on the VV Dataset:}
Fig. \ref{fig:VV}  shows the qualitative comparison for an image from the VV dataset. Our finding are as follows: 1) RUAS~\cite{liu2021retinex} produces under-exposed trees. 2) RUAS~\cite{liu2021retinex}, SCI~\cite{ma2022toward}, and URetinexNet~\cite{wu2022uretinex} renders over-exposed skies. 3) LIME~\cite{guo2016lime}, Zero-DCE~\cite{guo2020zero}, and LLFlow~\cite{wang2022low} generates artifacts. 4) MBLLEN~\cite{lv2018mbllen} oversmooths the image. 5) LLFlow~\cite{wang2022low} yield color distortion. 6) PIE~\cite{fu2015probabilistic}, KinD~\cite{zhang2019kindling}, and SGZ~\cite{zheng2022semantic} have a promising perceptual quality. 

\begin{figure*}[t]
    \centering
     \subfloat[KinD~\cite{zhang2019kindling}]{
     \includegraphics[width=0.237\linewidth]{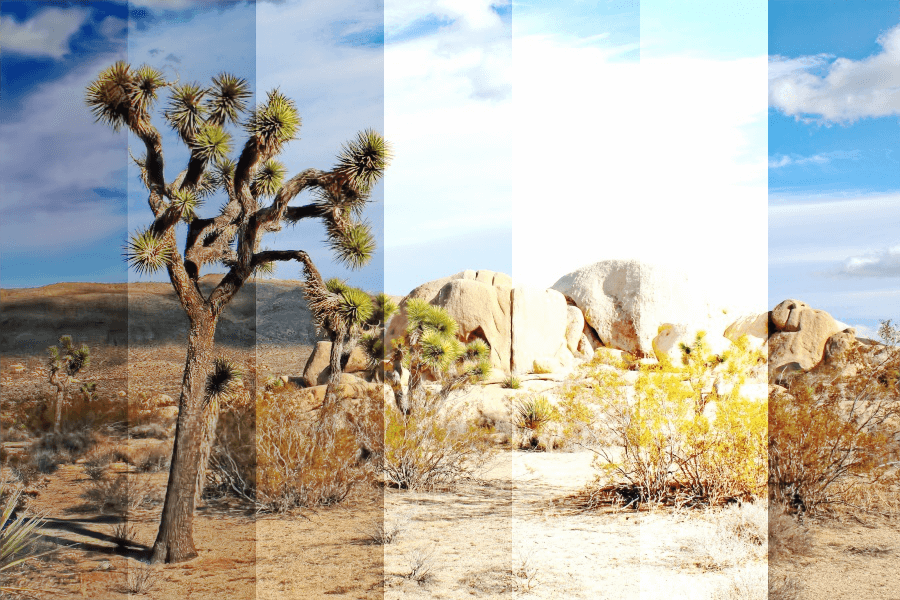}
     }
    \subfloat[Zero-DCE~\cite{guo2020zero}]{
    \includegraphics[width=0.237\linewidth]{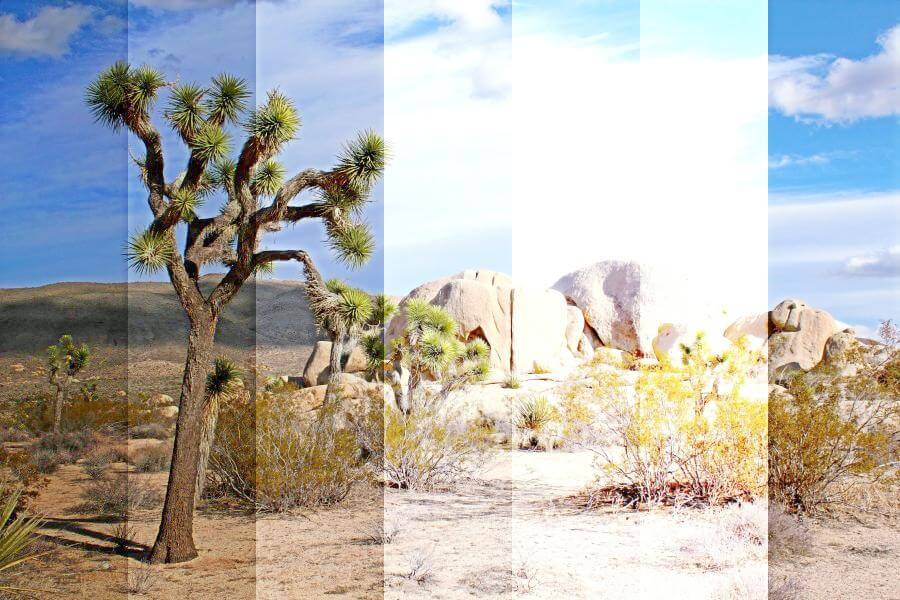}
    }
    \subfloat[KinD++~\cite{zhang2021beyond}]{
    \includegraphics[width=0.237\linewidth]{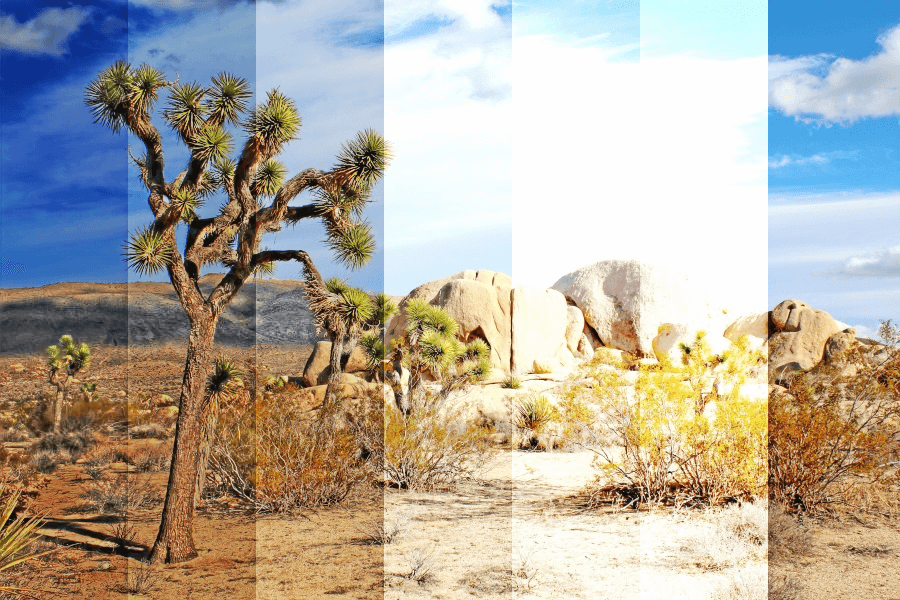}
    }
    \subfloat[RUAS~\cite{liu2021retinex}]{
    \includegraphics[width=0.237\linewidth]{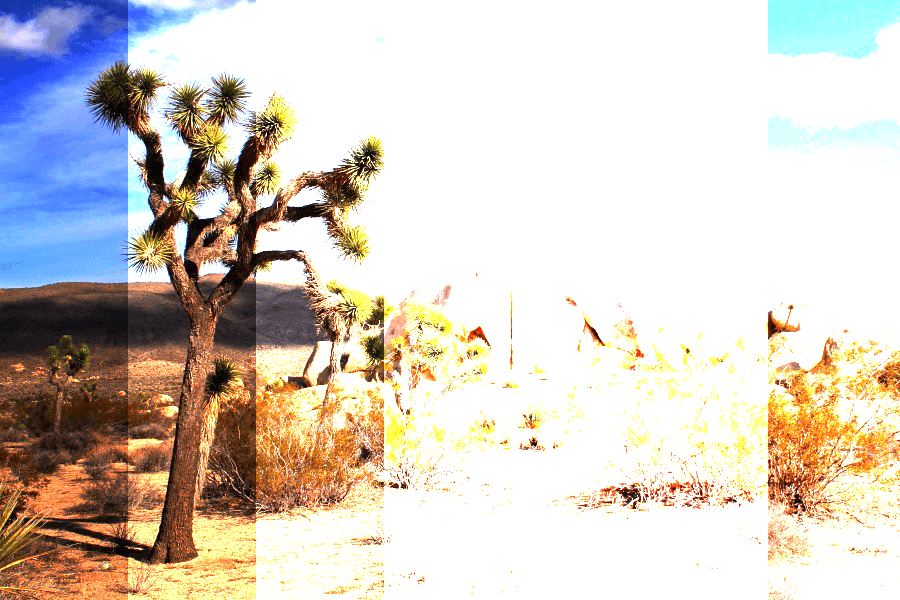}
    }
    \\
    \subfloat[LLFlow~\cite{wang2022low}]{
    \includegraphics[width=0.237\linewidth]{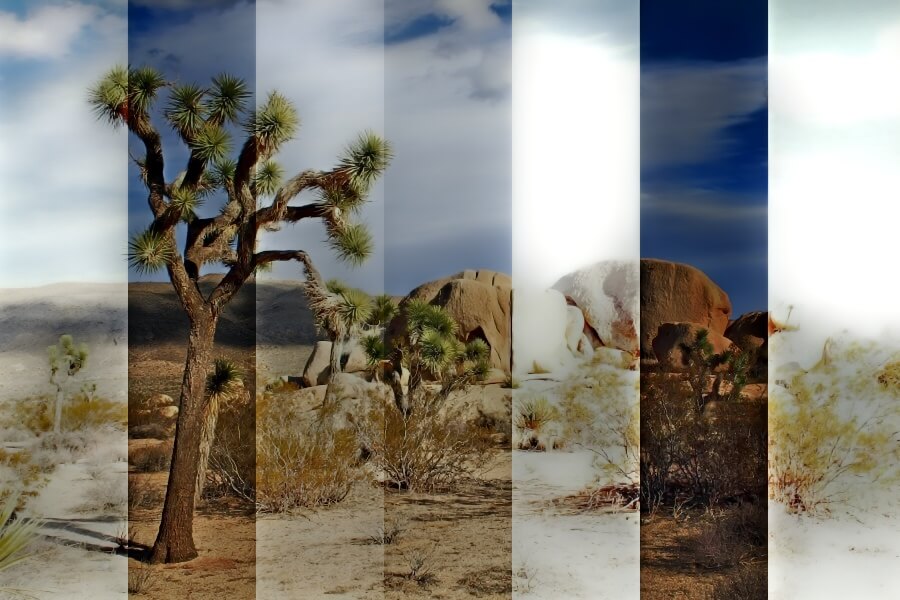}
    }
    \subfloat[SGZ~\cite{zheng2022semantic}]{
    \includegraphics[width=0.237\linewidth]{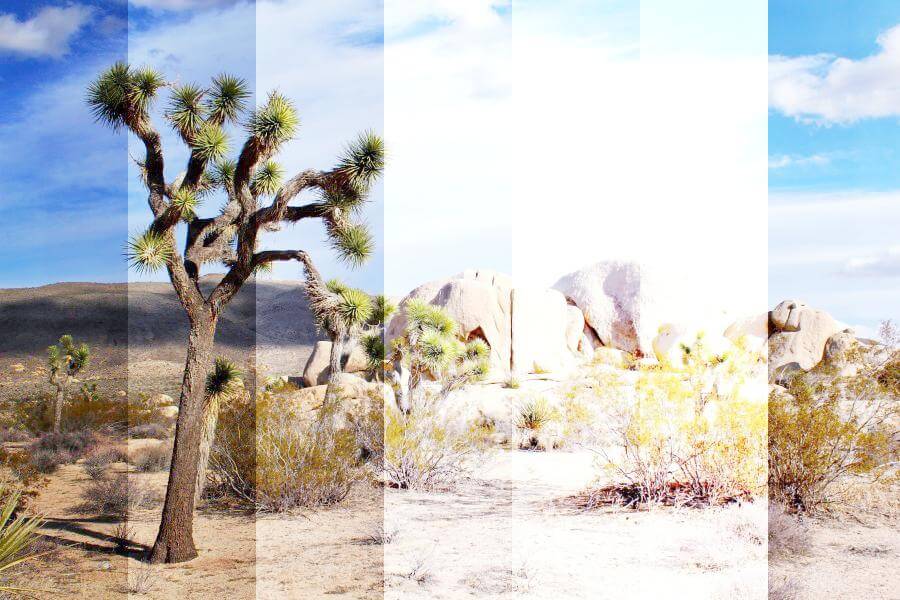}
    }
    \subfloat[SCI~\cite{ma2022toward}]{
    \includegraphics[width=0.237\linewidth]{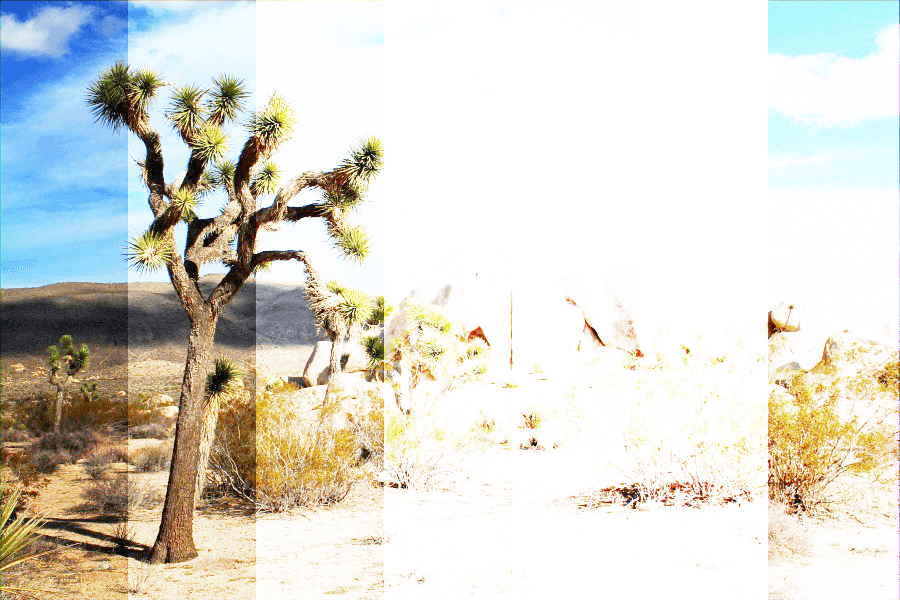}
    }
    \subfloat[URetinexNet~\cite{wu2022uretinex}]{
    \includegraphics[width=0.237\linewidth]{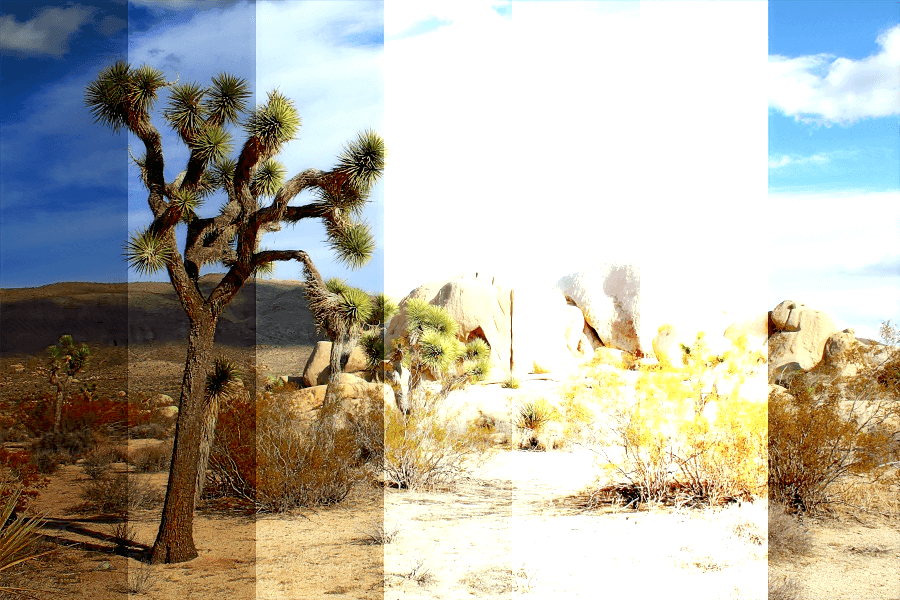}
    }
   
    \caption{\textbf{Visual Comparison on our SICE\_Grad Dataset.} Current low-light image enhancement methods struggles with uneven exposure in SICE\_Grad.}
    \label{fig:SICE_Grad}
\end{figure*}

\begin{figure*}[t]
    \centering
    \subfloat[KinD~\cite{zhang2019kindling}]{
    \includegraphics[width=0.237\linewidth]{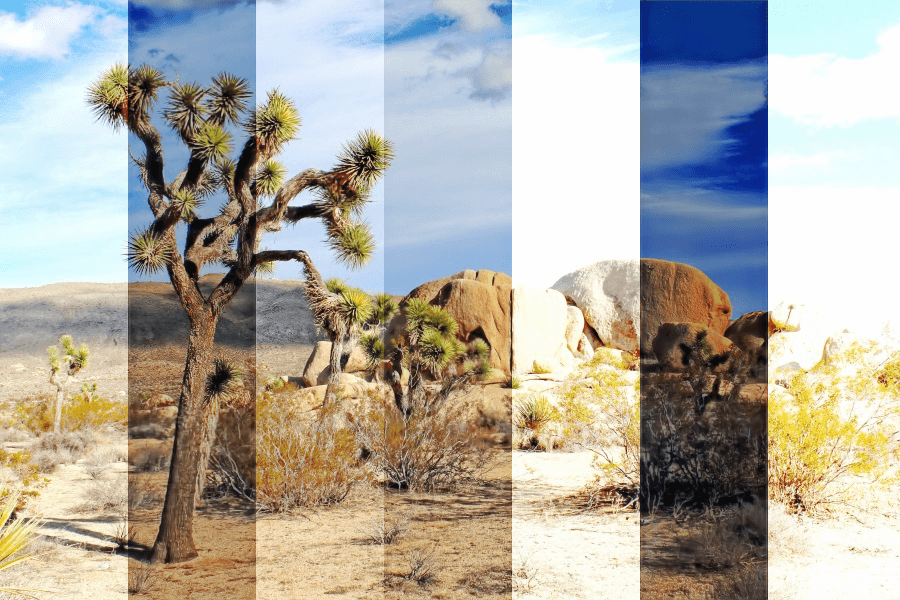}
    }
    \subfloat[Zero-DCE~\cite{guo2020zero}]{ 
    \includegraphics[width=0.237\linewidth]{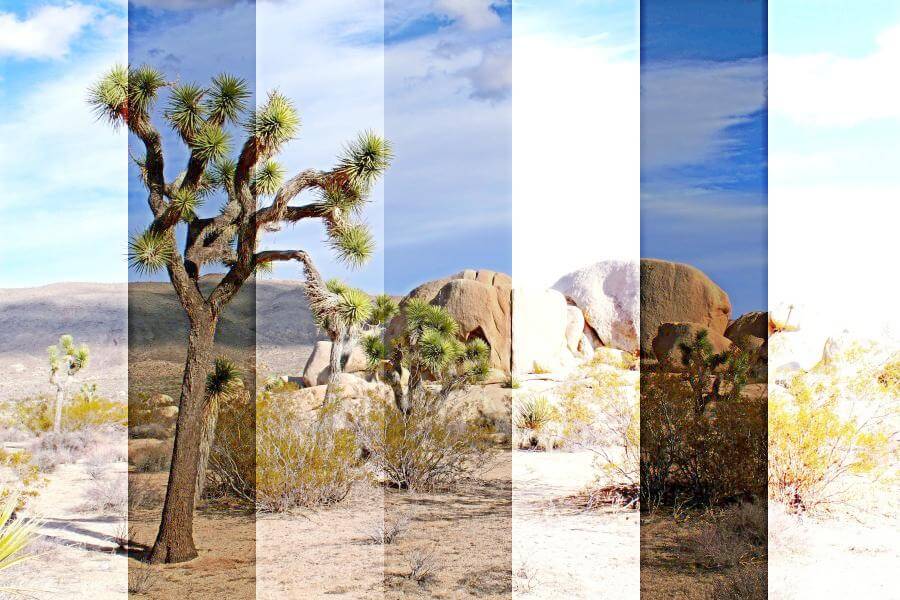}
    }
    \subfloat[KinD++~\cite{zhang2021beyond}]{
    \includegraphics[width=0.237\linewidth]{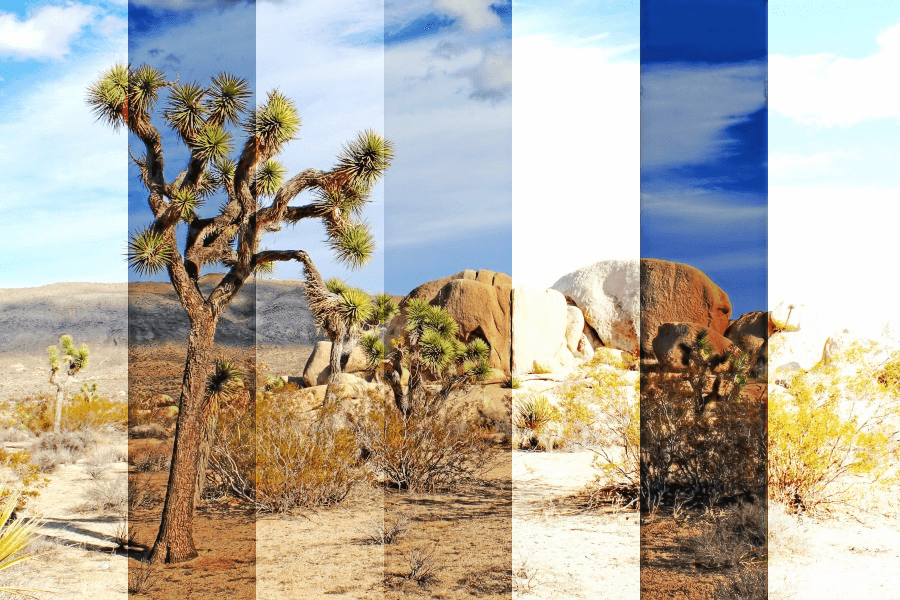}
    }
    \subfloat[RUAS~\cite{liu2021retinex}]{
    \includegraphics[width=0.237\linewidth]{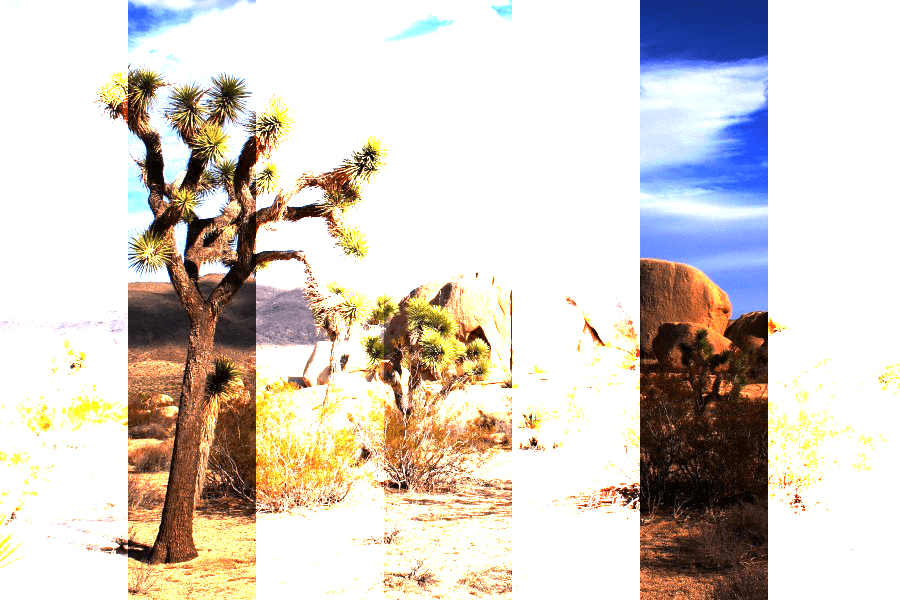}
    }
   \\
    \subfloat[LLFlow~\cite{wang2022low}]{
    \includegraphics[width=0.237\linewidth]{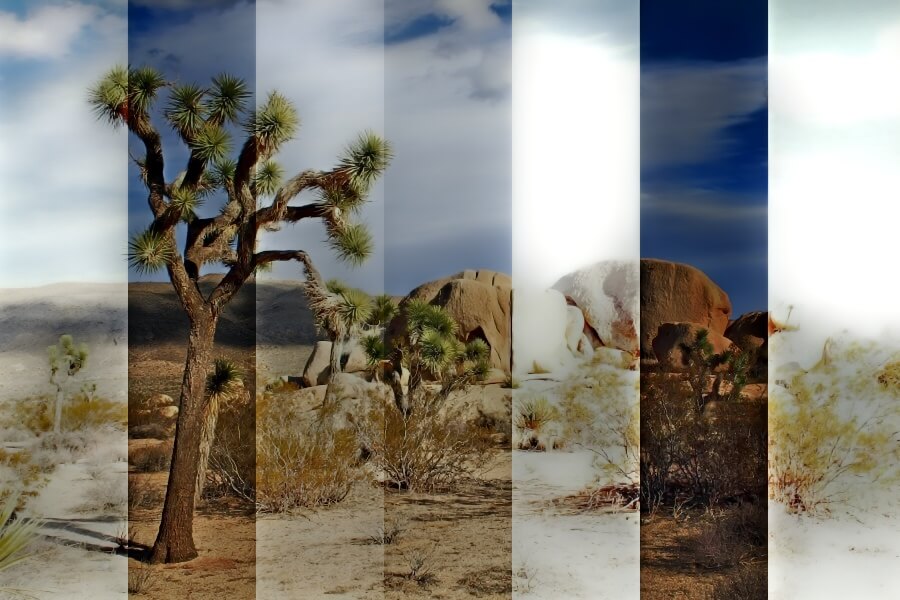}
    }
    \subfloat[SGZ~\cite{zheng2022semantic}]{
    \includegraphics[width=0.237\linewidth]{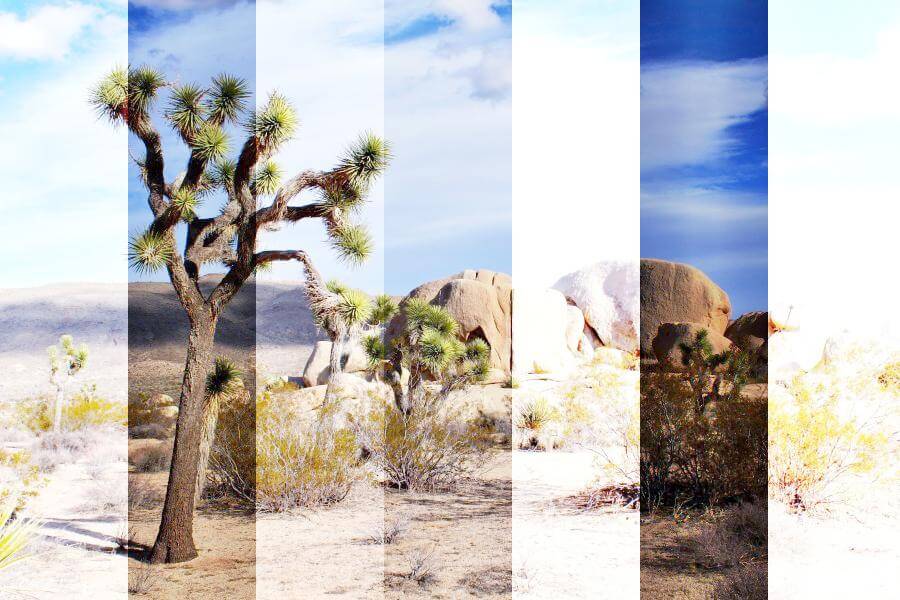}
    }
    \subfloat[SCI~\cite{ma2022toward}]{ 
    \includegraphics[width=0.237\linewidth]{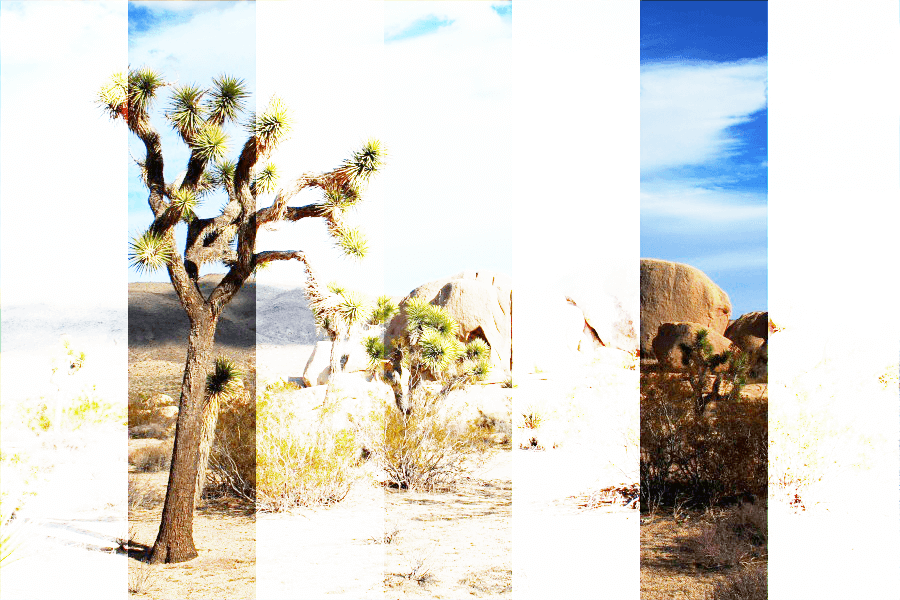}
    }
    \subfloat[URetinexNet~\cite{wu2022uretinex}]{ 
    \includegraphics[width=0.237\linewidth]{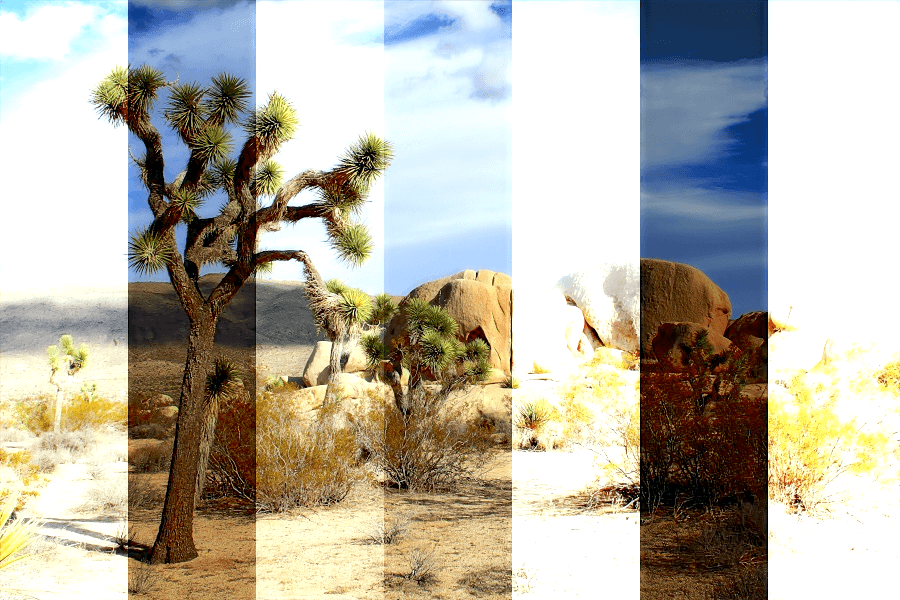}
    }
   
    \caption{\textbf{Visual Comparison on our SICE\_Mix Dataset.} Current low-light image enhancement methods struggles with uneven exposure in SICE\_Mix.}
    \label{fig:SICE_Mix}
\end{figure*}

\noindent \textbf{Results on the SICE\_Grad and SICE\_Mix Dataset:}
Fig. \ref{fig:SICE_Grad} and Fig. \ref{fig:SICE_Mix} display the qualitative comparison for an image from our SICE\_Grad dataset and the SICE\_Mix dataset, respectively. We find that no method yields faithful result on SICE\_Grad or SICE\_Mix. In particular, most methods successfully enhanced the under-exposed regions but made the over-exposed region even brighter. The lack of contrast from the homogeneous over-exposure makes it hard to distinguish any detail in these enhanced regions.

\begin{figure*}[t]
     \centering
     \subfloat[KinD~\cite{zhang2019kindling}]{
     \includegraphics[width=0.237\linewidth]{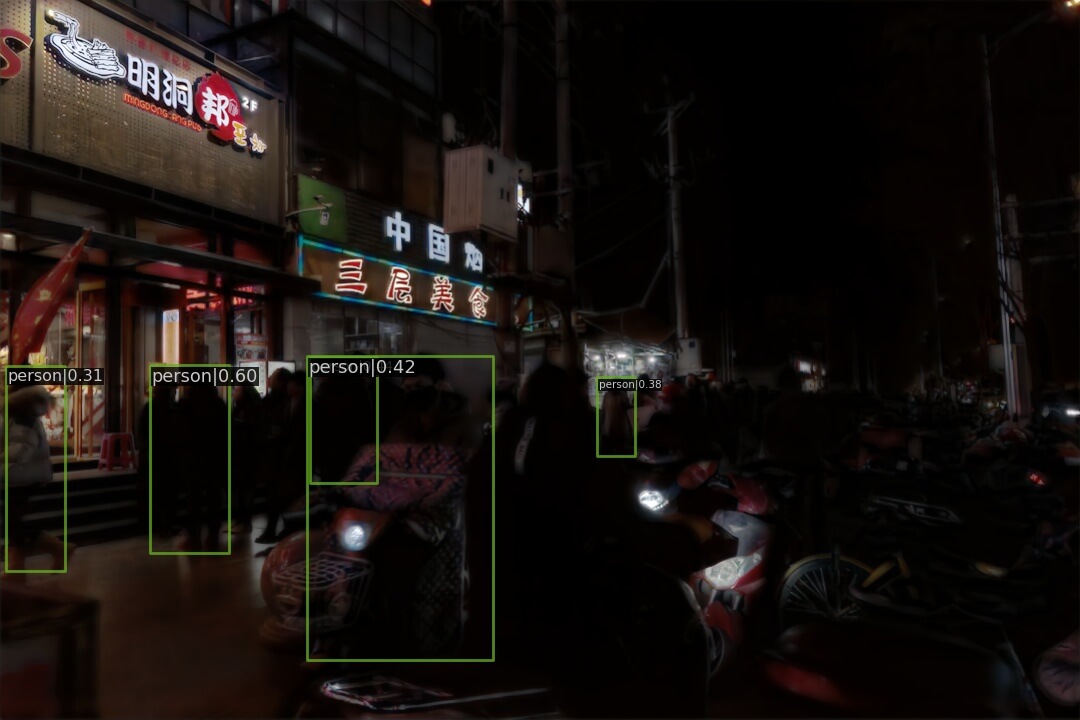}
     }
     \subfloat[Zero-DCE~\cite{guo2020zero}]{ 
     \includegraphics[width=0.237\linewidth]{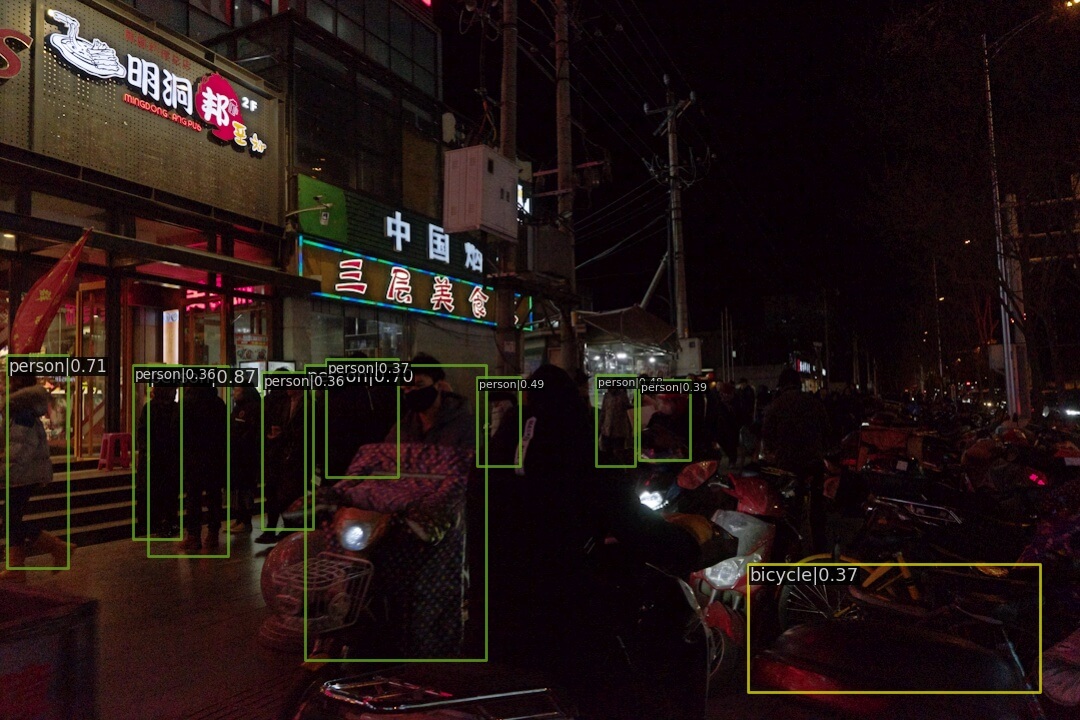}
     }
     \subfloat[KinD++~\cite{zhang2021beyond}]{
     \includegraphics[width=0.237\linewidth]{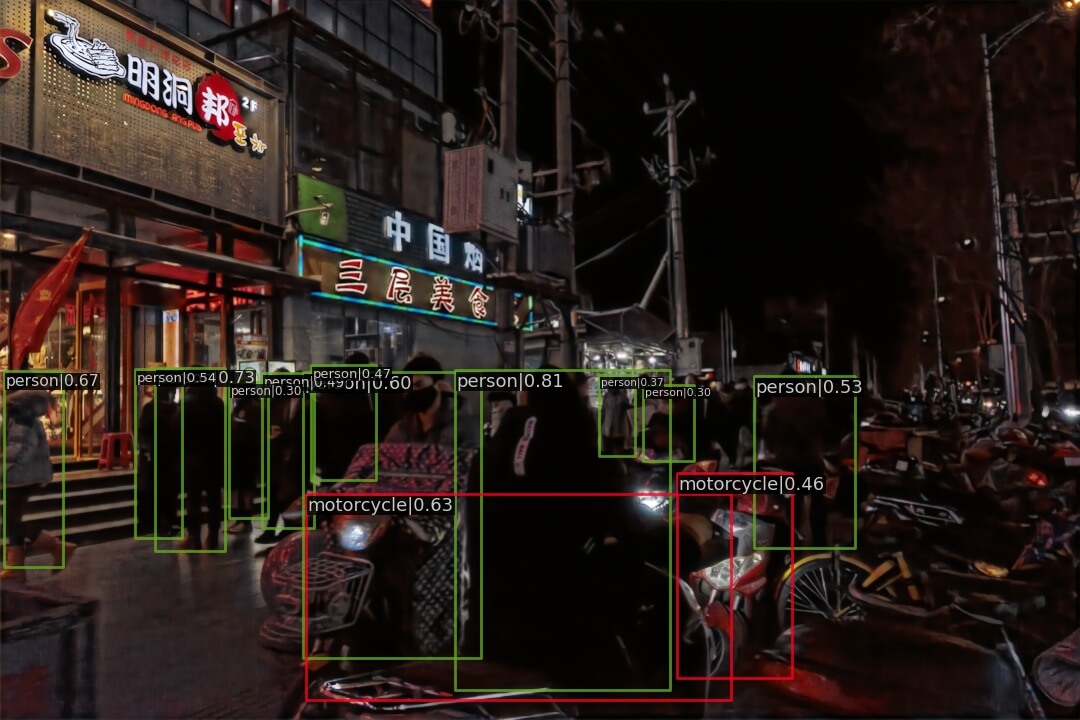}
     }
     \subfloat[RUAS~\cite{liu2021retinex}]{
     \includegraphics[width=0.237\linewidth]{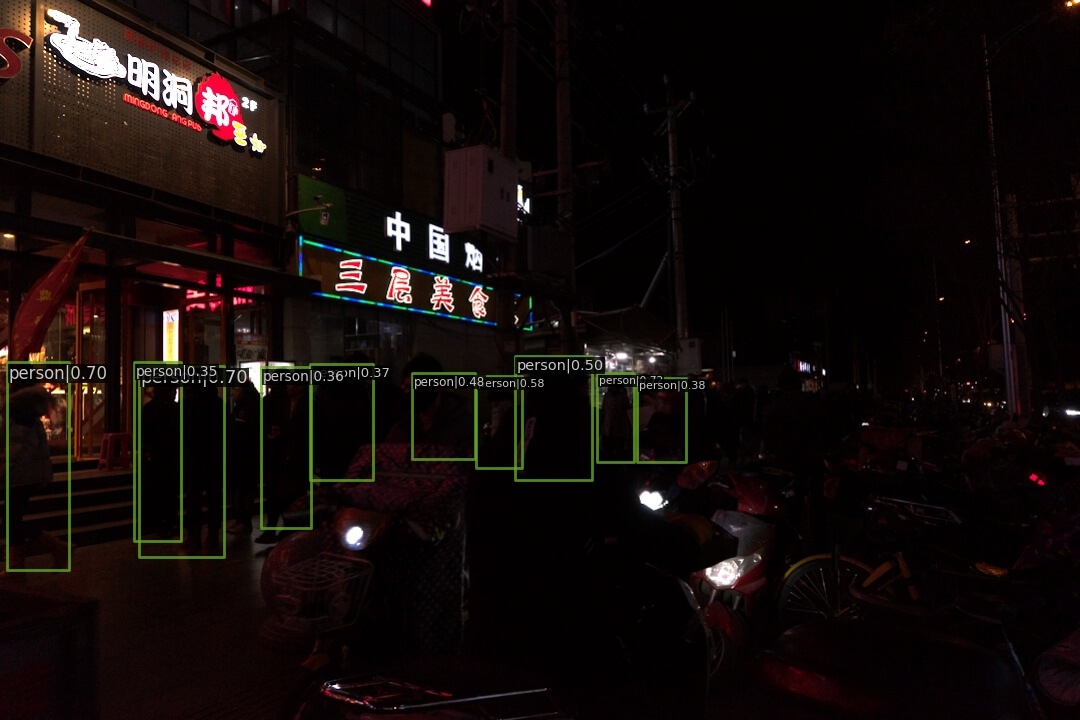}
     }
     \\
    \subfloat[LLFlow~\cite{wang2022low}]{ 
    \includegraphics[width=0.237\linewidth]{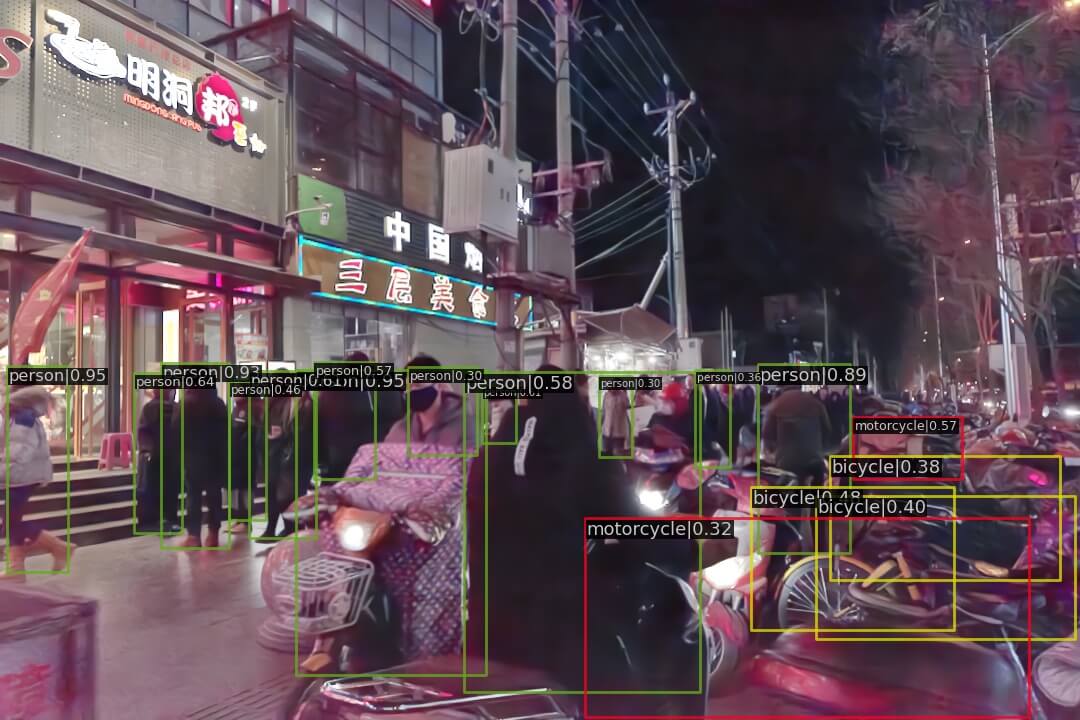}
    }
     \subfloat[SGZ~\cite{zheng2022semantic}]{
     \includegraphics[width=0.237\linewidth]{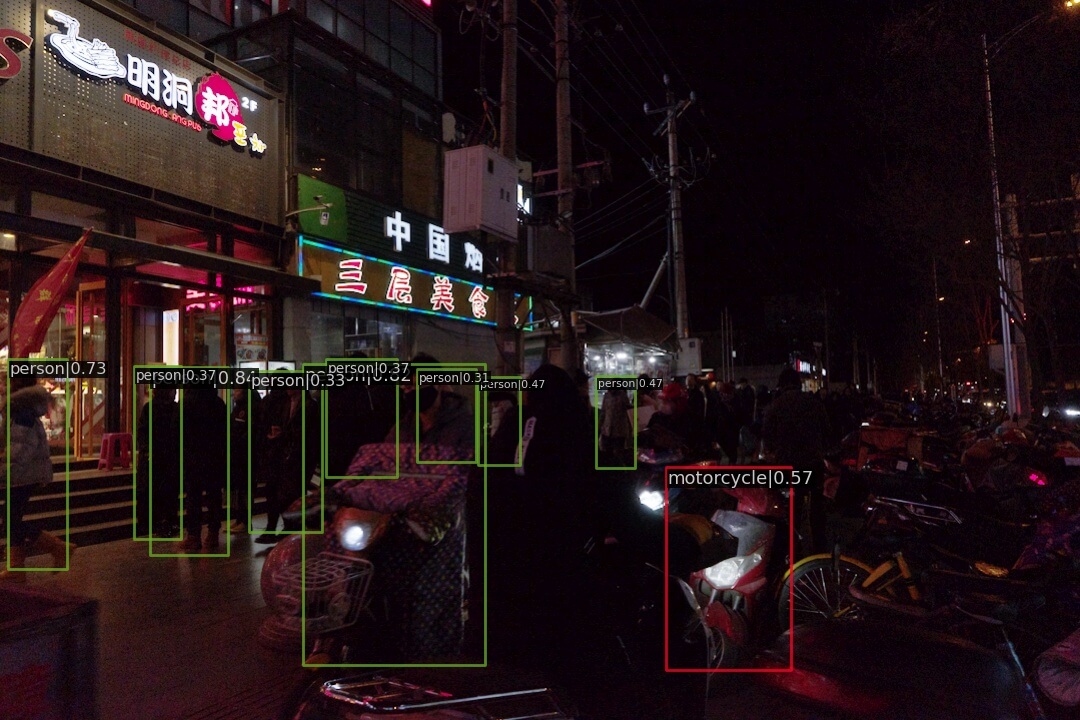}
     }
     \subfloat[SCI~\cite{ma2022toward}]{
     \includegraphics[width=0.237\linewidth]{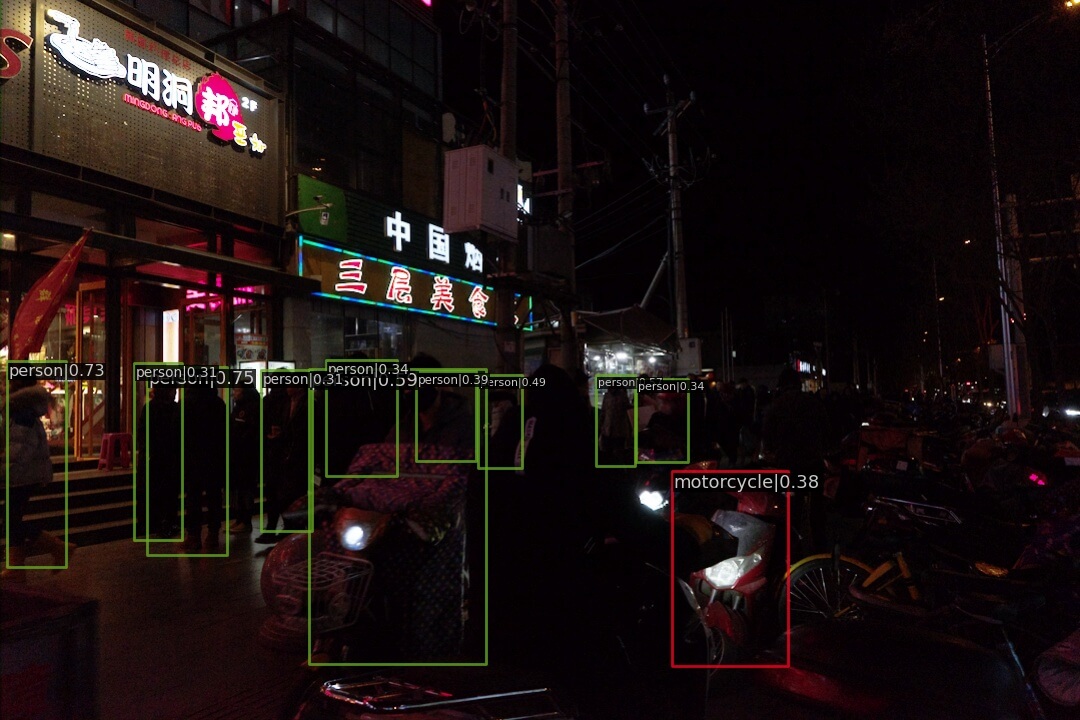}
     }
     \subfloat[URetinexNet~\cite{wu2022uretinex}]{
     \includegraphics[width=0.237\linewidth]{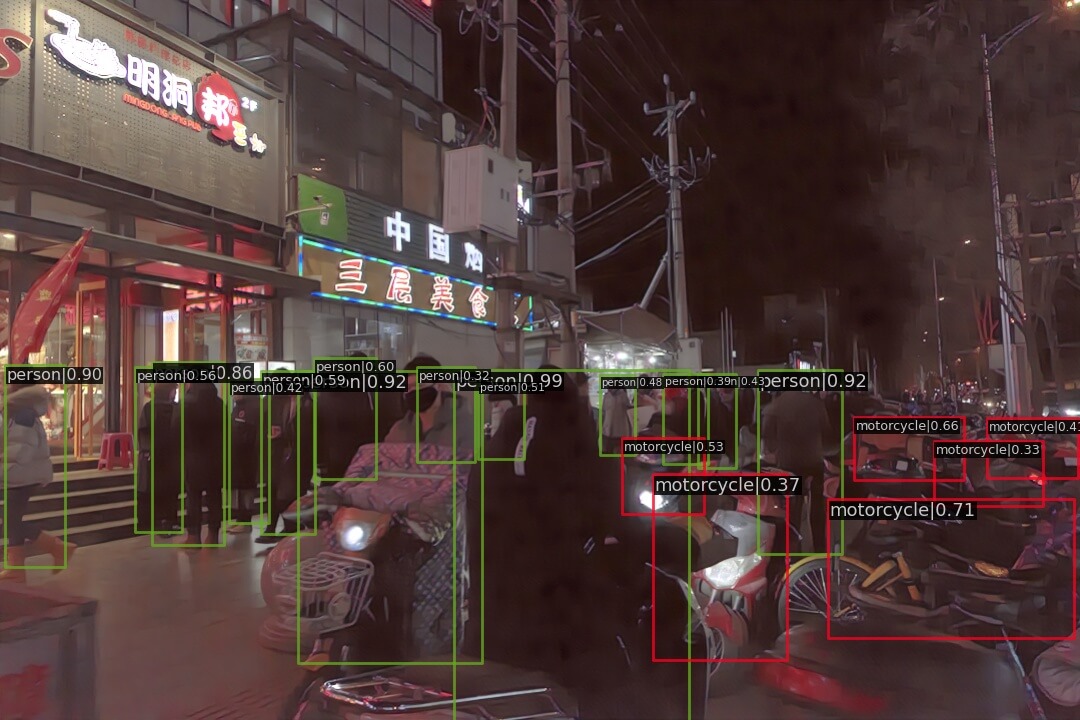}
     }
    \caption{\textbf{Object Detection Visual Comparison on the DarkFace~\cite{yang2020advancing} Dataset.} Image enhancement methods are used as a preprocessing step of object detection. }
    \label{fig:DarkFace_Det}
\end{figure*}

\noindent \textbf{Results on the DarkFace Dataset:} 
  Fig. \ref{fig:DarkFace_Det} shows the qualitative comparison (w/ object detection) for an image from the DarkFace dataset~\cite{yang2020advancing}. In particular, the bounding box in the figure is annotated with the predicted class and probability. Our findings are as follows: 1) KinD~\cite{zhang2019kindling}, Zero-DCE~\cite{guo2020zero}, RUAS~\cite{liu2021retinex}, SGZ~\cite{zheng2022semantic}, and SCI~\cite{ma2022toward} produces under-exposure images, especially for the right half. Therefore, many objects in their enhanced image are not detected. 2) KinD~\cite{zhang2019kindling} produces oversmoothed result. That is why the object detector is way off target in its enhanced image. 3) KinD++~\cite{zhang2021beyond}, URetinexNet~\cite{wu2022uretinex} and LLFlow~\cite{wang2022low} are good in terms of image enhancement. However, both KinD++~\cite{zhang2021beyond} and URetinexNet~\cite{wu2022uretinex} yield artifacts. That's why LLFlow's~\cite{wang2022low} enhancement yields better object detection results.

\begin{figure*}[t]
     \centering
     \subfloat[Dark]{
     \includegraphics[width=0.237\linewidth]{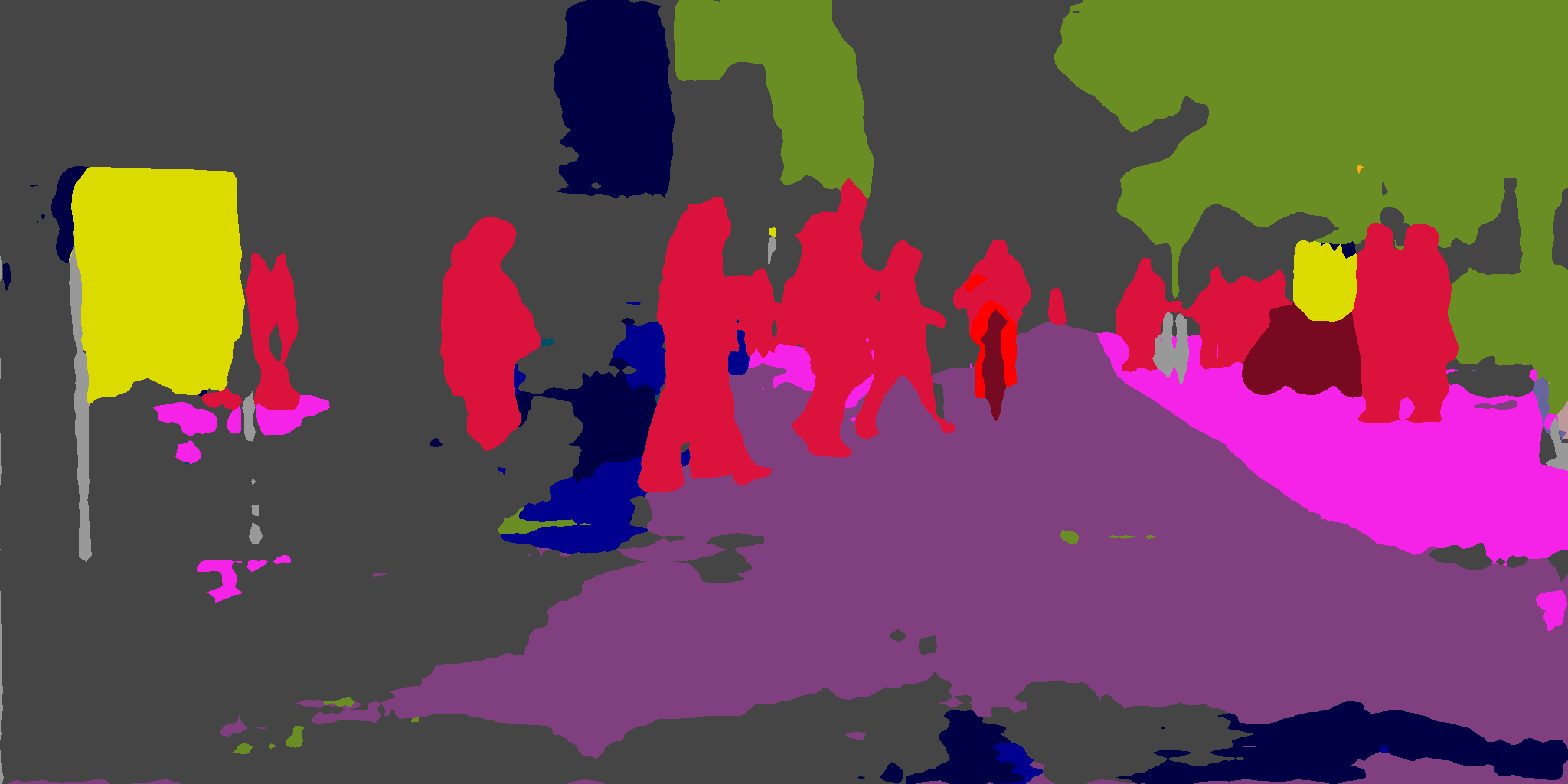}
     }
     \subfloat[PIE~\cite{fu2015probabilistic}]{
     \includegraphics[width=0.237\linewidth]{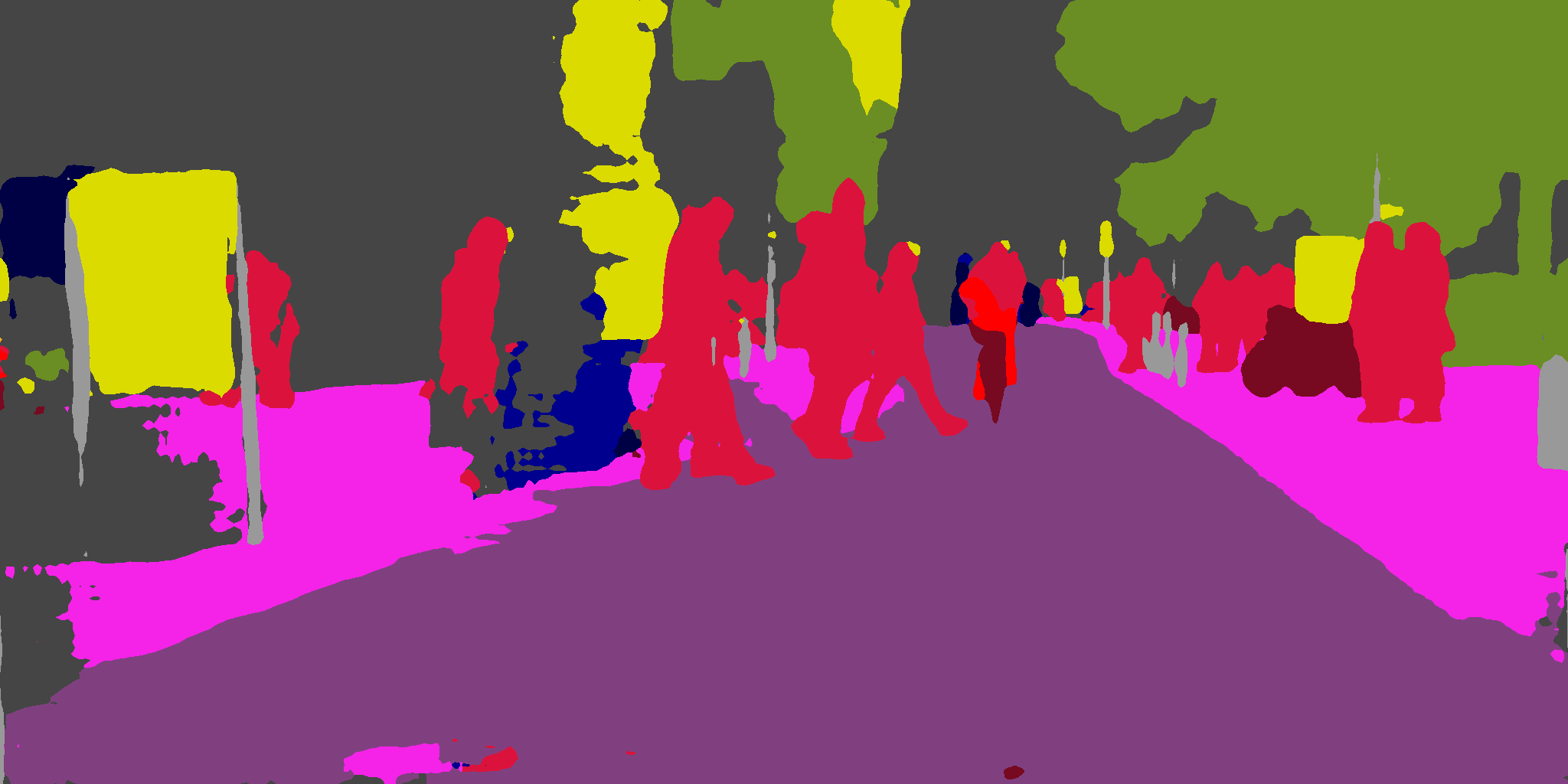}
     }
     \subfloat[RetinexNet~\cite{wei2018deep}]{
     \includegraphics[width=0.237\linewidth]{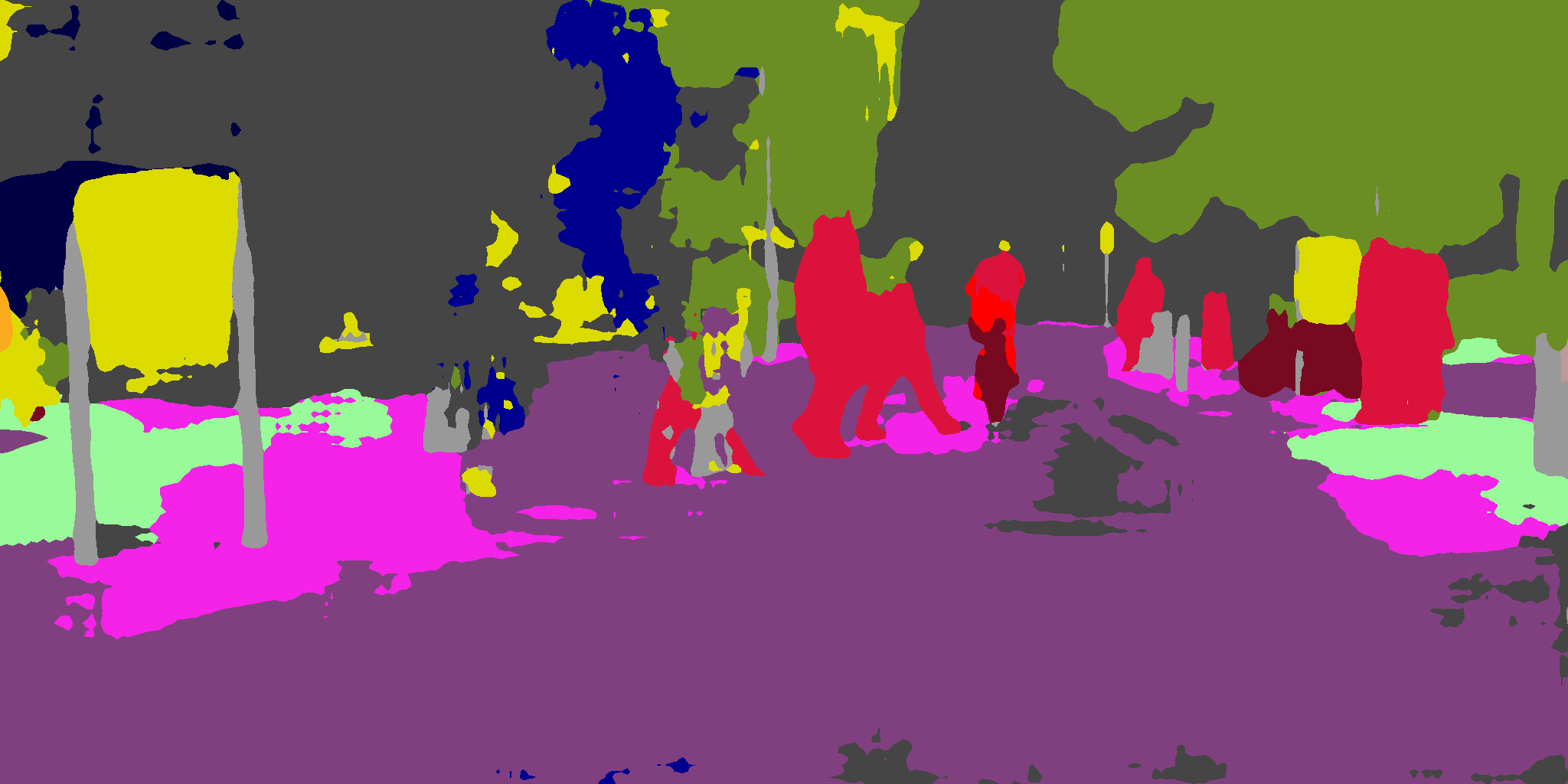}
     }
     \subfloat[MBLLEN~\cite{lv2018mbllen}]{
     \includegraphics[width=0.237\linewidth]{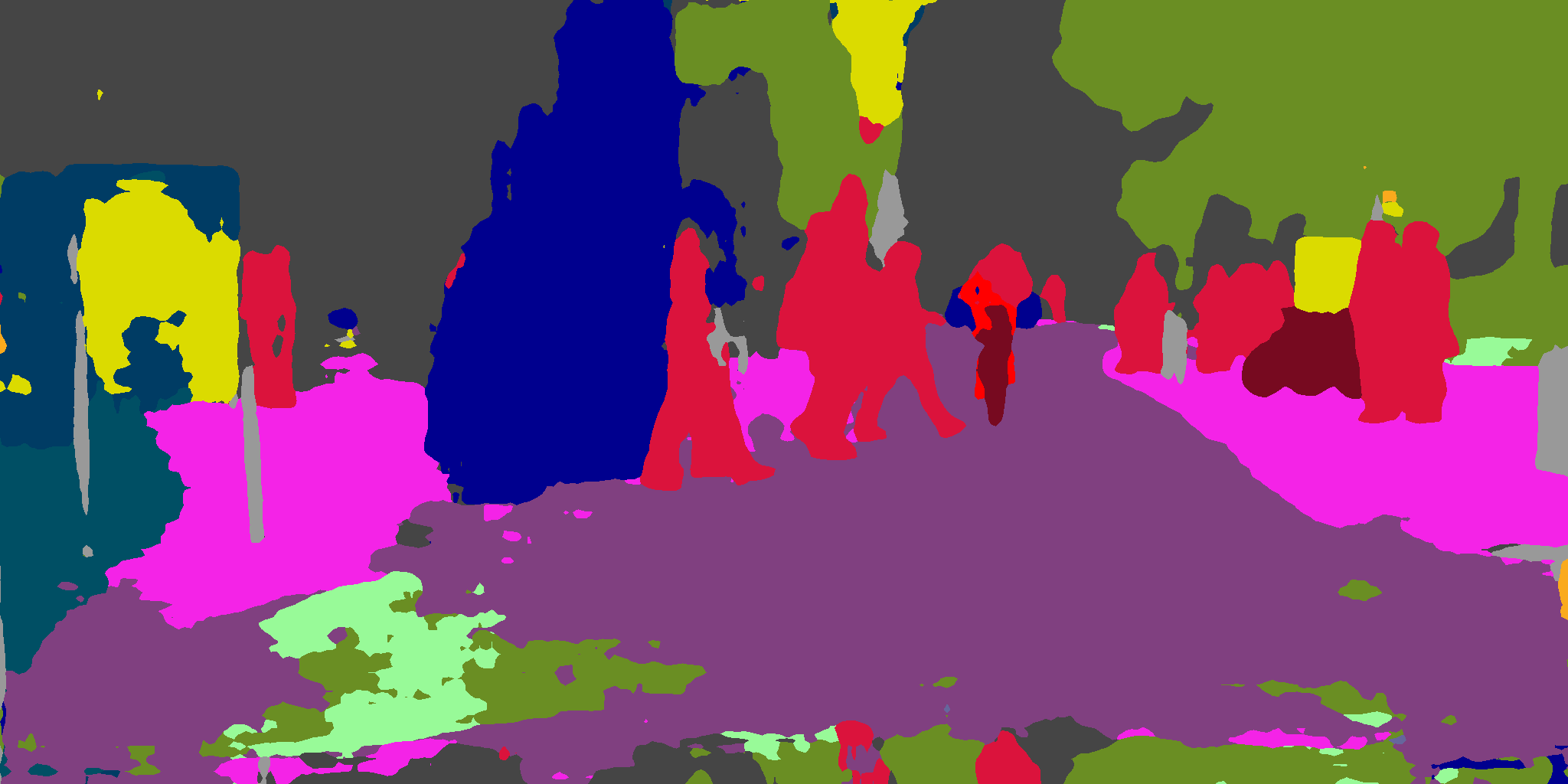}
     }
     \\
     \subfloat[KinD~\cite{zhang2019kindling}]{ 
     \includegraphics[width=0.237\linewidth]{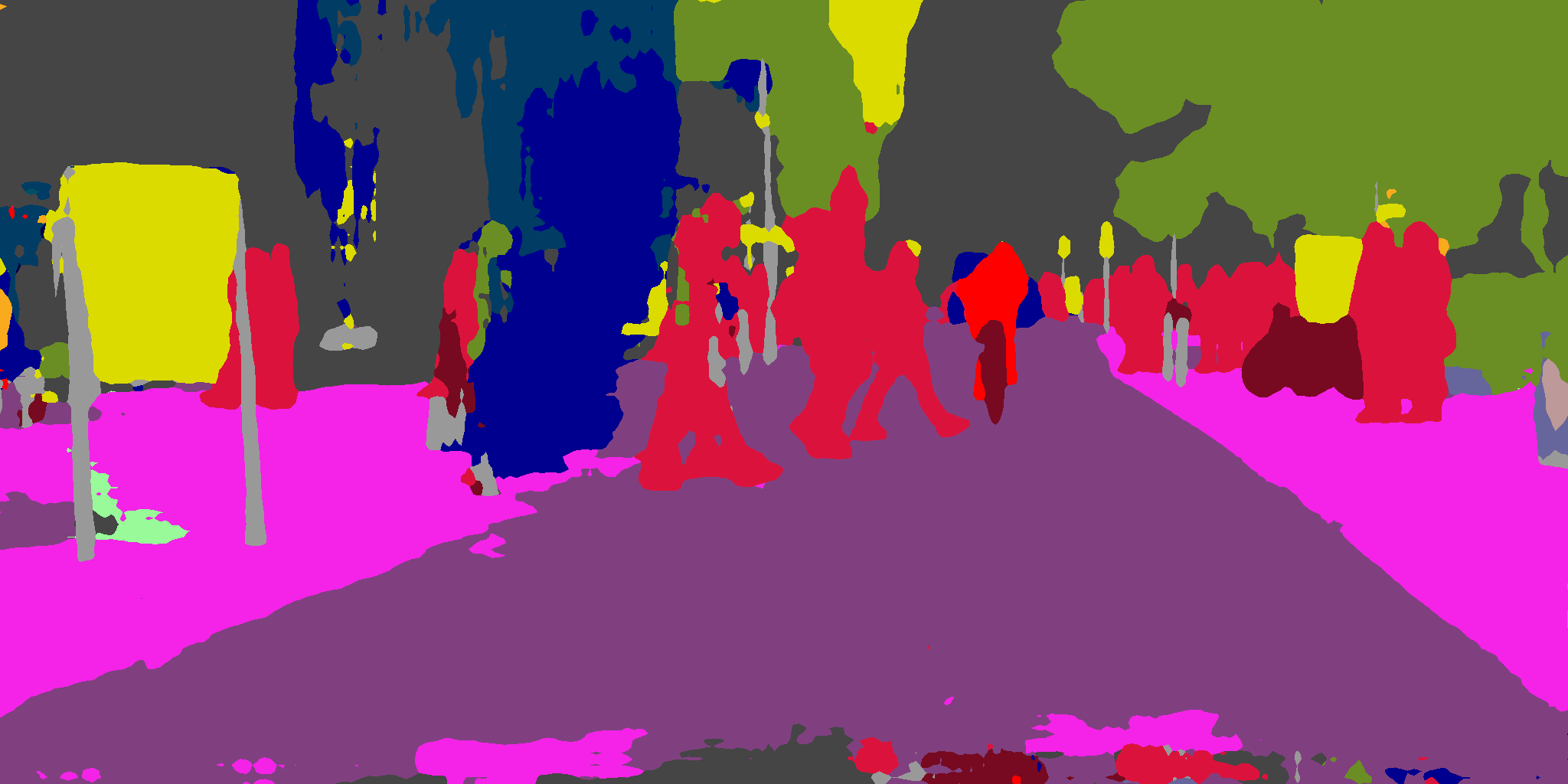}
     }
     \subfloat[Zero-DCE~\cite{guo2020zero}]{
     \includegraphics[width=0.237\linewidth]{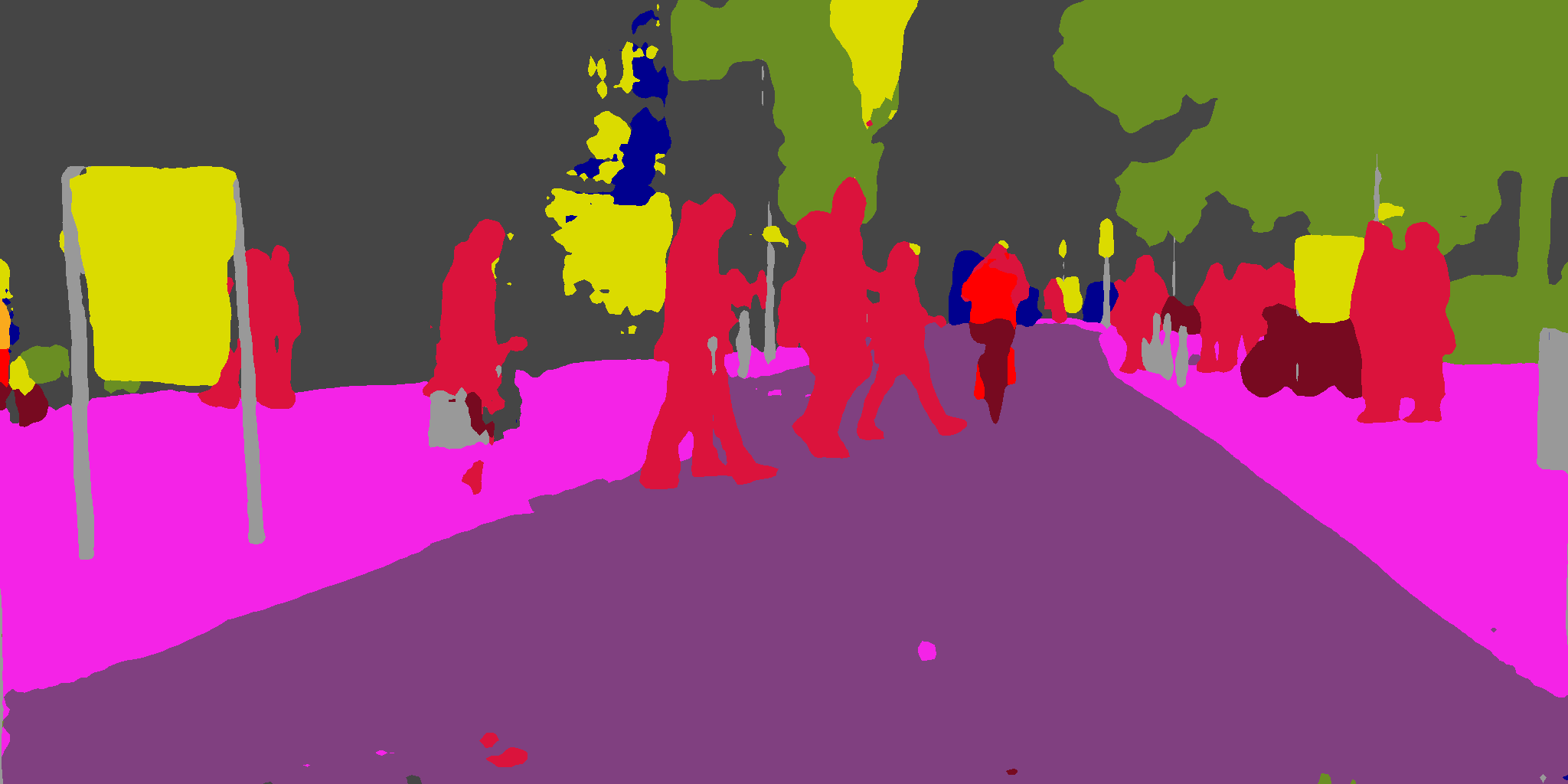}
     }
     \subfloat[SGZ~\cite{zheng2022semantic}]{ 
     \includegraphics[width=0.237\linewidth]{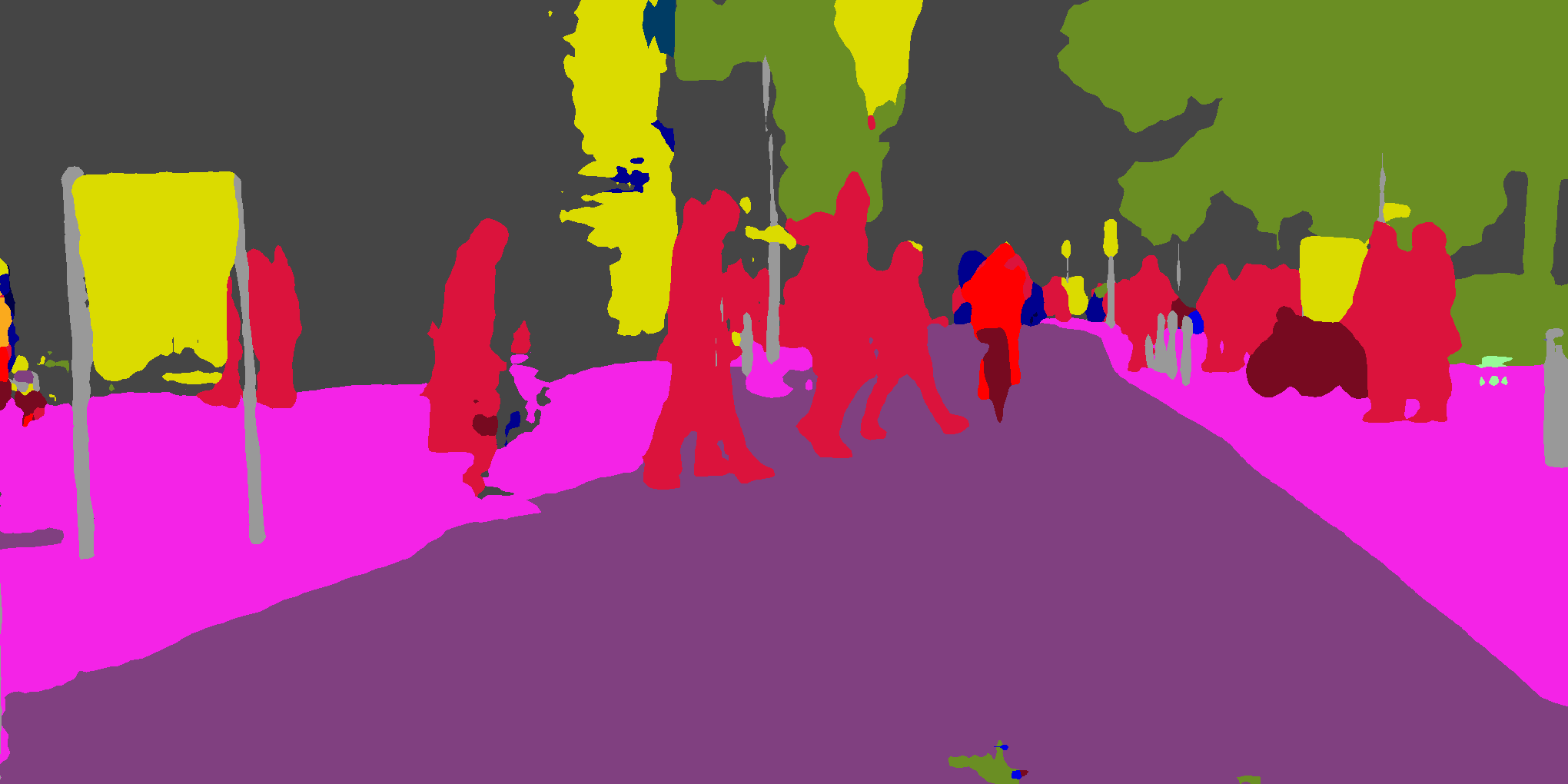}
     }
     \subfloat[Ground Truth]{
     \includegraphics[width=0.237\linewidth]{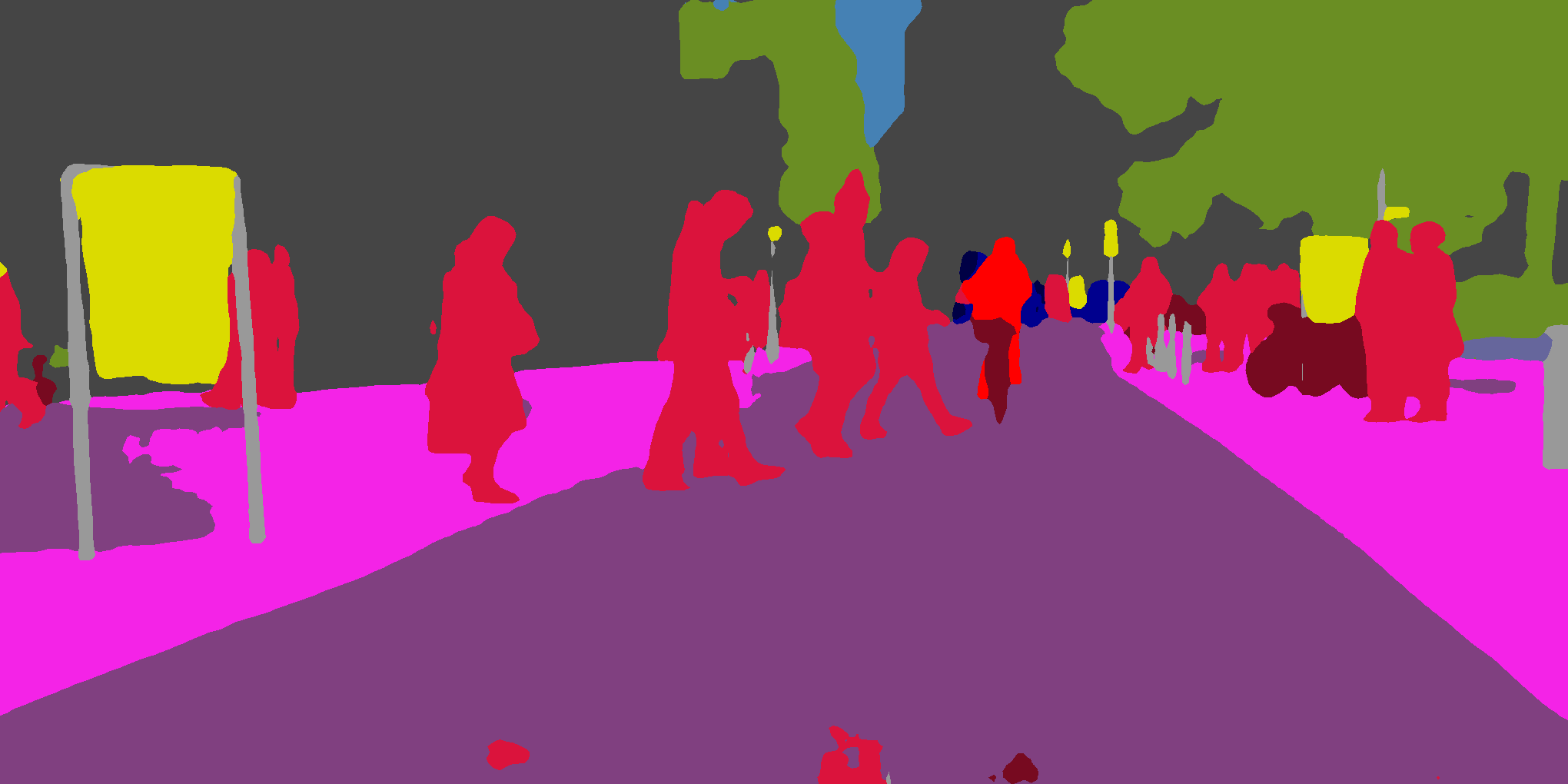}
     }
    \caption{\textbf{Semantic Segmentation Visual Comparison on the DCS~\cite{zheng2022semantic} Dataset.} Image enhancement methods are used as a preprocessing step of semantic segmentation. Ground Truth refers to the segmentation result on the ground truth image. }
    \label{fig:DCS_Seg}
\end{figure*}

\noindent \textbf{Results on the DCS Dataset:} 
Fig. \ref{fig:DCS_Seg} displays the qualitative comparison (w/ semantic segmentation) for an image from the DCS dataset~\cite{zheng2022semantic}. Particularly, different colors are used to distinguish pixels from different predicted categories. Our findings are as follows 1) RetinexNet~\cite{wei2018deep}, MBLLEN~\cite{lv2018mbllen}, and KinD~\cite{zhang2019kindling} causes large areas of incorrect segmentation on pedestrians and sidewalks. 2) Zero-DCE~\cite{guo2020zero} and SGZ~\cite{zheng2022semantic} are close to the GT. However, SGZ~\cite{zheng2022semantic} results in better segmentation results for the objects in the distance.

\begin{figure*}[t]
    \centering
    \subfloat[Dark]{
    \includegraphics[width=0.237\linewidth]{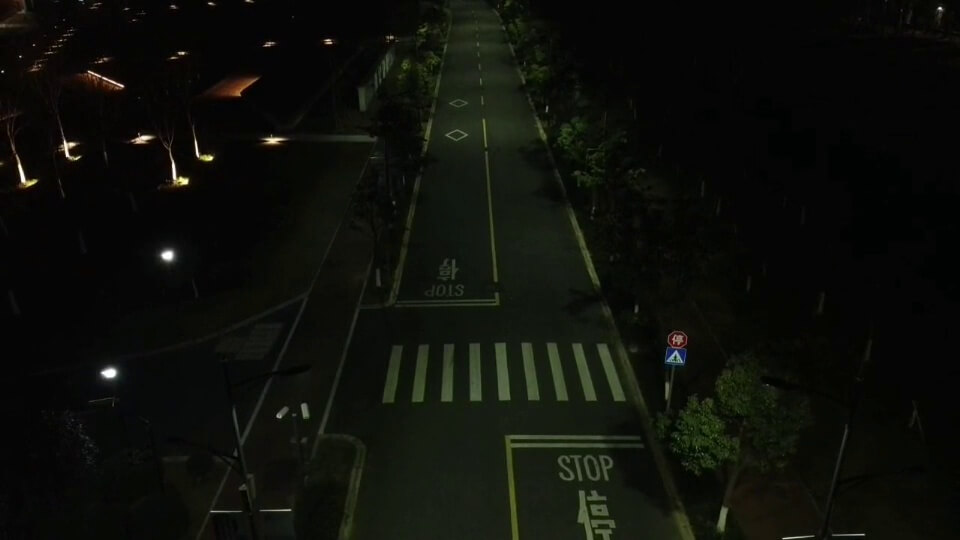}
    }
    \subfloat[RetinexNet~\cite{wei2018deep}]{
    \includegraphics[width=0.237\linewidth]{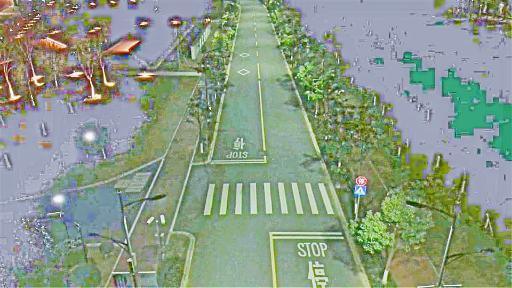}
    }
    \subfloat[MBLLEN~\cite{lv2018mbllen}]{
    \includegraphics[width=0.237\linewidth]{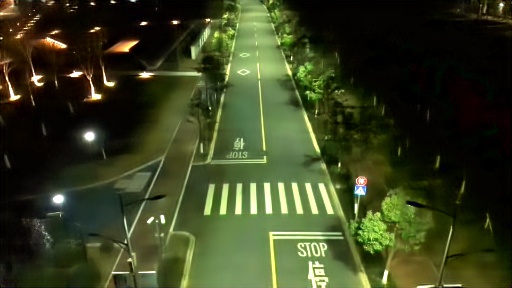}
    }
    \subfloat[KinD~\cite{zhang2019kindling}]{
    \includegraphics[width=0.237\linewidth]{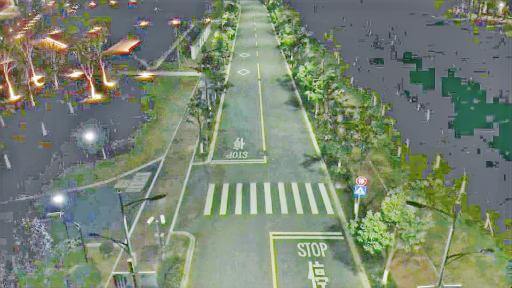}
    }
   \\
    \subfloat[Zero-DCE~\cite{guo2020zero}]{
    \includegraphics[width=0.237\linewidth]{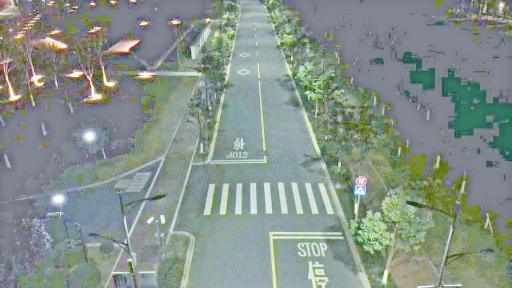}
    }
    \subfloat[KinD++~\cite{zhang2021beyond}]{
    \includegraphics[width=0.237\linewidth]{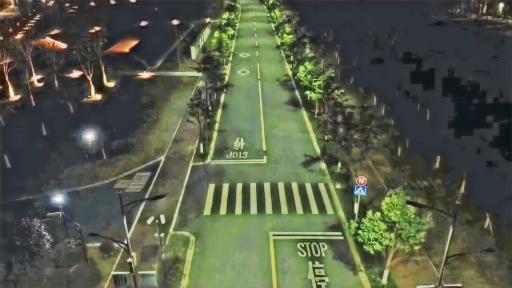}
    }
    \subfloat[EnlightenGAN~\cite{jiang2021enlightengan}]{ 
    \includegraphics[width=0.237\linewidth]{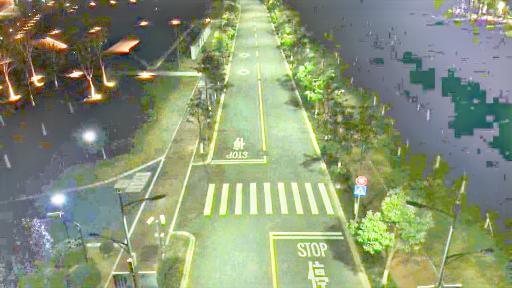}
    }
     \subfloat[SGZ~\cite{zheng2022semantic}]{
     \includegraphics[width=0.237\linewidth]{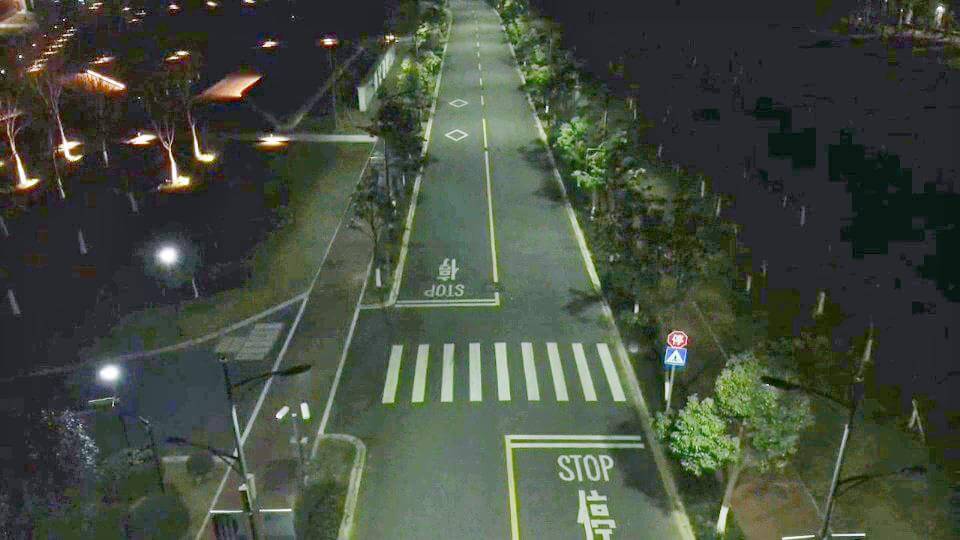}
     }
    \caption{\textbf{Visual Comparison for a Video Frame from our Night Wenzhou Dataset.} Most low-light image enhancement methods face difficulties enhancing images from the Night Wenzhou Dataset.  }
    \label{fig:aerial_video_3}
\end{figure*}

\noindent \textbf{Results on the Night Wenzhou Dataset:} 
Fig. \ref{fig:aerial_video_3} presents the qualitative comparison for a video frame from our Night Wenzhou dataset. Our findings are as below. 1) RetinexNet~\cite{wei2018deep}, KinD~\cite{zhang2019kindling}, and Zero-DCE~\cite{guo2020zero} produces images with poor contrast, extreme color deviation, oversmoothed details, and significant noises, blurs, and artifacts. 2) EnlightenGAN~\cite{jiang2021enlightengan} produces over-exposed images.  3) MBLLEN~\cite{lv2018mbllen}, KinD++~\cite{zhang2021beyond}, and SGZ~\cite{zheng2022semantic} produces images with good exposure. However, MBLLEN~\cite{lv2018mbllen} oversmooths the detail, whereas KinD++~\cite{zhang2021beyond} generates artifacts and has more color deviation than SGZ~\cite{zheng2022semantic}.

\section{Applications} \label{sec:applications} 




    \subsection{Visual Surveillance}

Visual surveillance systems have to operate 24 hours per day to capture all the essential information. However, at low-light conditions like night or dusk, it is challenging for visual surveillance systems to collect images or videos with good contrast and sufficient details~\cite{yang2019coarse}. This may lead to failure to catch the illegal or criminal acts, or failure to capture the party responsible for a specific accident. Over the past years, Kinect depth map~\cite{hu2013kinect}, enhanced CNN-enabled learning~\cite{liu2021enhanced}, and SVD-DWT~\cite{halder2020low} have been used to address the visual surveillance challenges in low-light conditions.

    \subsection{Autonomous Driving}

Real-world autonomous driving often has to operate at nighttime with low-light conditions. Autonomous vehicles perform poorly at low-light conditions because their visual recognition algorithms are trained with visual data captured in good illumination conditions (e.g., sunny day). For example, an autonomous vehicle may fail to recognize (and possibly collide into) a moving pedestrian who is wearing black pants and jacket on a road without street light. Recently, Retinex-Net~\cite{pham2020low}, LE-Net~\cite{li2021deep}, and  SFNet-N~\cite{wang2022sfnet} have been developed to improve the recognition performance for autonomous vehicles at low-light conditions.

    \subsection{Unmanned Aerial Vehicle}

Unmanned Aerial Vehicle (UAV) is often used in aerial photography and military reconnaissance. Both tasks are usually performed in low-light conditions (e.g. dawn, dusk, or night) for better framing or concealment. For the same reason as autonomous driving, UAVs have poor performance at these low-light conditions. For instance, a UAV flying at low-light conditions may collide into a non luminous building or a tree with dark leaves. Recently, HighlightNet~\cite{fu2022highlightnet}, LighterGAN~\cite{wang2021lightergan}, LLNet~\cite{lore2017llnet} and the IOU-predictor network~\cite{du2019visdrone} have been developed to improve the recognition performance for UAV at low-light conditions.

    \subsection{Photography} 


Photography is often done in dawn and dusk. That's because the light at these periods are more gentle, and can help emphasize the shapes of objects. However, the low-light conditions at dawn and dusk significantly degrade the quality of mages captured at these periods. In particular, a higher ISO and a longer exposure time will not only enhance the image brightness but also introduce noise, blurs and artifacts~\cite{yuan2007image}. Recent methods like MorphNet~\cite{gordon2018morphnet}, RNN~\cite{ren2019low} and a learning-based auto white balancing algorithm~\cite{liba2019handheld} is shown helpful for photography at low-light conditions. 

    \subsection{Remote Sensing}

Remote sensing operates at all-weather, and is often used for climate change detection, urban 3D modeling, and global surface monitoring~\cite{campbell2011introduction}. However, low-light conditions make remote sensing challenging, since the object visibility are degraded, especially from the far-distance of the remote satellite. For example, It is difficult for remote sensing satellites to accurately detect the intensity shift of human activities in regions with few artificial light sources. Methods like RICNN~\cite{cheng2016learning}, SR~\cite{haut2018new} and RSCNN~\cite{hu2020rscnn} can help improve the quality of remote sensing images which boost their interpretability.

    \subsection{Microscopic Imaging} 

Microscopic imaging has been widely applied to automatic cell segmentation, fluorescence labeling tracking, and high-throughput experiments~\cite{pepperkok2006high}. However, microscopic imaging is challenging at low-light conditions due to the poor visibility of micro-level details. For instance, it is difficult to identify the cell structures with the microscope under low-light conditions. VELM~\cite{shotton1988video}, SalienceNet~\cite{bouilhol2022saliencenet} and CARE network~\cite{weigert2018content} can be applied to improve the visibility outcomes of optical microscopy and the video documentation of organelle motion processes at low-light conditions. 

    \subsection{Underwater Imaging}

Underwater imaging often occurs low-light conditions, since the lightening in deep water in very weak. This is a challenge for underwater imaging, since the visibility in these conditions is very poor, especially for the object in the far distance~\cite{li2019underwater}. For example, with the existing camera technology, it is difficult to clearly capture the details of coral reefs 30 meters below the sea level~\cite{goodman2013coral}. L$^{2}$UWE~\cite{marques2020l2uwe}, UIE-Net~\cite{wang2017deep}, and  WaterGAN~\cite{li2017watergan} have been proven useful for underwater imaging at low-light conditions. 




\section{Future Prospects} \label{sec:future_prospects} %

    \subsection{Uneven Exposure}

    In the Low-Light Image Enhancement (LLIE) domain, addressing uneven exposure remains a pressing issue. While current strategies proficiently brighten under-exposed regions, they tend to over-amplify already bright areas. The ideal LLIE technique would concurrently amplify dim sections and reduce brightness in over-illuminated zones. Future investigations could chart a promising course by creating a comprehensive real-world dataset, spotlighting images with both over- and under-exposure. Our contributions with SICE\_Grad and SICE\_Mix are initial steps in this direction, but there's a pressing need for more encompassing datasets. 
    
    Henceforth, embracing frameworks such as the Laplacian Pyramid, exemplified in DSLR~\cite{lim2020dslr}, or the multi-branch fusion approach from TBEFN~\cite{lu2020tbefn}, can help grasp the variances of multi-scale exposure. Additionally, the emerging realm of vision transformers, notably highlighted in~\cite{dosovitskiy2020image}, offers a unique perspective with their superior capacity to comprehend global exposure nuances compared to traditional CNNs. The preliminary work of IAT~\cite{cui2022illumination} in this space underscores potential avenues for refining techniques to model intricate exposure scenarios.

    


    \subsection{Preserving and Utilizing Semantic Information}

    Brightening low-light images without sacrificing their semantic information is a pivotal challenge in low-light image enhancement (LLIE). This careful balance affects both human interpretation and the performance of high-level algorithms. The significance of semantic information is evident in Yang et al.'s work~\cite{yang2018visual}, where semantic priors aid in navigation tasks. Similarly, Xie et al.~\cite{xie2018learning} underscore the potential mishaps in domain adaptation when semantic details are overlooked. SAPNet~\cite{zheng2022sapnet} combines image processing with semantic segmentation, emphasizing the importance of retaining semantic details.
    
    Going forward, we envision the development of datasets enriched with semantic annotations to guide LLIE models. Additionally, the integration of segmentation techniques with LLIE promises images that are both luminous and semantically robust. The convergence of these strategies could pave the way for the next generation of LLIE solutions.





    \subsection{Low-Light Video Enhancement} 
        

    Low-light video enhancement (LLVE) presents unique challenges not found in image enhancement. Real-time processing is pivotal for videos, but many existing methods struggle to achieve this criterion. Even when real-time requirements are met, these methods can introduce temporal inconsistencies like flickering artifacts~\cite{li2021low}. Chen et al.~\cite{chen2019seeing} ventured into extreme low-light video enhancement, underscoring the challenge of obtaining dynamic scene ground truths. Triantafyllidou et al.~\cite{triantafyllidou2020low} proposed a synthetic approach to mitigate data collection hurdles, creating dynamic video pairs. Jiang et al.~\cite{jiang2019learning} focused on temporal consistency but their solution leans on specific preconditions or synthetic datasets. Zhang et al.~\cite{zhang2021learning} explored inferred motion from single images to remedy temporal inconsistencies, indicating the persistent challenge in ensuring stability across diverse scenes.
    
    In essence, while advancements address aspects of temporal consistency, achieving it robustly in varied scenes remains an open challenge. The focus should shift toward enabling real-time processing, capturing diverse scene intricacies, and managing motion-related complexities. Future strategies might also benefit from lightweight architectures~\cite{howard2017mobilenets} and more comprehensive low-light video datasets.



    \subsection{Benchmark datasets}

    Presently, there's an absence of a universally-accepted benchmark dataset for LLIE. This poses challenges on two fronts. Firstly, many LLIE models are trained on proprietary datasets in a supervised way, which might lead to domain-specific solutions that struggle to generalize. Secondly, testing on custom datasets can introduce bias, as they could be overly tailored to a specific method, thus not offering a fair playing field for comparisons.
    
    Drawing inspiration from the success of CityScapes~\cite{cordts2016cityscapes} and BDD100k~\cite{yu2020bdd100k}, a reliable benchmark should include diverse real-world images and videos, ensure a balanced train-test split, and come with comprehensive annotations. Importantly for LLIE, this dataset should also span a wide range of lighting conditions and exposures to truly test the robustness and versatility of enhancement methods.



    \subsection{Better Evaluations Metrics}

    Current evaluation metrics for LLIE, such as PSNR and SSIM, often fail to capture the nuances important for human perceptual judgments. As a result, many researchers turn to user studies, which provide a more realistic assessment of image quality but are expensive and time-consuming. Deep learning-based metrics, harnessing the advanced perceptual capabilities of architectures like CNNs or transformers, offer an efficient alternative.
    
    For instance, transformer-based metrics such as MUSIQ~\cite{ke2021musiq} and MANIQA~\cite{yang2022maniqa}, and CNN-based metrics like HyperIQA~\cite{su2020blindly} and KonIQ++~\cite{su2021koniq++}, suggest a path forward. Future endeavors should be targeted at developing metrics of this kind that align with human judgments, while also being efficient and scalable for LLIE evaluations.

    \subsection{Low-Light and Adverse Weather}



    Vision in joint adverse weather and low-light conditions is inherently challenging. When combined, these conditions introduce unique complexities as the effects of limited lighting intertwine with disturbances from bad weather like rain, snow, haze, etc. A few pioneering works have ventured into this domain. Li et al.~\cite{li2015nighttime} proposes a fast haze removal method for nighttime images, leveraging a novel maximum reflectance prior to estimate ambient illumination and restore clear images.  ForkGAN~\cite{zheng2020forkgan} introduces a fork-shaped generator for task-agnostic image translation, enhancing multiple vision tasks in rainy night without explicit supervision.
    
    Current challenges in this field encompass handling multiple degradations at once, the lack of diverse datasets for all conditions, and the need for real-time processing, especially in applications like autonomous driving. Future research is likely to focus on developing versatile models, enriching datasets to reflect real-world adversities, fine-tuning network structures, and using domain adaptation to bridge knowledge across varying conditions.


    

    \subsection{Near-Infrared (NIR) Light Techniques}

    Emerging methods underscore the transformative role of NIR in low-light image enhancement. Wang et al.~\cite{wang2019stereoscopic} utilize a dual-camera system combining conventional RGB and near-infrared/near-ultraviolet captures to produce low-noise images without visible flash disturbances. Wan et al.~\cite{wan2022purifying} employ NIR enlightened guidance and disentanglement to refine low-light images by separating structure and color components. 

    Despite these innovations, challenges remain: seamless hardware integration for synchronized RGB-NIR capture, precise image fusion without quality compromise, and the development of robust dual-spectrum datasets. Looking forward, research should optimize camera configurations, advance fusion algorithms, and curate comprehensive datasets. Addressing these can unlock NIR's full potential in low-light imaging.

    \subsection{Noise Distribution Modeling} 

    Recent advances in noise distribution modeling have taken significant strides in addressing low-light denoising. Wei et al.~\cite{wei2020physics} present a model that emulates the real noise from CMOS photosensors, focusing on synthesizing more realistic training samples. Feng et al.~\cite{feng2022learnability} respond to the learnability constraints of limited real data, introducing shot noise augmentation and dark shading correction. Lastly, Monakhova et al.~\cite{monakhova2022dancing} push the boundaries with a GAN-tuned physics-based noise model, capturing photorealistic video under starlight.
    
    However, the contemporary domain still grapples with refining noise models, enhancing the richness of training datasets, and managing the complexities of noise distributions in varied scenarios. Future endeavors should concentrate on improving noise modeling accuracy, expanding diverse training datasets, and integrating physics-based and data-driven techniques.

    \subsection{Joint Enhancement and Detection} 

    Joint enhancement and detection in low-light conditions are gaining traction in the research community. Wang et al.~\cite{wang2022unsupervised} have proposed an adaptation technique to train face detectors for low-light scenarios without the need for specific low-light annotations. Ma et al.~\cite{ma2022pia} focus on a parallel architecture, using an illumination allocator to achieve simultaneous low-light enhancement and object detection. 
    
    
    The current challenge lies in effectively transferring knowledge from normal lighting to low-light conditions and ensuring these models generalize well in diverse scenarios. Future work should focus on optimizing architectures for both enhancement and detection tasks, improving adaptability, and further exploring joint preprocessing techniques, as seen in other fields like deblurring~\cite{zheng2021deblur} and denoising~\cite{liu2021end}.

        


\section{Conclusion} \label{sec:conclusion}

This paper presents a comprehensive survey of low-light image and video enhancement. Firstly, we propose SICE\_Grad and SICE\_Mix to simulate the challenging over-/under-exposure scenes under-performed by current LLIE methods. Secondly, we introduce Night Wenzhou, a large-scale, high-resolution video dataset with various illuminations and degradation. Thirdly, we analyze the critical components of LLIE methods, including learning strategy, network structures, loss functions, evaluation metrics, training and testing datasets, etc. We also discuss the emerging system-level applications for low-light image and video enhancement algorithms. Finally, we conduct qualitative and quantitative comparisons of LLIE methods on the benchmark datasets and the proposed datasets to identify the open challenges and suggest future directions.

\bibliography{egbib.bib}
\bibliographystyle{IEEEtran}




\end{document}